\documentclass{article}
\usepackage[T1]{fontenc}
\usepackage[utf8]{inputenc}
\usepackage{amsmath,amsfonts,amsthm,amssymb}
\usepackage{setspace}
\usepackage{fullpage}
\usepackage{fancyhdr}
\usepackage{xspace, xcolor}

\usepackage{verbatim}
\usepackage{lastpage}
\usepackage{extramarks}
\usepackage{natbib}
\usepackage{diagbox}
\usepackage{booktabs}
\usepackage{titletoc}

\usepackage{times}
\usepackage{setspace}
\usepackage{graphicx,float,wrapfig}
\usepackage{algorithm}
\usepackage{enumitem}
\usepackage[noend]{algpseudocode}
\usepackage{subcaption}
\usepackage{xspace}
\usepackage{multirow}

\usepackage{xltabular}
\newcolumntype{L}{>{\RaggedRight\hspace{0pt}}X}

\allowdisplaybreaks

\usepackage{hyperref}
\hypersetup{
	colorlinks   = true, %
	urlcolor     = red, %
	linkcolor    = blue, %
	citecolor   = blue %
}

\usepackage{listings}

\usepackage{microtype}
\usepackage{graphicx}
\usepackage{subcaption}
\usepackage{booktabs} %
\usepackage{dblfloatfix}

\usepackage{hyperref}

\usepackage{amsmath}
\usepackage{amssymb}
\usepackage{mathtools}
\usepackage{amsthm}

\usepackage[capitalize,noabbrev]{cleveref}

\theoremstyle{plain}

\theoremstyle{definition}

\theoremstyle{remark}

\newcommand{\ourmethod}{DC-CoT\xspace}
\newcommand{\pt}{parallel thinking\xspace}
\newcommand{\firstfilter}{\textit{Include-Easy}\xspace}
\newcommand{\secondfilter}{\textit{Remove-Easy}\xspace}

\newcommand{\EOS}{\texttt{[EOS]}}
\newcommand{\infer}{\textsc{Infer}}

\newcommand{\plen}{\textup{lpl}}
\newcommand{\termination}{\texttt{finish}}
\newcommand{\Wstarti}[1]{\texttt{"<worker\_#1>"}}
\newcommand{\Wendi}[1]{\texttt{"</worker\_#1>"}}

\newcommand{\singlethreadone}{S_1}
\newcommand{\wonetext}{W_1}
\newcommand{\wtwotext}{W_2}
\newcommand{\wthreetext}{W_3}

\newcommand{\singlethreadtwo}{S_2}

\newcommand{\Wstart}{\texttt{"<spawn\_workers>"}}
\newcommand{\Wend}{\texttt{"</spawn\_workers>"}}
\newcommand{\prompt}{P}
\newcommand{\tempprompt}{\prompt_{\textup{temp}}}
\newcommand{\completion}{\mathrm{S}}

\newcommand{\initrolloutdata}{\mathcal{D}_{\textup{init}}}
\newcommand{\parallelrewritedata}{\mathcal{D}_{\textup{rewrite}}}

\newcommand{\initialproblemset}{\mathcal{P}_{\textup{init}}}
\newcommand{\problem}{p}
\newcommand{\WA}[1]{W_{#1}^{\textup{(gen)}}}
\newcommand{\WB}[1]{W_{#1}^{\textup{(prefill)}}}
\newcommand{\WAfig}[1]{W_{#1}^{\textup{(gen)}}}
\newcommand{\WBfig}{W_{123}^{\textup{(pre)}}}

\newcommand{\model}{M}
\newcommand{\basemodel}{M_{\textup{base}}}
\newcommand{\sftmodel}{M_{\textup{SFT}}}

\newcommand{\rewritemodel}{M_{\textup{rewrite}}}

\newcommand{\epslow}{\varepsilon_{\textup{low}}}
\newcommand{\epshigh}{\varepsilon_{\textup{high}}}
\newcommand{\refmodel}{\pi_{\textup{ref}}}
\newcommand{\currentpolicy}{\pi_{\theta}}
\newcommand{\oldpolicy}{\pi_{\textup{old}}}
\newcommand{\response}{o}
\newcommand{\responseseq}{o_{\textup{seq}}}
\newcommand{\maxplen}{L_{\textup{max}}}
\newcommand{\penaltycutoff}{L_{\textup{cutoff}}}
\newcommand{\reward}{R}

\newcommand{\rewardfn}{g}
\newcommand{\correctformatreward}{f}
\newcommand{\groupsize}{G}
\newcommand{\normadv}{\hat{A}}
\newcommand{\klpenalty}{\textsc{KL}}
\newcommand{\advmean}{\mu}
\newcommand{\advstd}{\sigma}

\newcommand{\lengthpenaltyfn}{\ell}
\newcommand{\clip}{\textup{clip}}

\newcommand{\lpcoeff}{C_L}

\renewcommand{\Return}{\textbf{return }}

\newcommand{\claudemodel}{\texttt{claude-sonnet-4-5@20250929}}

\def\shownotes{0}  %
\ifnum\shownotes=1
\newcommand{\authnote}[2]{[#1: #2]}
\else
\newcommand{\authnote}[2]{}
\fi

\usepackage{url}
\usepackage{xurl}
\usepackage{amsfonts}
\usepackage{algorithm}
\usepackage{algpseudocodex}
\usepackage{enumitem}
\usepackage{tcolorbox}
\tcbuselibrary{breakable}
\usepackage{fancyvrb}
\usepackage{fvextra}
\usepackage{listings}
\usepackage{array}
\usepackage{multirow}
\usepackage{diagbox}
\usepackage{xspace}
\usepackage[T1]{fontenc}

\usepackage[table]{xcolor}
\usepackage{makecell}

\newtcolorbox[auto counter]{mybox}[2][]{
    colback=white,
    colframe=gray,
    coltitle=black,
    colbacktitle=gray!30,
    title=Box~\thetcbcounter: #2,
    fonttitle=\bfseries,
    label=#1,
    breakable
}

\usepackage{standalone}
\usepackage{tikz}
\usetikzlibrary{shapes,arrows,positioning,calc}

\usepackage{cleveref}

\setlist[itemize]{leftmargin=*, labelsep=0.5em, itemsep=2pt, topsep=2pt}

\crefname{tcb@cnt@mybox}{Box}{Boxes}
\Crefname{tcb@cnt@mybox}{Box}{Boxes}

\newcommand{\arrowfont}{\Huge}  %

\newcommand{\verticaldist}{4.0cm}    %
\newcommand{\horizontaldist}{2.0cm}  %

\tikzset{
    inference_box/.style={
        rectangle,
        rounded corners,
        draw=black,
        thick,
        align=left,
        inner sep=6pt,
        font=\Huge  %
    },
    inference_problem/.style={inference_box, fill=yellow!20, text width=20cm},
    inference_solution/.style={inference_box, fill=green!20, text width=40cm},
    inference_worker/.style={inference_box, fill=red!20, text width=20cm},
    inference_arrow/.style={->, >=stealth, thick}
}

\newcommand{\gridheight}{1.2cm}  %
\newcommand{\gridfont}{\normalsize}  %

\tikzset{
    gridcell/.style={
        rectangle,
        draw=black,
        thick,
        minimum height=\gridheight,
        align=center,
        font=\gridfont,
        inner xsep=0.3cm,  %
        inner ysep=0.3cm   %
    }
}

\newcommand{\attentioncellcm}{0.95}          %

\newcommand{\attentiongencellcm}{1.12}       %

\newcommand{\attentionwidecellcm}{1.62}      %

\newcommand{\attentionfont}{\small}

\newcommand{\attentiontickgap}{0.55}         %

\newcommand{\attentionaxisfont}{\small}
\newcommand{\attentionaxisgap}{0.55}         %

\usetikzlibrary{shapes.geometric, arrows.meta, fit, positioning, calc, backgrounds}
\pgfdeclarelayer{farback}
\pgfdeclarelayer{back}
\pgfsetlayers{farback,back,main}
\makeatletter
\newcommand{\flowTitleFont}{%
  \fontsize{\numexpr\f@size+2\relax}{\numexpr\f@size+4\relax}\selectfont
}
\makeatother
\tikzset{
  outerbox/.style={
    rectangle, rounded corners, 
    draw=none,
    fill=black!10,
    inner xsep=8pt, inner ysep=6pt
  },
  subbox1/.style={
    rectangle, rounded corners, 
    draw=black!40, line width=1pt,
    fill=orange!30,
    inner xsep=6pt, inner ysep=-8pt
  },
  subbox2/.style={
    rectangle, rounded corners, 
    draw=black!40, line width=1pt,
    fill=yellow!40,
    inner xsep=6pt, inner ysep=-8pt
  },
  subbox3/.style={
    rectangle, rounded corners, 
    draw=black!40, line width=1pt,
    fill=green!20,
    inner xsep=6pt, inner ysep=-8pt
  }
}

\usepackage[textsize=tiny]{todonotes}

\begin{document}

\title{Divide-and-Conquer CoT: RL for Reducing Latency \\ via Parallel Reasoning}

\author{Arvind Mahankali \\
Stanford University \\
\texttt{amahanka@stanford.edu}
\and
Kaiyue Wen \\
Stanford University \\
\texttt{kaiyuew@stanford.edu}
\and
Tengyu Ma \\
Stanford University \\
\texttt{tengyuma@stanford.edu}
}
\date{}

\maketitle
\begin{abstract}
Long chain-of-thought reasoning (Long CoT) is now fundamental to state-of-the-art LLMs, especially in mathematical reasoning. However, LLM generation is highly sequential, and long CoTs lead to a high latency. We propose to train Divide-and-Conquer CoT (DC-CoT) to reduce the latency. With DC-CoT, the model can act as a director that identifies distinct subtasks  that can be performed in parallel in its reasoning process, and then spawns workers to execute the subtasks. Our goal is to achieve high accuracy, with a low longest path length, which is a theoretical measure of the latency needed for the response. We start with a long CoT base model (DeepScaleR-1.5B-Preview), and first use SFT with a small curated demonstration set to initialize its ability to spawn workers in a certain format. Because SFT degrades the accuracy significantly, we design a multi-stage RL algorithm, with various data filtering strategies, to recover the accuracy while decreasing the longest path length. Across several benchmarks including AIME 2024 and HMMT 2025, DC-CoT achieves similar accuracy as DeepScaleR-1.5B-Preview while decreasing longest path length by 35-40\%. \footnote{Our code, SFT dataset and models are publicly available at \url{https://github.com/amahankali10/DC_CoT_RL_for_Low_Latency_CoT_with_Parallel_Reasoning}.}
\end{abstract}

\section{Introduction} \label{sec:intro}

Long chain-of-thoughts (CoTs) have led to a paradigm shift in language model (LM) reasoning \citep{openai2024openaio1card, deepseekai2025deepseekr1incentivizingreasoningcapability}. Long CoTs allow LMs to think for longer and try many useful problem solving strategies, such as checking their own work and enumerating cases. Thus, long CoT models trained with reinforcement learning (RL) excel in mathematical reasoning, drastically improving performance on benchmarks such as AIME \citep{openaiO3systemcard}. 

However, long CoTs are quite slow, due to the sequential nature of LM generation \citep{leviathan2023fastinferencetransformersspeculative} and the large number of tokens involved. Over the course of 2025, token budgets in long CoTs have increased from 32K for DeepSeek R1 \citep{deepseekai2025deepseekr1incentivizingreasoningcapability} to 80K for Minimax M1 \citep{minimax2025minimaxm1scalingtesttimecompute} and even 256K for Kimi K2 Thinking \citep{moonshotai_kimik2_thinking}. Reducing the latency of generation using reasoning models is therefore a pressing question.

While long CoTs are all \textit{sequential}, many tasks admit a \textit{parallelizable} reasoning structure. For instance, many math problems naturally decompose into independent sub-cases that are solvable in parallel. Alternatively, there may be multiple unique plausible approaches that are worth trying in parallel as opposed to sequentially. %

This paper aims to significantly reduce the latency of long CoTs by \textbf{training} models to perform parallel thinking. We hypothesize that prompting-based scaffolding methods \textbf{alone}—whether heuristic approaches~\citep{anthropic2025multiagentsystem,Yan2025DontBuildMultiAgents,Lin2026ScalingLongRunningAutonomousCoding} or principled prompt optimization~\citep{agrawal2025gepareflectivepromptevolution}, which do not change model weights—are fundamentally limited in achieving effective or optimal parallel thinking. This limitation arises from existing models’ inability to decompose challenging goals into parallelizable sub-tasks, as they were never trained to do so. We posit that \pt is an advanced capability that must be explicitly taught through targeted, large-scale training.

\begin{figure}[h]
\centering
\resizebox{0.7\textwidth}{!}{\newlength{\vsep}
\setlength{\vsep}{0.8cm}

\begin{tikzpicture}[
    node distance=\vsep,
    box/.style={rectangle, draw, thick, minimum width=3.5cm, minimum height=1.5cm, align=left, font=\normalsize, inner xsep=8pt, inner ysep=6pt},
    worker/.style={rectangle, draw, thick, minimum width=4.1cm, text width=3.8cm, minimum height=1.2cm, align=left, font=\normalsize, fill=blue!10, inner xsep=8pt, inner ysep=6pt},
    bigbox/.style={rectangle, draw, thick, rounded corners=10pt, inner sep=10pt},
    arrow/.style={-Stealth, thick},
    dirwide/.style={box, minimum width=7.5cm, text width=7.0cm, minimum height=2.2cm, align=left},
    dirL/.style={box, minimum width=3.5cm, text width=3.1cm, minimum height=1.5cm, align=left},
    dirR/.style={box, minimum width=3.5cm, text width=3.1cm, minimum height=1.5cm, align=left}
]

\node[dirwide] (P) at (0,0) {
    \textbf{Problem}\\[0.2cm]
    Find the sum of all real roots $x$ of the equation:
    \[
    \left(2^{x} - 4\right)^{3} + \left(4^{x} - 2\right)^{3} = \left(4^{x} + 2^{x} - 6\right)^{3}
    \]
};

\node[dirwide, below=of P, fill=red!20] (D1) {
    \textbf{Director: Initial Phase}\\[0.15cm]
    \textbf{$<$think$>$}\\
    \textit{[Thinking process that analyzes the problem, and eventually decomposes the problem into subtasks]}\\
    Worker 1 will explore...\\
    Worker 2 will explore...\\
    Worker 3 will explore...\\
    \textbf{$<$spawn\_workers$>$}
};

\node[worker, right=1.5cm of D1, yshift=1.5cm] (W1) {
    \textbf{Worker 1}\\[0.2cm]
    \textbf{$<$worker\_1$>$}\\
    \textit{[reasoning process that solves the assigned task]}\\
    \textbf{$<$/worker\_1$>$}
};

\node[worker, below=of W1] (W2) {
    \textbf{Worker 2}\\[0.2cm]
    \textbf{$<$worker\_2$>$}\\
    \textit{[reasoning process that solves the assigned task]}\\
    \textbf{$<$/worker\_2$>$}
};

\node[worker, below=of W2] (W3) {
    \textbf{Worker 3}\\[0.2cm]
    \textbf{$<$worker\_3$>$}\\
    \textit{[reasoning process that solves the assigned task]}\\
    \textbf{$<$/worker\_3$>$}
};

\node[dirwide, below=\vsep of D1, fill=red!20] (D2) {
    \textbf{Director: Read Workers' Thoughts}\\[0.15cm]
    \textbf{$<$/spawn\_workers$>$}\\
    \textit{[Read workers' thoughts and continue reasoning]}
};

\node[dirR, minimum height=3.6504cm, anchor=north east, fill=red!20] (R) at ($(D2.south east)+(0,-\vsep)$) {
    \textbf{Director: Respawn}\\[0.15cm]
    \textit{[Identify subtasks]}\\
    Worker 1 will...\\
    Worker 2 will...\\
    Worker 3 will...\\
    \textbf{$<$spawn\_workers$>$}
};

\node[dirL, minimum height=3.6504cm, anchor=north west, fill=green!20] (L) at ($(D2.south west)+(0,-\vsep)$) {
    \textbf{Director: Answer}\\[0.15cm]
    \textbf{$<$/think$>$}\\
    \textbf{$<$answer$>$}\\
    \textit{[final answer text]}\\
    \textbf{$<$/answer$>$}
};

\node[bigbox, fit=(W1)(W2)(W3)] (W) {};

\draw[arrow] (P) -- (D1);
\draw[arrow] (D1.east) -- (W.west);
\draw[arrow] (W.west) -- (D2.east);

\draw[arrow] (D2.south) -- (L.north);
\draw[arrow] (D2.south) -- (R.north);

\draw[arrow] (R.east) to[out=0, in=-90] (W.south);

\end{tikzpicture}}
\caption{Our parallel inference procedure for \textbf{\ourmethod}. The model performs some reasoning before assigning tasks to workers (\textcolor{red}{Director --- Initial Phase}). Then, the workers perform the assigned tasks (\textcolor{blue}{Worker 1}, \textcolor{blue}{Worker 2}, \textcolor{blue}{Worker 3}). The director then reads the workers' outputs (\textcolor{red}{Director --- Read Workers' Thoughts}). Finally, the director may either finalize its answer (\textcolor{green}{Director --- Final Answer}) or launch another round of parallel generation (\textcolor{red}{Director --- Re-spawn Workers}).}
\label{fig:intro_flowchart}
\end{figure}
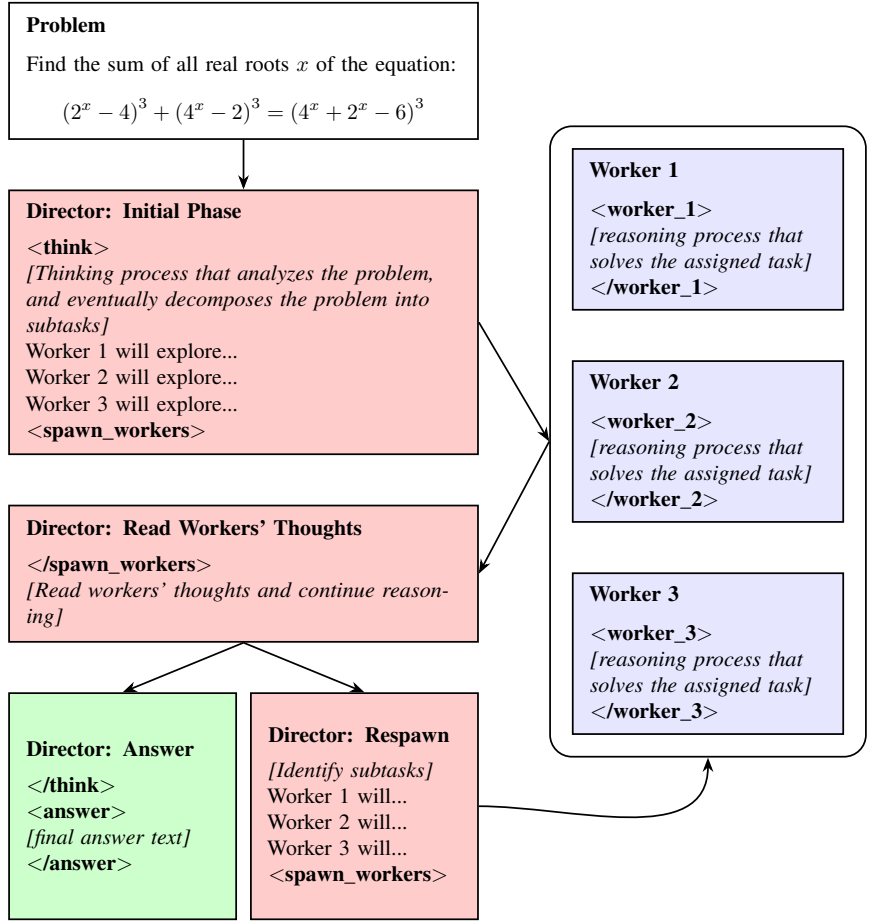

We introduce Divide-and-Conquer CoT (\ourmethod), a method to teach LMs to identify opportunities for parallelizable subtasks in their thinking processes,  solve the subtasks in parallel threads, then aggregate the outputs of these threads. As shown in \Cref{fig:intro_flowchart}, a single model will serve two roles: director and worker. The director performs sequential thinking until it identifies useful subtasks that can be performed in parallel. It will then spawn multiple workers, and specify a subtask for each worker. Each worker inherits its context from the director, and the workers complete their subtasks in parallel. Afterwards, the thinking processes of the workers will be appended to the context of the director. The director aggregates the workers' results, then thinks more before completing the response. This enables sequential reasoning for inherently sequential aspects, while allowing parallel reasoning when applicable to reduce latency.

\begin{figure}[t]
    \centering
    \includegraphics[width=0.6\textwidth]{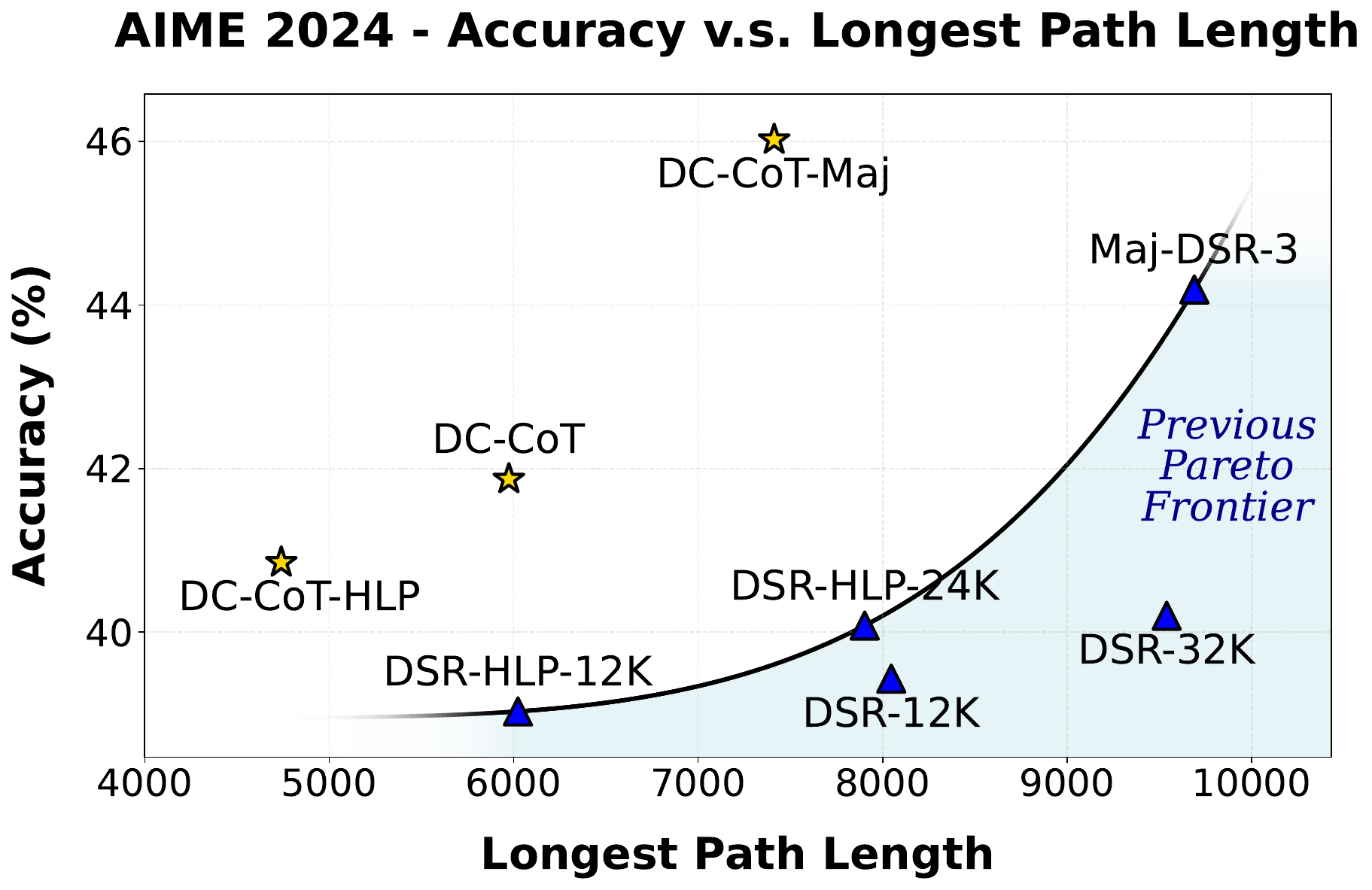}
    \caption{Accuracy (pass@1) and longest path length for \ourmethod and baselines on AIME 2024. \textbf{DSR-32K} and \textbf{DSR-12K} denote DeepScaleR-1.5B-Preview evaluated with response lengths $32,768$ and $12,000$ respectively. \textbf{\ourmethod-HLP} denotes \ourmethod trained with additional RL with a high length penalty, and \textbf{DSR-HLP-24K} and \textbf{DSR-HLP-12K} denotes baselines trained similarly, starting from DeepScaleR-1.5B-Preview. \textbf{Maj-DSR-3} denotes \textbf{DSR-12K} evaluated using majority voting (maj@3), and \textbf{\ourmethod-Maj} denotes \textbf{\ourmethod} evaluated using maj@3. \textbf{\ourmethod} achieves higher accuracy than \textbf{DSR-32K} while requiring less longest path length.}
    \label{fig:aime_2d_plot}
\end{figure}

We train our model to accurately identify parallelism and generate responses in this fashion. Starting from a sequential reasoning model (in our case DeepScaleR-1.5B-Preview \citep{deepscaler2025}), we first perform SFT on synthetic demonstrations (\Cref{subsec:sft_dataset} and \Cref{subsec:sft_train_seq_position_id_attn_mask}) to teach the model to follow the parallel format of \Cref{fig:intro_flowchart}. However, while SFT gives our model the ability to follow the format to a certain degree, it cannot fundamentally improve the \pt capability of our model --- in fact, it significantly degrades the accuracy compared to the base model.

Therefore, we perform RL, which is critical to improving the model's \pt capability and recovering the accuracy. We quantify the model's \pt capability by its accuracy and the average \textbf{longest path length} of its responses, defined as the number of tokens on the longest path from the beginning to the end of the response (see \Cref{sec:inference} for details). This is a theoretical proxy for the \textit{latency} of the responses. We design a multi-stage RL algorithm (\Cref{subsec:rl_algorithm_explanation}), which uses combinations of correctness and longest path length as reward functions to maximize accuracy and minimize longest path length.

We highlight that improving our model's parallel thinking capability through RL is nontrivial, as we face the following technical challenges. First, we initially use DAPO \citep{yu2025dapoopensourcellmreinforcement}, but find that the accuracy plateaus at a level much lower than the accuracy of our sequential base model. We believe this is caused by DAPO's tendency to steadily increase the entropy of the policy for almost all tokens. We speculate that an excessively high entropy for all tokens is detrimental, since some tokens should be essentially deterministic under an ideal policy. We later find that continuing RL training with \textbf{CISPO} \citep{minimax2025minimaxm1scalingtesttimecompute} improves the accuracy monotonically while keeping the entropy stable (unlike DAPO or GRPO \citep{shao2024deepseekmathpushinglimitsmathematical}), though using CISPO by itself does not suffice to recover the accuracy to the level of the sequential base model.

We next find that the data filtering strategy affects whether the model prioritizes accuracy or longest path length. Initially, we filter out problems where all the responses are wrong, while still including problems where all the responses are correct, as such problems can still provide a signal for the model to reduce its longest path length, via the longest path length penalty term in the reward function (see \Cref{subsec:rl_algorithm_explanation}). However, the model eventually plateaus in accuracy and starts to solely improve its longest path length. We speculate that there is a conflict between easier problems, for which the advantages in RL are dominated by the longest path length, and more difficult problems, where the advantage depends more on accuracy --- since we include problems where all responses are correct, the former dominates. We address this issue by updating our data filtering strategy --- partway through the training, we start filtering out those easy problems where all responses are correct.

Concurrent and independent works propose AsyncThink \citep{chi2025eraagenticorganizationlearning} and Native Parallel Reasoner \citep{wu2025nativeparallelreasonerreasoning}, or NPR, which both also train models end-to-end with RL to parallelize their reasoning. %
Unlike these works that use non-thinking variants of Qwen3-4B, we start from a long CoT reasoning model that is already trained with RL. We speculate that the long CoTs cause some of the technical challenges we describe above, due to the positive correlation between length and accuracy (and the resulting conflict between improving accuracy and reducing length).  We show that it is possible to build on a long CoT model, recovering and continuously improving the accuracy while reducing the inference latency. Interestingly, the very recent Kimi K2.5~\citep{kimi_k_2_5_agent_swarm} also uses a multi-stage strategy to train an orchestrator to decompose and delegate subtasks, prioritizing parallelization initially and then accuracy later, reminiscent of the purpose of our multi-stage data filtering strategy. Unlike Kimi K2.5, we allow the workers/sub-agents to learn from RL as well.

Our results on AIME 2024 \citep{maa_aime2024} are shown in \Cref{fig:aime_2d_plot}. Our method, \ourmethod, achieves a better accuracy and longest path length compared to DeepScaleR-1.5B-Preview evaluated at response lengths 32K and 12K (denoted as \textbf{DSR-32K} and \textbf{DSR-12K} respectively) --- in particular, \ourmethod achieves a $37.4\%$ reduction in longest path length compared to \textbf{DSR-32K}. Additionally, continuing RL from DC-CoT with a high length penalty (denoted as \textbf{\ourmethod-HLP}) gives a Pareto improvement compared to similarly-trained baselines trained from DeepScaleR-1.5B-Preview (\textbf{DSR-HLP-24K} and \textbf{DSR-HLP-12K}). Finally, \ourmethod is complementary to test-time scaling through independent sampling, achieving further gains with majority voting. \ourmethod with majority voting (denoted as \textbf{\ourmethod-Maj}) achieves better accuracy and longest path length compared to DeepScaleR-1.5B-Preview with response length 12K and majority voting (denoted as \textbf{Maj-DSR-3}). We observe similar benefits when using \ourmethod on other benchmarks (see \Cref{fig:all_2d_plots}, \Cref{tab:main_results}, and \Cref{tab:change_compared_to_baselines}). In \Cref{appendix:latency_table}, we also show that DC-CoT improves over DSR-32K and DSR-12K in latency, measured in wall-clock time.

\section{Related Work} \label{sec:related_work}

We discuss works on reinforcement learning with verifiable rewards (RLVR) and long CoT for mathematical reasoning in \Cref{appendix:additional_related_work}.

\paragraph{Delegation in Multi-agent Systems.} Several works study multi-agent systems \citep{anthropic2025multiagentsystem, kim2025sciencescalingagentsystems, zou2025latentcollaborationmultiagentsystems, gao2025singleagentmultiagentsystemsboth}, where different LMs collaborate to solve problems, dividing an over-arching task into subtasks. \citet{anthropic2025multiagentsystem} find that multi-agent systems can lead to large improvements in deep research compared to sequential single-agent systems. Other works \citep{nguyen2025sfrdeepresearcheffectivereinforcementlearning, gao2025singleagentmultiagentsystemsboth, kim2025sciencescalingagentsystems} study the circumstances under which multi-agent systems will outperform single-agent systems, and vice versa. Training-free methods using multi-agent systems have been proposed to improve efficiency \citep{narayan2025minionscostefficientcollaborationondevice, zou2025latentcollaborationmultiagentsystems} and accuracy \citep{du2023improvingfactualityreasoninglanguage} compared to single-agent systems. Some recent works~\citep{soylu2024finetuningpromptoptimizationgreat,ziems2025multimodulegrpocomposingpolicy} explore optimizing the weights of each agent in multi-agent systems given a fixed workflow. Other works train multi-agent systems with RL \citep{dang2025multiagentcollaborationevolvingorchestration, zhou2025mem1learningsynergizememory, sun2025scalinglonghorizonllmagent}. \citet{dang2025multiagentcollaborationevolvingorchestration} train an orchestrator with RL to call LLM sub-agents while minimizing total resource usage. \citet{sun2025scalinglonghorizonllmagent} train a coordinator and its sub-agents end-to-end to manage the context length of the coordinator, while \citet{zhou2025mem1learningsynergizememory} train a single agent with RL to compress its own context length. \citet{hu2025owloptimizedworkforcelearning} train a multi-agent system with a planner and workers to accurately perform tasks such as coding and deep research. Our goal is to reduce the inference latency, which \citet{dang2025multiagentcollaborationevolvingorchestration}, \citet{sun2025scalinglonghorizonllmagent}, \citet{zhou2025mem1learningsynergizememory} and \citet{hu2025owloptimizedworkforcelearning} do not study --- additionally, we train the sub-agents through RL, unlike \citet{dang2025multiagentcollaborationevolvingorchestration}.

\paragraph{Parallel Sampling for Inference-Time Scaling.} Several works study how to improve the accuracy of LMs, via \textit{parallel sampling} \citep{wang2023selfconsistencyimproveschainthought,yao2023treethoughtsdeliberateproblem, brown2024largelanguagemonkeysscaling, snell2024scalingllmtesttimecompute, zhang2024accessinggpt4levelmathematical, wu2025modeconditioningunlockssuperiortesttime, wen2025parathinkernativeparallelthinking, qi2025learningreasonparallelsamples, zhao2025majorityrightrltraining, hassid2025dontoverthinkitpreferring} or via parallel sampling followed by iterative summarization and refinement~\citep{madaan2025rethinkingthinkingtokensllms, pacore2025} (more details in \Cref{appendix:additional_related_work}). Our work is orthogonal and complementary to theirs because we aim to identify \textit{distinct} and parallelizable subtasks, which cannot be achieved by independent sampling. Indeed, parallel sampling can be combined with our approach to achieve additional gains, as shown in Section~\ref{subsec:results}.

\paragraph{Parallel Decoding with LLMs to Reduce Latency.} Improving the latency of LM inference is an active area of research \citep{leviathan2023fastinferencetransformersspeculative, cai2024medusasimplellminference, chen2023acceleratinglargelanguagemodel, zhang2024draftandverify}. Several works specifically study how an LM can identify opportunities for parallelization in its own generation, through prompting \citep{ning2024skeletonofthoughtpromptingllmsefficient}, training on demonstrations \citep{liu2024aparllmsautoparallelautoregressive}, and preference optimization using solution quality and latency to compute a reward \citep{jin2025learningpromisescalinglanguage}. These works study instruction following models. Parallelization of inference has also been studied for reasoning models that use long CoTs \citep{biju2025sprintenablinginterleavedplanning, yang2025multiverselanguagemodelssecretly, zheng2025parallelr1parallelthinkingreinforcement, pan2025learningadaptiveparallelreasoning}. \citet{biju2025sprintenablinginterleavedplanning} and \citet{yang2025multiverselanguagemodelssecretly} curate demonstrations, used for SFT, which can teach reasoning models to parallelize their reasoning traces. \citet{zheng2025parallelr1parallelthinkingreinforcement} train a reasoning model with RL to perform parallel inference, as part of a curriculum used to improve sequential reasoning. \citet{pan2025learningadaptiveparallelreasoning} show that in the Countdown task, LMs trained with RL to search in parallel achieve a better accuracy-latency tradeoff than sequential methods.

\section{Parallel Inference Architecture} \label{sec:inference}

We now explain our architecture for parallelizable inference. In the following, we use $\EOS$ to denote the end-of-sequence token. We define a single-thread inference primitive $\infer(\model, P, \termination, L)$, which performs standard LM inference. Here $\model$ is a language model, $P$ is a prompt, $\termination$ is a termination string, and $L$ is an upper bound on the amount of generated tokens. The subroutine $\infer(\model, P, \termination, L)$ performs generation using the model $\model$ and prompt $P$ until one of the following occurs: (1) $L$ tokens are generated, (2) $\termination$ is output, or (3) the $\EOS$ token is output. The output is a list of tokens $S$, whose length is denoted by $|S|$.

Our architecture is implemented entirely using calls to $\infer$. Thus, we directly use vLLM \citep{kwon2023efficient}, without any under-the-hood modifications, unlike Multiverse \citep{yang2025multiverselanguagemodelssecretly}, which uses non-standard attention masks during generation, thus requiring a customized version of SGLang. We adjust our training-time algorithm to maintain consistency with inference (see \Cref{subsec:sft_train_seq_position_id_attn_mask}).

An illustration of our architecture is shown in \Cref{fig:intro_flowchart}. A more detailed version with examples of thoughts generated by the model is shown in \Cref{fig:inference_algorithm_example} in \Cref{appendix:detailed_inference_figure}. Formal pseudocode is given in \Cref{algorithm:inference_algorithm} in \Cref{appendix:inference_pseudocode}.

\paragraph{Initial Single-Thread Phase.} We begin by using $\infer(\model, \prompt, \Wstart, L)$ to perform inference in single-thread mode. Here the model performs initial reasoning, through which it may identify subtasks that are useful towards solving the problem and can be done in parallel. This corresponds to the ``Director: Initial Phase'' box in \Cref{fig:intro_flowchart}, and in the rest of the paper, we use $\singlethreadone$ to refer to this part. This call to $\infer$ terminates when $\Wstart$ is output --- in essentially all cases, our model will output $\Wstart$ in this stage rather than $\EOS$.

\paragraph{Parallel Generation with Workers.} After reaching $\Wstart$, our algorithm spawns multiple workers, which will each continue generation \textit{in parallel} based on the contents of $\singlethreadone$. In our experiments, we use $3$ workers. For each worker $i \in [3]$, the prompt of the $i^{\textup{th}}$ worker will be $\prompt_i := \prompt + \singlethreadone + \Wstarti{i}$. In other words, each worker sees the original prompt and the result of the initial single-thread generation, as well as a special tag to distinguish this worker from other workers. The $i^{th}$ worker then performs standard inference by calling $\infer(\model, \prompt_i, \Wendi{i}, L)$ --- these $3$ calls to $\infer$ are executed in parallel. In the rest of the paper, we use $\wonetext$ to refer to $\Wstarti{1}$, together with the portion generated by $\infer$ --- this is the text corresponding to ``Worker 1'' in \Cref{fig:intro_flowchart}. We define $\wtwotext$ and $\wthreetext$ analogously.

\paragraph{Returning to Single-Thread.} After all of the workers finish their generation, the model (in its role as director) sees the work done by the workers, and performs further reasoning (see the ``Director: Read Workers' Thoughts'' box in \Cref{fig:intro_flowchart}). The model may finalize its answer and end generation (see ``Director: Answer'' box) or may identify new subtasks that can be done in parallel, output $\Wstart$, and start a new round of multi-thread generation (see ``Director: Respawn'' box). While there can be arbitrarily many rounds of parallel generation, our model typically spawns workers for 1 or sometimes 2 rounds. In the case where only $1$ round occurs, we use $\singlethreadtwo$ to refer to all the text after $\Wend$ inclusive, corresponding to boxes ``Director: Read Workers' Thoughts'' and ``Director: Final Answer'' in \Cref{fig:intro_flowchart}.

\paragraph{Longest Path Length.} We use the longest path length of a response, defined below, as a proxy for the wall-clock time for generating it. Suppose for simplicity that workers are only spawned for a single round --- the definition generalizes naturally to multiple rounds. Let $\singlethreadone$, $\wonetext$, $\wtwotext$, $\wthreetext$ and $\singlethreadtwo$ be as defined earlier in this section. Then, we define the longest path length of $\response$ as $\plen(\response) := |S_1| + \max_{i \in [3]} |W_i| + |S_2|$. We enforce a longest path length budget during inference by setting the allowed output length at each call to $\infer$.

\section{Training Methods} \label{sec:training_methods}

\newcommand{\trainflowchartfigref}{\Cref{fig:training_algorithm_flowchart}}
\begin{figure}[h]
    \centering
    \begin{tikzpicture}
\def\mainwidth{15.6cm}
\def\subboxwidth{14.73cm}
\def\innertextwidth{\dimexpr\subboxwidth-12pt\relax}
\newcommand{\flowItemizeStart}{%
  \begin{itemize}
    \setlength{\itemsep}{1pt}%
    \setlength{\topsep}{0pt}%
    \setlength{\partopsep}{0pt}%
    \setlength{\parskip}{0pt}%
    \setlength{\parsep}{0pt}%
}
\newcommand{\flowItemizeEnd}{\end{itemize}}
\node (section1content) [text width=\innertextwidth, align=left, minimum height=2.0cm] {%
  {\bfseries 1. Prepare SFT Dataset (\Cref{subsec:sft_dataset})}
  \vspace{-3pt}
  \flowItemizeStart
    \item Sample $1$ correct solution with sequential CoT for each given problem using $\basemodel$
    \item Use a more powerful model $\rewritemodel$ to rewrite CoTs into the parallel format as in \Cref{fig:intro_flowchart}, obtaining dataset $\parallelrewritedata$  
  \flowItemizeEnd
};
\begin{pgfonlayer}{back}
\node (section1box) [subbox1, fit=(section1content), minimum width=\subboxwidth, minimum height=2.0cm] {};
\end{pgfonlayer}
\node (section2content) [text width=\innertextwidth, align=left, below=0.2cm of section1box, minimum height=2.0cm] {%
  {\bfseries 2. SFT (\Cref{subsec:sft_train_seq_position_id_attn_mask})}
  \vspace{-3pt}
  \flowItemizeStart
    \item SFT starting from $\basemodel$ on $\parallelrewritedata$ using attention masks which respect the parallelism shown in \Cref{fig:intro_flowchart}
  \flowItemizeEnd
};
\begin{pgfonlayer}{back}
\node (section2box) [subbox2, fit=(section2content), minimum width=\subboxwidth, minimum height=2.0cm] {};
\end{pgfonlayer}
\node (section3content) [text width=\innertextwidth, align=left, below=0.2cm of section2box, minimum height=2.0cm] {%
  {\bfseries 3. RL (\Cref{subsec:rl_algorithm_explanation})}
  \vspace{-3pt}
  \flowItemizeStart
    \item RL with $4$ stages using correctness reward and format/longest path length penalties
    \item Each stage has varying hyperparameters and data filtering criteria  
  \flowItemizeEnd
};
\begin{pgfonlayer}{back}
\node (section3box) [subbox3, fit=(section3content), minimum width=\subboxwidth, minimum height=2.0cm] {};
\end{pgfonlayer}
\coordinate (topcenter) at ($(section1box.north west)!0.5!(section1box.north east)$);
\node (overalltitlefit) [font=\bfseries\flowTitleFont, text opacity=0, text width=\mainwidth, align=center]
  at ($(topcenter)+(0,0.24cm)$) {\trainflowchartfigref: Training Procedure};
\begin{pgfonlayer}{farback}
\node (outercontainer) [outerbox, fit=(overalltitlefit)(section1box)(section2box)(section3box)] {};
\end{pgfonlayer}
\node (overalltitle) [font=\bfseries\flowTitleFont, anchor=north, yshift=-4pt, text width=\mainwidth, align=center]
  at (outercontainer.north) {\trainflowchartfigref: Training Procedure};
\end{tikzpicture}
    \phantomcaption
    \label{fig:training_algorithm_flowchart}
\end{figure}
\noindent This section describes our training methods, which are summarized in \Cref{fig:training_algorithm_flowchart}.

\subsection{Supervised Fine-tuning (SFT)} \label{subsec:sft_for_spawning_workers}

\subsubsection{SFT Dataset} \label{subsec:sft_dataset}

In this section, we elaborate on step 1 of \Cref{fig:training_algorithm_flowchart}.

\paragraph{Initial Generation of Sequential Long CoT Rollouts.} We generate sequential CoTs using $\basemodel$, which can be any reasoning model. For an initial set of problems $\initialproblemset$, we perform rejection sampling, generating up to $3$ responses per problem, until we obtain a correct response. If a problem $\problem$ has at least one correct response $\responseseq$, we include $(\problem, \responseseq)$ in $\initrolloutdata$ --- otherwise, we omit $\problem$.

\paragraph{Rewriting Sequential CoTs to Include Parallelization.} Given the $(\problem, \responseseq)$ pairs in $\initrolloutdata$, we rewrite the responses $\responseseq$ to spawn workers where it would be useful, obtaining a dataset $\parallelrewritedata$ of responses $(\problem, \response)$ where the response $\response$ performs parallel thinking.

Given a sequential CoT response $\responseseq$, we tell our rewriting model $\rewritemodel$ to largely preserve the content and thinking process of $\responseseq$, while producing $\response$. The response $\response$ should satisfy the following criteria. (A) It should begin by reasoning in a sequential manner, in the same way as $\responseseq$ does --- this is shown in the ``Director: Initial Phase'' box in \Cref{fig:intro_flowchart}. (B) The reasoning processes of the workers in $\response$ should be excerpted almost verbatim from $\responseseq$. Intuitively, criteria (A) and (B) mean that the solutions in $\parallelrewritedata$ have a similar structure to responses produced by $\basemodel$. We hypothesize that training $\basemodel$ on rewritten responses satisfying (A) and (B) mitigates the degradation in the accuracy of $\basemodel$ caused by SFT, though not eliminating it entirely.

We also ask $\rewritemodel$ to include in $\response$, prior to spawning workers, a justification for why the tasks it has assigned to the workers can be performed in parallel. One goal is to reduce cases where the tasks are not in fact parallelizable, e.g. where the task assigned to Worker 2 requires the answer to Worker 1's task. This also allows for the possibility that our model $\sftmodel$ trained on $\parallelrewritedata$ can self-reflect, and after proposing an initial assignment of tasks to workers, it can decide that the assignment is suboptimal, rather than committing to spawning workers.

To convey the above information to $\rewritemodel$, we give $\rewritemodel$ a detailed prompt, shown in \Cref{box:claude_parallel_rewrite_prompt} in \Cref{appendix:sft_data_gen_prompts}. The prompt contains an example of a sequential CoT $\responseseq$, together with an example of a rewritten response $\response$ that spawns workers.

We again use rejection sampling when generating the rewritten responses with $\rewritemodel$. For each pair $(\problem, \responseseq)$ in $\initrolloutdata$, we attempt up to $3$ times to generate a rewritten response $\response$ using $\rewritemodel$. We verify that $\response$ has the correct answer. Additionally, we verify that $\response$ satisfies the format shown in \Cref{fig:intro_flowchart}. We consider $\response$ to satisfy the format if it spawns workers, if it includes all relevant tags shown in \Cref{fig:intro_flowchart}, and if these tags are nested and matched appropriately.

\subsubsection{SFT Training Sequences, Position IDs, and Attention Mask} \label{subsec:sft_train_seq_position_id_attn_mask}

We now elaborate on step 2 of \Cref{fig:training_algorithm_flowchart}. For simplicity, we consider a response where workers are only called for 1 round, and define $S_1$, $W_1$, $W_2$, $W_3$ and $S_2$ as in \Cref{sec:inference}.

In our inference procedure, two distinct operations take place. (A) When $W_1$, $W_2$ and $W_3$ are initially generated, each of them has $S_1$ as the prefix. (B) $W_1$, $W_2$ and $W_3$ are later included in the prefix used to generate $S_2$.

For $i \in [3]$, let $\WA{i}$ denote the KV cache vectors corresponding to $W_i$ in operation (A), and let $\WB{i}$ denote the KV cache vectors corresponding to $W_i$ in step (B). Then, $\WA{i}$ and $\WB{i}$ are distinct vectors, for the following reason. When the $W_i$ are initially generated during operation (A), they are generated in parallel --- they share the original prompt and $S_1$ as their context, and are otherwise separate LM generation tasks. Thus, the $W_i$ only attend to $S_1$ and not each other. Meanwhile, in operation (B), we perform standard LM generation to generate $S_2$, using $S_1$, $W_1$, $W_2$ and $W_3$ as the context. Thus, during the prefill stage, a standard causal mask will be applied to the context, meaning $W_2$ attends to $W_1$, and $W_3$ attends to $W_1$ and $W_2$. 

\begin{figure}[h]
    \centering
    \resizebox{0.4\textwidth}{!}{\begin{tikzpicture}[x=1cm,y=1cm]

    \pgfmathsetmacro{\s}{\attentioncellcm}
    \pgfmathsetmacro{\m}{\attentiongencellcm}
    \pgfmathsetmacro{\w}{\attentionwidecellcm}

    \pgfmathsetmacro{\totalw}{2*\s + 3*\m + \w}
    \pgfmathsetmacro{\totalh}{2*\s + 3*\m + \w}

    \pgfmathsetmacro{\tickgap}{\attentiontickgap}
    \pgfmathsetmacro{\axisgap}{\attentionaxisgap}

    \coordinate (gridNW) at (0,0);
    \coordinate (gridNE) at (\totalw,0);
    \coordinate (gridSW) at (0,-\totalh);
    \coordinate (gridSE) at (\totalw,-\totalh);

    \pgfmathsetmacro{\xposZero}{0.5*\s}
    \pgfmathsetmacro{\xposOne}{\s + 0.5*\m}
    \pgfmathsetmacro{\xposTwo}{\s + 1.5*\m}
    \pgfmathsetmacro{\xposThree}{\s + 2.5*\m}
    \pgfmathsetmacro{\xposFour}{\s + 3*\m + 0.5*\w}
    \pgfmathsetmacro{\xposFive}{\s + 3*\m + \w + 0.5*\s}

    \pgfmathsetmacro{\yposZero}{-0.5*\s}
    \pgfmathsetmacro{\yposOne}{-(\s + 0.5*\m)}
    \pgfmathsetmacro{\yposTwo}{-(\s + 1.5*\m)}
    \pgfmathsetmacro{\yposThree}{-(\s + 2.5*\m)}
    \pgfmathsetmacro{\yposFour}{-(\s + 3*\m + 0.5*\w)}
    \pgfmathsetmacro{\yposFive}{-(\s + 3*\m + \w + 0.5*\s)}

    \newcommand{\colcenter}[1]{%
        \ifnum#1=0 \xposZero\else
        \ifnum#1=1 \xposOne\else
        \ifnum#1=2 \xposTwo\else
        \ifnum#1=3 \xposThree\else
        \ifnum#1=4 \xposFour\else
        \xposFive\fi\fi\fi\fi\fi
    }

    \newcommand{\rowcenter}[1]{%
        \ifnum#1=0 \yposZero\else
        \ifnum#1=1 \yposOne\else
        \ifnum#1=2 \yposTwo\else
        \ifnum#1=3 \yposThree\else
        \ifnum#1=4 \yposFour\else
        \yposFive\fi\fi\fi\fi\fi
    }

    \newcommand{\cellsize}[1]{%
        \ifnum#1=0 \s\else
        \ifnum#1=1 \m\else
        \ifnum#1=2 \m\else
        \ifnum#1=3 \m\else
        \ifnum#1=4 \w\else
        \s\fi\fi\fi\fi\fi
    }

    \foreach \x in {0,...,5} {
        \foreach \y in {0,...,5} {

            \pgfmathsetmacro{\xpos}{\colcenter{\x}}
            \pgfmathsetmacro{\ypos}{\rowcenter{\y}}
            \pgfmathsetmacro{\cellw}{\cellsize{\x}}
            \pgfmathsetmacro{\cellh}{\cellsize{\y}}
            \pgfmathsetmacro{\halfwidth}{0.5*\cellw}
            \pgfmathsetmacro{\halfheight}{0.5*\cellh}

            \pgfmathsetmacro{\isdiag}{0}
            \ifnum\y=\x \pgfmathsetmacro{\isdiag}{1} \fi

            \pgfmathsetmacro{\isgreen}{0}
            \ifnum\y=0 \ifnum\x=0 \pgfmathsetmacro{\isgreen}{1} \fi \fi
            \ifnum\y=1 \ifnum\x=0 \pgfmathsetmacro{\isgreen}{1} \fi \fi
            \ifnum\y=1 \ifnum\x=1 \pgfmathsetmacro{\isgreen}{1} \fi \fi
            \ifnum\y=2 \ifnum\x=0 \pgfmathsetmacro{\isgreen}{1} \fi \fi
            \ifnum\y=2 \ifnum\x=2 \pgfmathsetmacro{\isgreen}{1} \fi \fi
            \ifnum\y=3 \ifnum\x=0 \pgfmathsetmacro{\isgreen}{1} \fi \fi
            \ifnum\y=3 \ifnum\x=3 \pgfmathsetmacro{\isgreen}{1} \fi \fi
            \ifnum\y=4 \ifnum\x=0 \pgfmathsetmacro{\isgreen}{1} \fi \fi
            \ifnum\y=4 \ifnum\x=4 \pgfmathsetmacro{\isgreen}{1} \fi \fi
            \ifnum\y=5 \ifnum\x=0 \pgfmathsetmacro{\isgreen}{1} \fi \fi
            \ifnum\y=5 \ifnum\x=4 \pgfmathsetmacro{\isgreen}{1} \fi \fi
            \ifnum\y=5 \ifnum\x=5 \pgfmathsetmacro{\isgreen}{1} \fi \fi

            \pgfmathsetmacro{\isred}{0}
            \ifnum\y=2 \ifnum\x=1 \pgfmathsetmacro{\isred}{1} \fi \fi
            \ifnum\y=3 \ifnum\x=1 \pgfmathsetmacro{\isred}{1} \fi \fi
            \ifnum\y=3 \ifnum\x=2 \pgfmathsetmacro{\isred}{1} \fi \fi
            \ifnum\y=4 \ifnum\x=1 \pgfmathsetmacro{\isred}{1} \fi \fi
            \ifnum\y=4 \ifnum\x=2 \pgfmathsetmacro{\isred}{1} \fi \fi
            \ifnum\y=4 \ifnum\x=3 \pgfmathsetmacro{\isred}{1} \fi \fi
            \ifnum\y=5 \ifnum\x=1 \pgfmathsetmacro{\isred}{1} \fi \fi
            \ifnum\y=5 \ifnum\x=2 \pgfmathsetmacro{\isred}{1} \fi \fi
            \ifnum\y=5 \ifnum\x=3 \pgfmathsetmacro{\isred}{1} \fi \fi

            \ifdim\isdiag pt=1pt
                \fill[white]
                    (\xpos-\halfwidth, \ypos+\halfheight) --
                    (\xpos+\halfwidth, \ypos+\halfheight) --
                    (\xpos+\halfwidth, \ypos-\halfheight) -- cycle;

                \fill[green!30]
                    (\xpos-\halfwidth, \ypos+\halfheight) --
                    (\xpos-\halfwidth, \ypos-\halfheight) --
                    (\xpos+\halfwidth, \ypos-\halfheight) -- cycle;

                \draw[black, thick]
                    (\xpos-\halfwidth, \ypos-\halfheight) rectangle
                    (\xpos+\halfwidth, \ypos+\halfheight);
                \draw[black, thick, dotted]
                    (\xpos-\halfwidth, \ypos+\halfheight) --
                    (\xpos+\halfwidth, \ypos-\halfheight);
            \else
                \def\fillcol{white}
                \ifdim\isgreen pt=1pt \def\fillcol{green!30}\fi
                \ifdim\isred pt=1pt   \def\fillcol{red!30}\fi

                \fill[\fillcol]
                    (\xpos-\halfwidth, \ypos-\halfheight) rectangle
                    (\xpos+\halfwidth, \ypos+\halfheight);
                \draw[black, thick]
                    (\xpos-\halfwidth, \ypos-\halfheight) rectangle
                    (\xpos+\halfwidth, \ypos+\halfheight);
            \fi
        }
    }

    \pgfmathsetmacro{\ticky}{-(\totalh + \tickgap)}
    \node[anchor=base, font=\attentionfont] at (\xposZero,  \ticky) {$S_1$};
    \node[anchor=base, font=\attentionfont] at (\xposOne,   \ticky) {$\WAfig{1}$};
    \node[anchor=base, font=\attentionfont] at (\xposTwo,   \ticky) {$\WAfig{2}$};
    \node[anchor=base, font=\attentionfont] at (\xposThree, \ticky) {$\WAfig{3}$};
    \node[anchor=base, font=\attentionfont] at (\xposFour,  \ticky) {$\WBfig$};
    \node[anchor=base, font=\attentionfont] at (\xposFive,  \ticky) {$S_2$};

    \pgfmathsetmacro{\tickx}{-\tickgap}
    \node[anchor=center, font=\attentionfont, rotate=-90] at (\tickx, \yposZero)  {$S_1$};
    \node[anchor=center, font=\attentionfont, rotate=-90] at (\tickx, \yposOne)   {$\WAfig{1}$};
    \node[anchor=center, font=\attentionfont, rotate=-90] at (\tickx, \yposTwo)   {$\WAfig{2}$};
    \node[anchor=center, font=\attentionfont, rotate=-90] at (\tickx, \yposThree) {$\WAfig{3}$};
    \node[anchor=center, font=\attentionfont, rotate=-90] at (\tickx, \yposFour)  {$\WBfig$};
    \node[anchor=center, font=\attentionfont, rotate=-90] at (\tickx, \yposFive)  {$S_2$};

    \pgfmathsetmacro{\xmid}{0.5*\totalw}
    \pgfmathsetmacro{\ymid}{-0.5*\totalh}

    \pgfmathsetmacro{\axisfromtick}{\axisgap}

    \pgfmathsetmacro{\keyy}{\ticky - \axisfromtick}
    \node[anchor=base, font=\attentionaxisfont] at (\xmid, \keyy) {Key};

    \pgfmathsetmacro{\queryx}{\tickx - \axisfromtick}
    \node[anchor=base, font=\attentionaxisfont, rotate=-90] at (\queryx, \ymid) {Query};

\end{tikzpicture}}
    \caption{Attention mask at training time. Blocks which are included in the attention computation are shown in \textcolor{green}{green}, while blocks which are excluded are shown in \textcolor{red}{red}. The attention mask is also causal --- entries above the diagonal are masked out. Here, $\WBfig$ refers to the concatenation of $\WB{1}$, $\WB{2}$ and $\WB{3}$.}
    \label{fig:parallel_attn_mask}
\end{figure}

Thus, for the sake of consistency with inference-time behavior, we include $W_1$, $W_2$ and $W_3$ twice in the training sequences, with the first occurrence corresponding to operation (A) (i.e. the $\WA{i}$ for $i \in [3]$) and the second corresponding to (B) (i.e. the $\WB{i}$). To enforce the independence of the $\WA{i}$, as well as the fact that $S_2$ attends to the $\WB{i}$ but not the $\WA{i}$, we use the attention mask in \Cref{fig:parallel_attn_mask}. Similarly, at training time, we let the position IDs corresponding to the $\WA{i}$ all start from $|S_1|$, while those of $\WB{1}$ start from $|S_1|$, those of $\WB{2}$ start from $|S_1| + |W_1|$, and so on.

\subsection{Reinforcement Learning (RL)} \label{subsec:rl_algorithm_explanation}

We next perform RL starting from $\sftmodel$. We find that the model's accuracy degrades after performing SFT with $\parallelrewritedata$ --- thus, we aim to recover the accuracy. Additionally, many of $\sftmodel$'s responses do not spawn workers (for instance, on a batch of problems for our RL training dataset, $\sftmodel$ only spawns workers for roughly $65\%$ of responses). Thus, we would like to teach $\sftmodel$ to spawn workers more often, and in a way which reduces its longest path length. The reward for a response $\response$ is determined by the correctness and longest path length of $\response$, as well as whether $\response$ satisfies the format defined by \Cref{fig:intro_flowchart}.

Given a problem and response generated by $\currentpolicy$, we construct a training sequence, as well as the accompanying attention mask and position IDs, in a manner similar to \Cref{subsec:sft_train_seq_position_id_attn_mask}. We do not compute the loss on tokens which are not generated by the policy itself --- for instance, using the terminology of \Cref{subsec:sft_train_seq_position_id_attn_mask}, we compute the loss on the tokens in $\WA{i}$, but not on those in $\WB{i}$, for $i \in [3]$.

In our RL training, we use DAPO \citep{yu2025dapoopensourcellmreinforcement} (a variant of GRPO \citep{shao2024deepseekmathpushinglimitsmathematical}), and CISPO \citep{minimax2025minimaxm1scalingtesttimecompute} --- implementation details are deferred to \Cref{appendix:grpo_cispo}.

\paragraph{Penalty for overlong responses. }
We use an overlong length penalty similar to \citet{yu2025dapoopensourcellmreinforcement}. Here, we penalize responses with excessive longest path length. For a response $\response$, we add to the reward a length penalty $\lengthpenaltyfn(\response) := \lpcoeff \cdot \frac{\plen(\response) - \penaltycutoff}{\maxplen - \penaltycutoff}$ for some cutoff $\penaltycutoff$, if $\plen(\response) > \penaltycutoff$. We let $\lpcoeff = 0.1$ for most of our experiments.

\paragraph{Reward function.} Given a response $\response$, its reward $\rewardfn(\response)$ is defined as follows. We first define a \textit{correctness-format} reward as follows:
\[
\correctformatreward(\response) := 
\begin{cases}
1.0 & \text{if $\response$ is correct and satisfies format} \\
0.5 & \text{if $\response$ is correct but does not satisfy format} \\
0.0 & \text{otherwise}
\end{cases}
\]
Then, the overall reward is $\rewardfn(\response) := \max(0, \correctformatreward(\response) - \lengthpenaltyfn(\response))$.

\paragraph{Data filtering.} Following DAPO \citep{yu2025dapoopensourcellmreinforcement}, we filter problems $\problem$ according to certain criteria in each iteration of RL. We use two different data filtering strategies: (1) \textbf{\firstfilter}, where we filter out problems $\problem$ where there are no responses $\response$ that are both correct \textit{and} satisfy the format, i.e. where none of the responses $\response$ satisfy $f(\response) = 1.0$, and (2) \textbf{\secondfilter}, where we filter out problems $\problem$ where all of the responses are correct or all of the responses are wrong. One key difference is that \firstfilter includes problems where all of the responses are correct, unlike \secondfilter. Filtering strategy \secondfilter does not take format into account. We use \firstfilter in the initial stages of training. Many of the SFT model $\sftmodel$'s responses do not spawn workers at all, or do not satisfy the format. Thus, even problems $\problem$ for which all of the responses have the correct answer provide a useful signal for the model to learn to spawn workers, follow the format, and reduce its longest path length. However, \firstfilter eventually leads to a plateau in the accuracy, as the model prioritizes longest path length. Thus, we switch to \secondfilter partway through RL training, as discussed below in more detail.

\paragraph{Multi-stage training to address accuracy plateaus.} We make changes to our RL algorithm at multiple intermediate steps, each time with the goal of addressing accuracy plateaus. Due to compute constraints, each time we make a change to our training algorithm, we build on our latest checkpoint, rather than starting from scratch. Similar multi-stage training is done by other works, such as ProRL \citep{liu2025prorlprolongedreinforcementlearning}. Each time we make a change to our algorithm and begin a new stage, we reset the reference model $\refmodel$, as well as the optimizer state, similar to ProRL \citep{liu2025prorlprolongedreinforcementlearning}. Our RL training run has $4$ stages in total. In \Cref{appendix:fewer_stage_RL}, we study versions of our RL training run that have fewer stages.

In Stage 1, we use DAPO, and filtering strategy \firstfilter. We set the upper clipping parameter $\epshigh$ to $0.28$, as done by \citet{yu2025dapoopensourcellmreinforcement}. The maximum allowed longest path length $\maxplen$ for any response $\response$ is $7,500$ in this stage (and in stages 2 and 3, as well). The length penalty cutoff $\penaltycutoff$ is $2,000$. For many responses, the SFT model does not spawn workers at all, or does not follow the format. Over the course of Stage 1, the accuracy improves, and the model learns to follow the format and spawn workers more often, while the longest path length decreases. However, we eventually observe a plateau in the accuracy after many steps (\Cref{fig:dapo_stage_1_plateau} in \Cref{appendix:intermediate_plateaus_rl}). We hypothesize that this accuracy plateau is caused by DAPO's tendency to keep increasing the entropy of our model. In \Cref{fig:entropy_statistics_stage_1} in \Cref{appendix:entropy_statistics}, we show various statistics related to the entropy of $\currentpolicy$ over the course of Stage 1. We find that the mean entropy increases monotonically, and even the $60^{\textup{th}}$ percentile of entropy rises drastically. Intuitively, this is harmful --- it suggests that DAPO increases the entropy for tokens for which the model should have a relatively high certainty. This analysis of entropy statistics is motivated by the work of \citet{wang20258020rulehighentropyminority}, who recommend only performing gradient updates for tokens in the higher percentiles of entropy, during RLVR.

Thus, in Stage 2, we continue from our Stage 1 checkpoint and use CISPO instead of DAPO. We find that CISPO improves the accuracy monotonically for several more steps, and CISPO does not exhibit the phenomenon of monotonically increasing entropy as DAPO does. Stage 2 eventually also encounters a plateau, and then a decline, in accuracy (\Cref{fig:data_filtering_stage_2_plateau} in \Cref{appendix:intermediate_plateaus_rl}). The longest path length continues to decrease sharply, at the expense of accuracy. We hypothesize that this is because of filtering strategy \firstfilter --- for a problem $\problem$ where all responses are correct, the advantage of a response is solely a function of its longest path length. Thus, in Stage 3, in order to resolve the plateau in accuracy, we switch to filtering strategy \secondfilter.

Stage 3 improves the accuracy, but eventually reaches a plateau, and the amount of truncated responses increases. Thus, in Stage 4, we now increase the allowed longest path length $\maxplen$ from $7,500$ to $12,000$, and increase the longest path length penalty cutoff $\penaltycutoff$ from $2,000$ to $6,500$.

\paragraph{High length penalty (HLP).} After finishing stage 4, we aim to decrease the longest path length significantly, without significantly reducing accuracy. Thus, we increase $\lpcoeff$ from $0.1$ to $0.9$, and we let $\penaltycutoff = 2000$ --- in other words, all responses with longest path length more than $2000$ are penalized, and the length penalty can range up to $0.9$. Our reward function in the HLP stage is
\[
\rewardfn(\response) := 
\begin{cases}
1 - \lengthpenaltyfn(\response) & \text{if $\response$ is correct and satisfies format} \\
0.01 &
\text{if $\response$ is correct but does not satisfy format} \\
0 & \text{otherwise}
\end{cases}
\]

Here, we only apply the length penalty to responses that are both correct and satisfy the format --- our reasoning is as follows. With our reward function $\rewardfn(\response)$, we ensure that responses that are correct, but do not satisfy the format, will not have higher reward than responses that are correct and satisfy the format. Since $1 - \lengthpenaltyfn(\response)$ could be as low as $0.1$ for a response $\response$ that is correct and satisfies the format, we must define $\rewardfn$ so that responses which are correct and do not satisfy the format always have reward less than $0.1$. Additionally, we would like all correct responses (regardless of format) to have higher reward than incorrect responses. Thus, in summary, all responses $\response$ that are correct, but do not satisfy the format, should have reward strictly greater than $0$ and less than $0.1$. We believe any value between $0$ and $0.1$ would work well --- we choose $0.01$ arbitrarily. Finally, we do not apply the length penalty to responses that are correct, but do not satisfy the format, to ensure that their reward does not fall below $0$. We continue to use filtering strategy \secondfilter.

\section{Experiments}

\subsection{Experimental Setup} \label{subsec:experimental_setup}

\paragraph{SFT Dataset --- Creation of $\initrolloutdata$ and $\parallelrewritedata$.} Our base model $\basemodel$ is DeepScaleR-1.5B-Preview \citep{deepscaler2025} and we obtain its responses on the first $15K$ problems in the DeepScaleR training dataset to obtain $\initrolloutdata$. We verify the accuracy of the responses using \texttt{math\_verify}. \footnote{\url{https://github.com/huggingface/Math-Verify}} We use temperature $0.6$ and set $\texttt{top\_p} = 0.95$ when sampling with vLLM \citep{kwon2023efficient}. We allow response length up to $24K$ and use the prompt shown in \Cref{box:sft_sequential_cot_prompt}. After rejection sampling, we end up with $9968$ problem-answer pairs --- we include the first $4K$ of these in $\initrolloutdata$. We use Claude Sonnet 4.5 as our rewriting model $\rewritemodel$. \footnote{The specific version we use is $\claudemodel$.} When generating with $\rewritemodel$, we use temperature $0.7$ and response length $12K$. Our final dataset $\parallelrewritedata$ contains $3411$ examples.

\paragraph{SFT Hyperparameters.} For SFT, we use the prompt in \Cref{box:sft_rl_prompt}. We use batch size $8$, and learning rate $5 \cdot 10^{-6}$, for $1800$ steps. We use linear learning rate warmup for $105$ steps. We construct the training sequences as described in \Cref{subsec:sft_train_seq_position_id_attn_mask}, and remove any training sequences with length more than $10,000$ tokens. We use $8$ H100 GPUs for SFT, and for future experiments, unless otherwise specified.

\paragraph{RL Details and Hyperparameters.} We use \texttt{verl} \citep{sheng2024hybridflow}, with modifications for our inference architecture (\Cref{sec:inference}) and the customized training sequences, attention mask and position IDs discussed in \Cref{subsec:sft_train_seq_position_id_attn_mask}. We use the DeepScaleR training dataset \citep{deepscaler2025}, and AIME 2024 \citep{maa_aime2024} for our validation dataset, performing $50$ rollouts per problem during each validation step. We use $\texttt{math\_verify}$ to verify accuracy. Our prompt is given in \Cref{box:sft_rl_prompt}. Stages 1, 2, 3 and 4 last for $700$, $480$, $240$ and $220$ steps respectively. The HLP stage lasts for $200$ steps. Learning rates and batch sizes are given in \Cref{appendix:extra_hyperparameters_details}.

\paragraph{Evaluation Benchmarks.} For evaluation, we use AIME 2024 \citep{maa_aime2024}, AMC 23 \citep{maa_amc12_problems_solutions}, HMMT Feb 2025 \citep{balunović2025matharenaevaluatingllmsuncontaminated}, MATH 500 \citep{hendrycks2021measuringmathematicalproblemsolving}, Minerva Math \citep{lewkowycz2022solvingquantitativereasoningproblems}, and Olympiad-Bench \citep{he-etal-2024-olympiadbench}. To evaluate $\text{pass}@1$ and $\text{maj}@3$, for AIME 2024, AMC 23 and HMMT 2025, we use $200$ rollouts per problem, and for MATH 500, Minerva Math, and Olympiad-Bench, we use $16$ rollouts per problem.

\paragraph{Baselines.} We report the accuracy and longest path length of DeepScaleR-1.5B-Preview \citep{deepscaler2025}, using a response length of $32,768$ tokens, denoted as \textbf{DSR-32K}, and $12K$ tokens, denoted as \textbf{DSR-12K}. We use the prompt shown in \Cref{box:deepscaler_prompt}, following DeepScaleR \citep{deepscaler2025}.

To further increase the length efficiency of the baseline, we apply an HLP stage to DeepScaleR-1.5B-Preview. We train and evaluate DeepScaleR-1.5B-Preview with a high length penalty allowing $24,000$ tokens per response (denoted as \textbf{DSR-HLP-24K}) and allowing $12,000$ tokens per response (denoted as \textbf{DSR-HLP-12K}). Hyperparameters for training  DSR-HLP-24K and DSR-HLP-12K are given in \Cref{appendix:extra_hyperparameters_details}.

We also evaluate DSR-12K using $\text{maj}@3$ --- we refer to this method as \textbf{Maj-DSR-3}. Given $N$ responses for a single problem (where $N$ is $200$ or $16$), we evaluate the accuracy/longest path length when taking the majority vote for all $\binom{N}{3}$ subsets of $3$ responses. Details are in \Cref{appendix:extra_hyperparameters_details}.

\subsection{Results}
\label{subsec:results}

\begin{figure*}[h]
\centering
\includegraphics[width=0.9\textwidth]{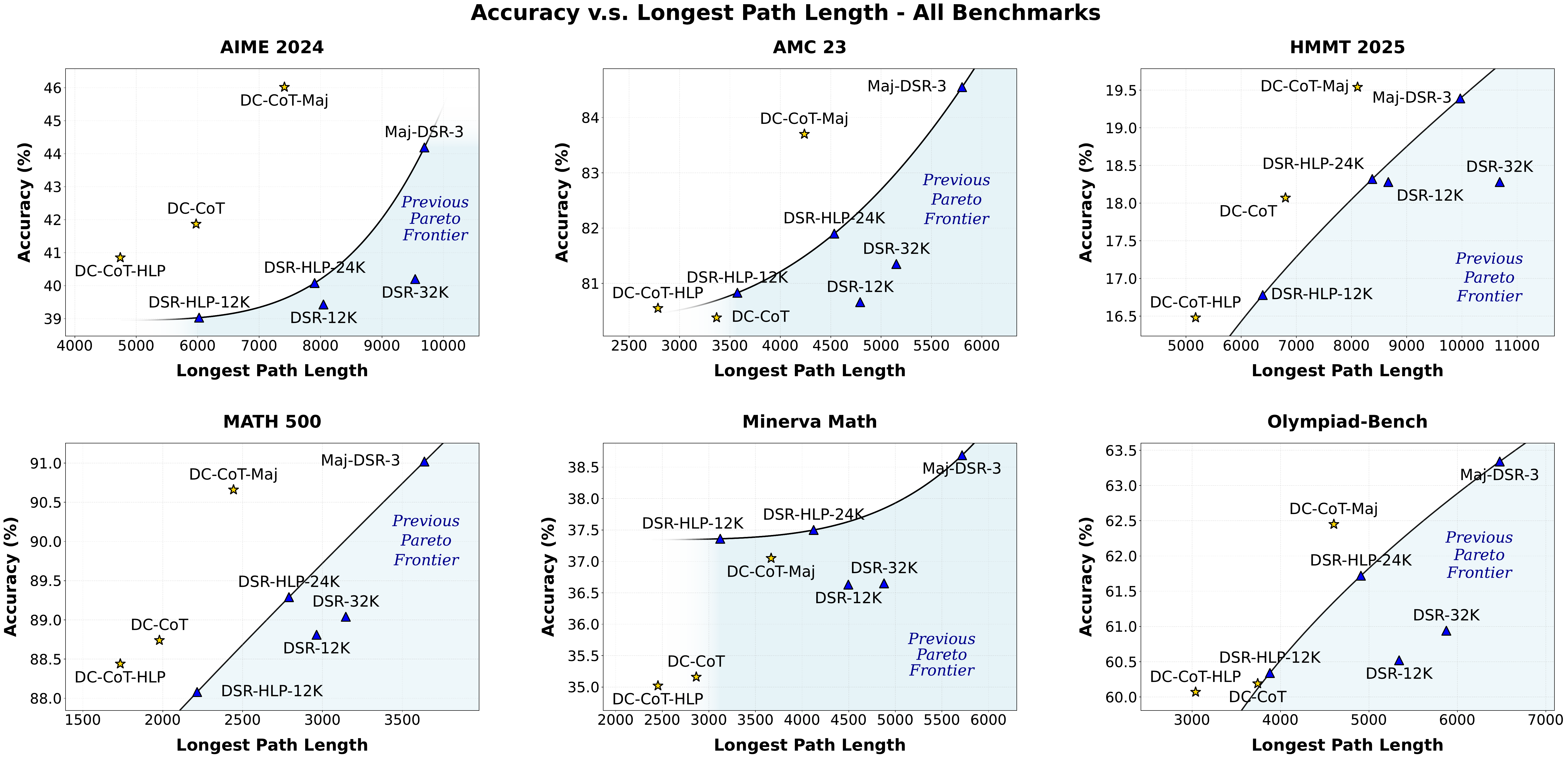}
\caption{Accuracy (pass@1) and longest path length for all methods on all benchmarks.}
\label{fig:all_2d_plots}
\end{figure*}

\begin{table*}[h]
\centering
\caption{Accuracy and longest path length (LPL) for all methods and benchmarks discussed in \Cref{subsec:experimental_setup}. Error bars for all accuracies and longest path lengths shown in \Cref{appendix:additional_tables_plots}, \Cref{tab:main_results_aime_amc_hmmt} and \Cref{tab:main_results_math_mm_ob}.}
\label{tab:main_results}

\small
\setlength{\tabcolsep}{3pt}
\renewcommand{\arraystretch}{1.15}

\begin{tabular}{l *{12}{c}}
\toprule
\textbf{Method}
& \multicolumn{2}{c}{\textbf{AIME 24}}
& \multicolumn{2}{c}{\textbf{AMC}}
& \multicolumn{2}{c}{\textbf{HMMT}}
& \multicolumn{2}{c}{\textbf{MATH}}
& \multicolumn{2}{c}{\textbf{MM}}
& \multicolumn{2}{c}{\textbf{OB}} \\
\cmidrule(lr){2-3}\cmidrule(lr){4-5}\cmidrule(lr){6-7}\cmidrule(lr){8-9}\cmidrule(lr){10-11}\cmidrule(lr){12-13}
& \textbf{Acc (\%)} & \textbf{LPL}
& \textbf{Acc (\%)} & \textbf{LPL}
& \textbf{Acc (\%)} & \textbf{LPL}
& \textbf{Acc (\%)} & \textbf{LPL}
& \textbf{Acc (\%)} & \textbf{LPL}
& \textbf{Acc (\%)} & \textbf{LPL} \\
\midrule

\textbf{DSR-32K}
& $40.20$ & $9540$
& $81.35$ & $5151$
& $18.28$ & $10685$
& $89.04$ & $3146$
& $36.65$ & $4879$
& $60.94$ & $5875$ \\

\textbf{DSR-12K}
& $39.43$ & $8047$
& $80.66$ & $4792$
& $18.28$ & $8666$
& $88.81$ & $2963$
& $36.63$ & $4496$
& $60.52$ & $5341$ \\

\textbf{Maj-DSR-3}
& $44.19$ & $9690$
& $84.55$ & $5803$
& $19.39$ & $9970$
& $91.02$ & $3638$
& $38.69$ & $5716$
& $63.34$ & $6479$ \\

\textbf{DSR-HLP-24K}
& $40.08$ & $7903$
& $81.90$ & $4535$
& $18.32$ & $8375$
& $89.29$ & $2790$
& $37.50$ & $4124$
& $61.72$ & $4909$ \\

\textbf{DSR-HLP-12K}
& $39.03$ & $6023$
& $80.83$ & $3574$
& $16.78$ & $6391$
& $88.08$ & $2215$
& $37.36$ & $3122$
& $60.34$ & $3880$ \\
\midrule

\textbf{\ourmethod}
& $41.87$ & $5974$
& $80.38$ & $3368$
& $18.07$ & $6803$
& $88.74$ & $1979$
& $35.16$ & $2865$
& $60.19$ & $3742$ \\

\textbf{\ourmethod-HLP}
& $40.85$ & $4740$
& $80.55$ & $2786$
& $16.48$ & $5175$
& $88.44$ & $1734$
& $35.02$ & $2454$
& $60.07$ & $3040$ \\

\textbf{\ourmethod-Maj}
& $46.02$ & $7412$
& $83.70$ & $4239$
& $19.54$ & $8109$
& $90.66$ & $2443$
& $37.05$ & $3668$
& $62.45$ & $4603$ \\

\bottomrule
\end{tabular}
\end{table*}

\begin{table*}[h]
\centering
\caption{Change in accuracy and longest path length of our methods compared to appropriate baselines. For \textbf{\ourmethod} we use \textbf{DSR-32K} as the baseline, for \textbf{\ourmethod-HLP} we use \textbf{DSR-HLP-12K} as the baseline, and for \textbf{\ourmethod-Maj} we use \textbf{Maj-DSR-3} as the baseline.}
\label{tab:change_compared_to_baselines}

\small
\setlength{\tabcolsep}{3pt}
\renewcommand{\arraystretch}{1.15}

\begin{tabular}{l l *{12}{c}}
\toprule
\textbf{Method} & \textbf{Baseline}
& \multicolumn{2}{c}{\textbf{AIME 24}}
& \multicolumn{2}{c}{\textbf{AMC}}
& \multicolumn{2}{c}{\textbf{HMMT}}
& \multicolumn{2}{c}{\textbf{MATH}}
& \multicolumn{2}{c}{\textbf{MM}}
& \multicolumn{2}{c}{\textbf{OB}} \\
\cmidrule(lr){3-4}\cmidrule(lr){5-6}\cmidrule(lr){7-8}\cmidrule(lr){9-10}\cmidrule(lr){11-12}\cmidrule(lr){13-14}
& & \textbf{Acc} & \textbf{LPL}
& \textbf{Acc} & \textbf{LPL}
& \textbf{Acc} & \textbf{LPL}
& \textbf{Acc} & \textbf{LPL}
& \textbf{Acc} & \textbf{LPL}
& \textbf{Acc} & \textbf{LPL} \\
\midrule

\textbf{\ourmethod} & \textbf{DSR-32K}
& +1.67 & -37.4\%
& -0.97 & -34.6\%
& -0.21 & -36.3\%
& -0.3 & -37.1\%
& -1.49 & -41.3\%
& -0.75 & -36.3\% \\

\textbf{\ourmethod-HLP} & \textbf{DSR-HLP-12K}
& +1.82 & -21.3\%
& -0.28 & -22.0\%
& -0.3 & -19.0\%
& +0.36 & -21.7\%
& -2.34 & -21.4\%
& -0.27 & -21.6\% \\

\textbf{\ourmethod-Maj} & \textbf{Maj-DSR-3}
& +1.83 & -23.5\%
& -0.85 & -27.0\%
& +0.15 & -18.7\%
& -0.36 & -32.8\%
& -1.64 & -35.8\%
& -0.89 & -29.0\% \\

\bottomrule
\end{tabular}
\end{table*}

\paragraph{Main results --- accuracy and longest path length.} \Cref{fig:all_2d_plots} and \Cref{tab:main_results} compare our methods (\textbf{\ourmethod}, \textbf{\ourmethod-HLP}) with the aforementioned baselines. \ourmethod consistently improves the accuracy-longest path length tradeoff compared to baselines. \ourmethod-HLP further improves the longest path length significantly, at the cost of a slight decrease in accuracy compared to \ourmethod. Across our benchmarks, \ourmethod matches (and sometimes exceeds) the accuracy of the strongest non-HLP baseline DSR-32K, while reducing the longest path length by roughly 35-40\% --- see \Cref{tab:change_compared_to_baselines} for percentage decreases in longest path length. \ourmethod-HLP achieves comparable accuracy to the HLP baseline DSR-HLP-12K trained and evaluated with the same longest path length, while achieving a roughly 20\% reduction in longest path length on all benchmarks. Additionally, we evaluate \ourmethod using maj@3, using an identical setup to Maj-DSR-3 --- we refer to this method as \textbf{\ourmethod-Maj}. As shown in \Cref{tab:change_compared_to_baselines}, \ourmethod-Maj achieves similar or higher accuracy than Maj-DSR-3, while achieving significant reductions in longest path length. \Cref{tab:total_tokens} in \Cref{appendix:additional_tables_plots} shows the total tokens used per response by our methods.

On AIME 2024 (\Cref{fig:aime_2d_plot}), the methods \ourmethod, \ourmethod-HLP and \ourmethod-Maj are beyond the Pareto frontier of accuracy and longest path length achieved by previous methods --- in particular, both \ourmethod and \ourmethod-HLP are a Pareto improvement compared to DSR-32K, DSR-12K, and DSR-HLP-24K. On HMMT 2025 (\Cref{fig:all_2d_plots}, top right), \ourmethod achieves similar accuracy (<0.3\% lower) to DSR-32K, DSR-12K and DSR-HLP-24K, while achieving a $36\%$ reduction in longest path length compared to DSR-32K, a $21.5\%$ reduction compared to DSR-12K, and an $18.8\%$ reduction compared to DSR-HLP-24K. Similarly, \ourmethod-HLP achieves a similar accuracy as DSR-HLP-12K, with $19\%$ lower longest path length. Finally, \ourmethod-Maj is a Pareto improvement over Maj-DSR-3, achieving slightly higher accuracy with $18.7\%$ less longest path length. On Minerva Math (MM), \ourmethod and \ourmethod-HLP obtain worse accuracy than the baselines. This may be because MM includes many problems that rely on scientific concepts, while our training dataset only includes math problems. We leave more diverse training datasets and improved performance on MM to future work.

\begin{figure}[h]
    \centering
    \includegraphics[width=0.8\textwidth]{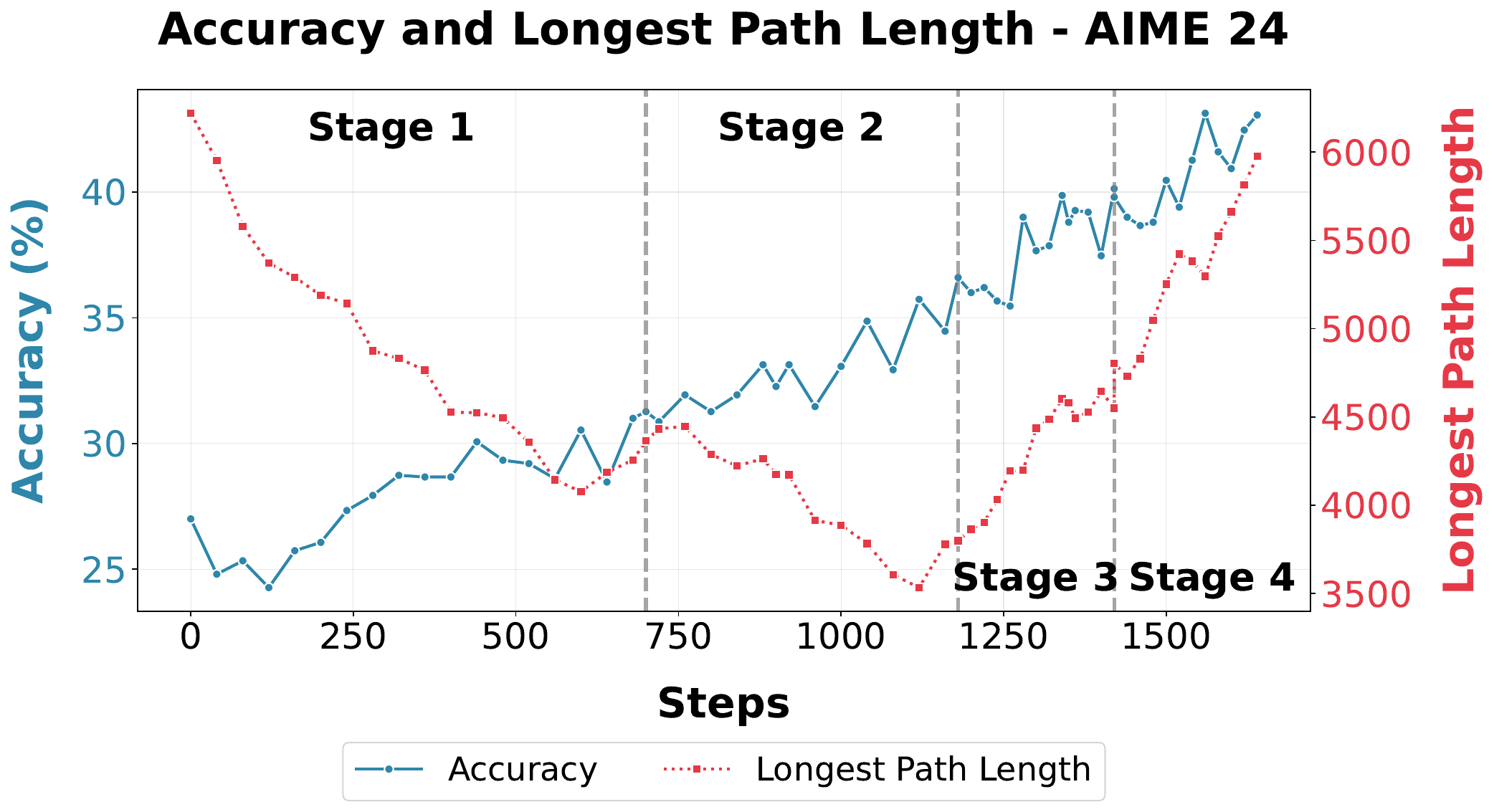}
    \caption{Accuracy and longest path length of \textbf{\ourmethod} on AIME 2024 during all stages of RL.}
    \label{fig:validation_accuracy_and_parallel_length}
\end{figure}

\begin{figure}[h]
    \centering
    \includegraphics[width=0.8\textwidth]{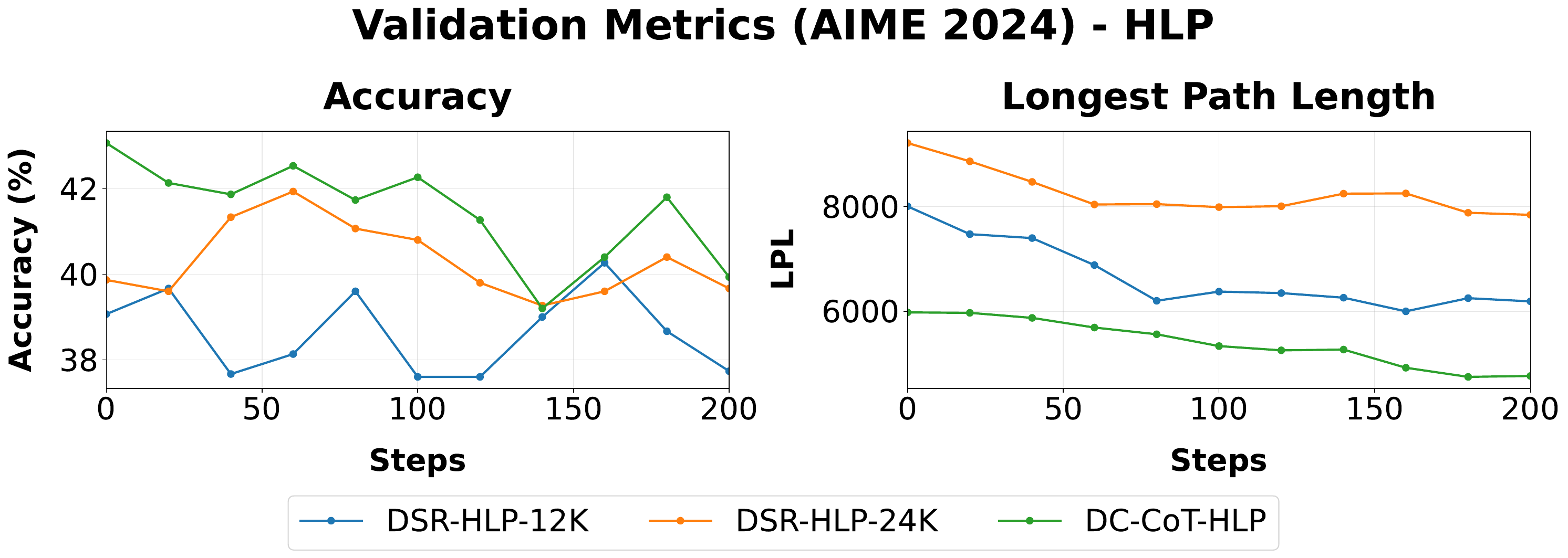}
    \caption{Accuracy (pass@1) and longest path length on AIME 2024 for \textcolor{blue}{DSR-HLP-12K}, \textcolor{orange}{DSR-HLP-24K}, and \textcolor{green}{\ourmethod-HLP}.}
    \label{fig:hlp_accuracy_lpl}
\end{figure}

\paragraph{RL training dynamics.} \Cref{fig:validation_accuracy_and_parallel_length} shows the accuracy and longest path length of \ourmethod over the course of RL, on AIME 2024. \Cref{fig:validation_dp} in \Cref{appendix:additional_tables_plots} shows the degree of parallelism (DP), which for a response $\response$ is defined as the ratio between the total tokens in $\response$ and the longest path length. The validation accuracy of our model improves monotonically, due to the multi-stage training discussed in \Cref{subsec:rl_algorithm_explanation}. The longest path length decreases, and the degree of parallelism increases, during Stages 1 and 2, when we use filtering strategy \firstfilter. However, when we switch to \secondfilter during Stages 3 and 4, the model uses more sequential thinking to help improve the accuracy. As a result, the longest path length increases and the degree of parallelism drops in these two stages.

The accuracy and longest path length of the methods trained with HLP (DSR-24K-HLP, DSR-12K-HLP, and \ourmethod-HLP) are shown in \Cref{fig:hlp_accuracy_lpl}, and the DP of \ourmethod-HLP is shown in \Cref{fig:hlp_dp} in \Cref{appendix:additional_tables_plots}. The accuracy of DSR-HLP-12K and DSR-HLP-24K stays roughly constant, while there is a slight decline in the accuracy of \ourmethod-HLP. There is a steady decrease in the longest path lengths for all three methods. Corresponding to this decrease in longest path length, the degree of parallelism for \ourmethod-HLP also increases during HLP, from slightly under $1.2$, to around $1.37$. We evaluate checkpoints from towards the end of HLP for all methods, based on our assessment of validation accuracy and longest path length. We use step $160$ for DSR-12K-HLP, step $180$ for DSR-24K-HLP, and step $180$ for \ourmethod-HLP in all tables and figures.

\paragraph{Additional Results in Appendices.} In \Cref{appendix:fewer_stage_RL}, we study versions of our RL run that use fewer stages --- the results suggest that all stages may play an important role in our final performance. In \Cref{appendix:latency_table}, we evaluate latency in wall-clock time for \textbf{\ourmethod}, \textbf{\ourmethod-HLP} and baselines, finding that improvements in longest path length translate to improvements in wall-clock time. In \Cref{appendix:decomposability}, we analyze rollouts qualitatively, finding that it is common for workers to cross-verify the same result through different approaches. We note that repeated verifications are common in standard CoTs, and contribute significantly to their length --- thus, parallelizing them is useful.

\paragraph{Limitations.} Our base model has 1.5B parameters. However, we believe larger models will benefit significantly from DC-CoT, since they will be more capable of proposing subtasks that lead to more efficient parallelization of solutions. Additionally, we train and evaluate on mathematical tasks. An interesting direction for future work would be to apply DC-CoT to settings such as coding agents and web search, which may be amenable to parallelization.

\section*{Acknowledgments}

The authors would like to thank Luke Bailey, Neil Band, Caroline Choi and Thomas Chen for helpful discussions. The authors would also like to thank support from NSF CIF 2212263 and the National Defense Science and Engineering Graduate (NDSEG) Fellowship, and the Marlowe cluster at Stanford University \citep{stanford_marlowe} for providing the H100 GPUs required for the experiments.

\bibliography{references}
\bibliographystyle{plainnat}

\newpage
\appendix
\onecolumn

\appendix

\section{Additional Related Work} \label{appendix:additional_related_work}

We discuss related works which are omitted from \Cref{sec:related_work}.

\paragraph{Parallel Sampling for Inference-Time Scaling} Several works study ways to improve the performance of LLMs at inference time using independent samples, as well as more advanced generation strategies \citep{brown2024largelanguagemonkeysscaling, snell2024scalingllmtesttimecompute, zhang2024accessinggpt4levelmathematical, wu2025modeconditioningunlockssuperiortesttime, qi2025learningreasonparallelsamples, zhao2025majorityrightrltraining}. 
\citet{wang2023selfconsistencyimproveschainthought} show that majority voting can achieve good performance when combined with CoT. \citet{yao2023treethoughtsdeliberateproblem} propose Tree-of-Thoughts (ToT), and show that allowing an LM to backtrack can lead to performance improvements compared to standard CoT (for instance, greatly improving the performance of GPT-4 on Game of 24).
\citet{brown2024largelanguagemonkeysscaling} study how independent sampling aggregation methods, such as majority voting and best-of-N with a reward model, scale as the number of samples increases. \citet{snell2024scalingllmtesttimecompute} study compute-optimal allocation of test-time compute, comparing sequential refinement with strategies such as beam search with a process reward model. \citet{wu2025modeconditioningunlockssuperiortesttime} aim to improve parallel sampling by heuristically guiding different samples to be drawn from different modes of the distribution, with the goal of ensuring that different samples pursue different strategies. \citet{wen2025parathinkernativeparallelthinking} train a model with SFT to generate multiple independent samples in parallel, then generate a summary --- by prepending the token $\texttt{<think\_i>}$ to the start of the $i^{th}$ sample, \citet{wen2025parathinkernativeparallelthinking} aim to have the samples pursue different strategies. \citet{qi2025learningreasonparallelsamples} and \citet{zhao2025majorityrightrltraining} instead use RL to teach their models to aggregate independent samples from another model to arrive at the correct answer. In our method, the orchestrator does reasoning after seeing the problem and prior to assigning tasks. Thus, the workers perform distinct tasks and can terminate once their assigned task is done, leading to an improvement in latency, unlike with parallel sampling. \citet{hassid2025dontoverthinkitpreferring} propose \textit{short-$m@k$} --- this method is an alternative to majority voting, where $k$ independent samples are generated, and once the shortest $m$ samples are finished, generation is terminated. The final answer is then obtained by majority voting among these $m$ responses. Our work is orthogonal to \citet{hassid2025dontoverthinkitpreferring}, as it is possible for our method to further improve the latency required by the shortest sample.

\paragraph{RLVR and Long CoT for Mathematical Reasoning} \citet{openai2024openaio1card} and \citet{deepseekai2025deepseekr1incentivizingreasoningcapability} introduce reasoning models, which are trained using RL to output long chains-of-thought (CoTs) which dramatically improve the models' capabilities, including in mathematical reasoning. \citet{deepseekai2025deepseekr1incentivizingreasoningcapability} train their model with RL using only verifiable rewards (RLVR), using GRPO \citep{shao2024deepseekmathpushinglimitsmathematical}. \citet{min2024imitateexploreselfimprovereproduction} and \citet{guha2025openthoughtsdatarecipesreasoning} curate demonstrations to teach models to perform long CoTs. Many works aim to achieve open-source reproductions of the RL procedure of \citet{deepseekai2025deepseekr1incentivizingreasoningcapability} using RLVR, as well as understand the impact of various hyperparameters and design choices \citep{deepscaler2025, yu2025dapoopensourcellmreinforcement, liu2025prorlprolongedreinforcementlearning, he2025skyworkopenreasoner1, hu2025prorlv2, hu2025brorlscalingreinforcementlearning, Polaris2025, liu2025itrickstrapsdeep, song2025fastcurlcurriculumreinforcementlearning, liu2025dlerdoinglengthpenalty, setlur2025e3learningexploreenables, cui2025entropymechanismreinforcementlearning, tissue2025entropyschdule}. \citet{he2025justrlscaling15bllm} aim to simplify existing recipes and identify which of the design choices are necessary for training stability and monotonic improvement. \citet{minimax2025minimaxm1scalingtesttimecompute} propose the CISPO algorithm, which is different from GRPO, using an objective akin to REINFORCE \citep{reinforce1992} but with truncated importance sampling  --- \citet{khatri2025artscalingreinforcementlearning}, among other findings, show that CISPO works well. \citet{yang2025pencillongthoughtsshort} show that on synthetic tasks, long CoTs can benefit from removing information that is no longer useful from the context.

\clearpage
\section{Formal Pseudocode for Inference Architecture} \label{appendix:inference_pseudocode}

In \Cref{algorithm:inference_algorithm}, we present formal pseudocode for the inference architecture discussed in \Cref{sec:inference}.

\begin{algorithm}
\caption{Algorithm for inference with a language model $\model$ that spawns worker threads.}
\label{algorithm:inference_algorithm}
\begin{algorithmic}[1]
\Require Language model $\model$, Prompt $\prompt$, number of workers $K$, longest path length budget $L$
\State $\completion \gets \infer(\model, \prompt, \Wstart, L)$ \label{alg:line:single_thread_init}
\State $\tempprompt \gets \prompt + \completion$
\State $L \gets L - |\completion|$ \label{alg:line:decrement_parallel_len1}
\State $\textsc{Response} \gets S$
\State 
\While{True}
\If{$\textsc{Response}$ contains \EOS token}
    \State \Return $\textsc{Response}$
\EndIf
\For{$i \gets [1 \ldots K]$ in parallel}
  \State \# Create prompts for each worker thread.
  \State $\prompt_i \gets \tempprompt + \Wstarti{i}$ \label{alg:line:create_worker_prompt}
  \State \# Do inference for each worker thread.
  \State $W_i \gets \infer(M, \prompt_i, \Wendi{i}, L)$ --- this is done asynchronously \label{alg:line:worker_inference}
\EndFor
\State \textbf{Wait} for $W_i$ for $i \in [K]$
\State $L \gets L - \max_{i \in [K]} |W_i|$ \label{alg:line:decrement_parallel_len2}
\For{$i \gets [1 \ldots K]$}
\State $\tempprompt \gets \tempprompt + \Wstarti{i} + W_i $ \label{alg:line:add_worker_to_prefix}
\State $\textsc{Response} \gets \textsc{Response} + \Wstarti{i} + W_i$
\EndFor
    \State $\tempprompt \gets \tempprompt + \Wend$
    \State $\textsc{Response} \gets \textsc{Response} + \Wend$
    \State $S \gets \infer(M, \tempprompt, \Wstart, L)$ \label{alg:line:back_to_single_thread}
    \State $\tempprompt \gets \tempprompt + S$
    \State $\textsc{Response} \gets \textsc{Response} + S$
    \State $L \gets L - |S|$ \label{alg:line:decrement_parallel_len3}
\EndWhile

\State \textbf{Return} \textsc{Response}
\end{algorithmic}
\end{algorithm}

\clearpage
\section{Detailed Inference Example} \label{appendix:detailed_inference_figure}

\begin{figure}[h]
    \centering
    \resizebox{\textwidth}{!}{\begin{tikzpicture}[node distance=\verticaldist and \horizontaldist]

    \node[inference_problem] (P) {
        \textbf{Problem (P)}\\[0.2cm]
        Find the sum of all real roots $x$ of the equation:
        \[
        \left(2^{x} - 4\right)^{3} + \left(4^{x} - 2\right)^{3} = \left(4^{x} + 2^{x} - 6\right)^{3}
        \]
    };

    \node[inference_solution, below=of P] (S1) {
        \textbf{Single-thread portion 1 (S1)}\\[0.15cm]
        \texttt{<think>}\\[0.1cm]
        Let me set $y = 2^x$, which means $4^x = (2^x)^2 = y^2$.\\[0.1cm]
        Now, let's denote:\\
        \quad $A = y - 4$\\
        \quad $B = y^2 - 2$\\
        \quad $C = y^2 + y - 6$\\[0.1cm]
        The equation becomes: $A^3 + B^3 = C^3$\\[0.05cm]
        Since $A + B = C$, this simplifies to: $-3AB \cdot C = 0$\\[0.1cm]
        So, either $C = 0$ or $AB = 0$.\\[0.15cm]
        Now, I can have the three workers explore these cases in parallel:\\
        \quad $\bullet$ Worker 1 will explore the case $C = 0$\\
        \quad $\bullet$ Worker 2 will explore the case $AB = 0$\\
        \quad $\bullet$ Worker 3 will verify the algebraic manipulation that $A + B = C$\\[0.05cm]
        These are independent investigations that don't depend on each other's results.\\[0.1cm]
        \texttt{<spawn\_workers>}
    };

    \node[inference_worker, below left=of S1, xshift=-2cm] (W1) {
        \textbf{Worker 1 (W1)}\\[0.15cm]
        \texttt{<worker\_1>}\\[0.1cm]
        \textbf{Case 1:} $C = 0$\\[0.1cm]
        $C = y^2 + y - 6 = 0$\\[0.05cm]
        Solve: $y = 2$ or $y = -3$\\
        Valid: $y = 2$ (since $y > 0$)\\[0.1cm]
        Thus, $y = 2$, so $2^x = 2 \Rightarrow x = 1$.\\[0.1cm]
        \texttt{</worker\_1>}
    };

    \node[inference_worker, below=of S1] (W2) {
        \textbf{Worker 2 (W2)}\\[0.15cm]
        \texttt{<worker\_2>}\\[0.1cm]
        \textbf{Exploring} $AB = 0$ \textbf{case}\\[0.1cm]
        \textbf{Case 1:} $A = 0 \Rightarrow y = 4 \Rightarrow 2^x = 4 \Rightarrow x = 2$\\[0.1cm]
        \textbf{Case 2:} $B = 0 \Rightarrow y^2 = 2 \Rightarrow y = \sqrt{2}$ (valid, $y > 0$)\\
        Thus, $2^x = \sqrt{2} = 2^{1/2} \Rightarrow x = \frac{1}{2}$\\[0.1cm]
        Therefore, from $AB = 0$, we get $x = 2$ and $x = \frac{1}{2}$.\\[0.1cm]
        \texttt{</worker\_2>}
    };

    \node[inference_worker, below right=of S1, xshift=2cm] (W3) {
        \textbf{Worker 3 (W3)}\\[0.15cm]
        \texttt{<worker\_3>}\\[0.1cm]
        \textbf{Verifying the algebraic manipulation:}\\[0.1cm]
        We have $A = y - 4$, $B = y^2 - 2$, $C = y^2 + y - 6$.\\[0.1cm]
        Compute $A + B$:\\
        $(y - 4) + (y^2 - 2) = y^2 + y - 6 = C$.\\
        So yes, $A + B = C$.\\[0.1cm]
        Therefore, $A^3 + B^3 = C^3$ becomes:\\
        $C^3 - 3AB \cdot C = C^3$\\[0.05cm]
        Subtract $C^3$: $-3AB \cdot C = 0$\\
        $\Rightarrow$ Either $C = 0$ or $AB = 0$.\\[0.1cm]
        \texttt{</worker\_3>}
    };

    \node[inference_solution, below=of W2, yshift=-1cm] (S2) {
        \textbf{Single-thread portion 2 (S2)}\\[0.15cm]
        \texttt{</spawn\_workers>}\\[0.1cm]
        Great, so we have confirmed the cases:\\[0.1cm]
        From $C = 0$, we get $x = 1$.\\
        From $AB = 0$, we get $x = 2$ and $x = \frac{1}{2}$.\\[0.1cm]
        We need to check if these are all the real roots.\\
        \ldots\\
        Therefore, these are all the real roots. Now, we need to sum all real roots:\\[0.05cm]
        $x = \frac{1}{2}, 1, 2$\\[0.1cm]
        Sum: $\frac{1}{2} + 1 + 2 = \frac{7}{2}$\\[0.1cm]
        Thus, the answer is $\frac{7}{2}$.\\[0.1cm]
        \texttt{</think>}\\[0.15cm]
        \texttt{<answer>}\\
        \ldots\\
        Thus, the sum of all real roots is $\boxed{\frac{7}{2}}$.\\
        \texttt{</answer>}
    };

    \draw[inference_arrow] (P) -- node[right] {\arrowfont Start solving} (S1);
    \draw[inference_arrow] (S1) -- node[left, pos=0.3] {\arrowfont Spawn} (W1);
    \draw[inference_arrow] (S1) -- node[right, pos=0.3] {\arrowfont Spawn} (W2);
    \draw[inference_arrow] (S1) -- node[right, pos=0.3] {\arrowfont Spawn} (W3);
    \draw[inference_arrow] (W1) -- node[left, pos=0.7] {\arrowfont Return} (S2);
    \draw[inference_arrow] (W2) -- node[right] {\arrowfont Return} (S2);
    \draw[inference_arrow] (W3) -- node[right, pos=0.7] {\arrowfont Return} (S2);

\end{tikzpicture}}
    \caption{An instantiation of the inference architecture in \Cref{fig:intro_flowchart} with example thoughts. These are a condensed version of the thoughts that our model produced when given this problem during RL training.}
    \label{fig:inference_algorithm_example}
\end{figure}
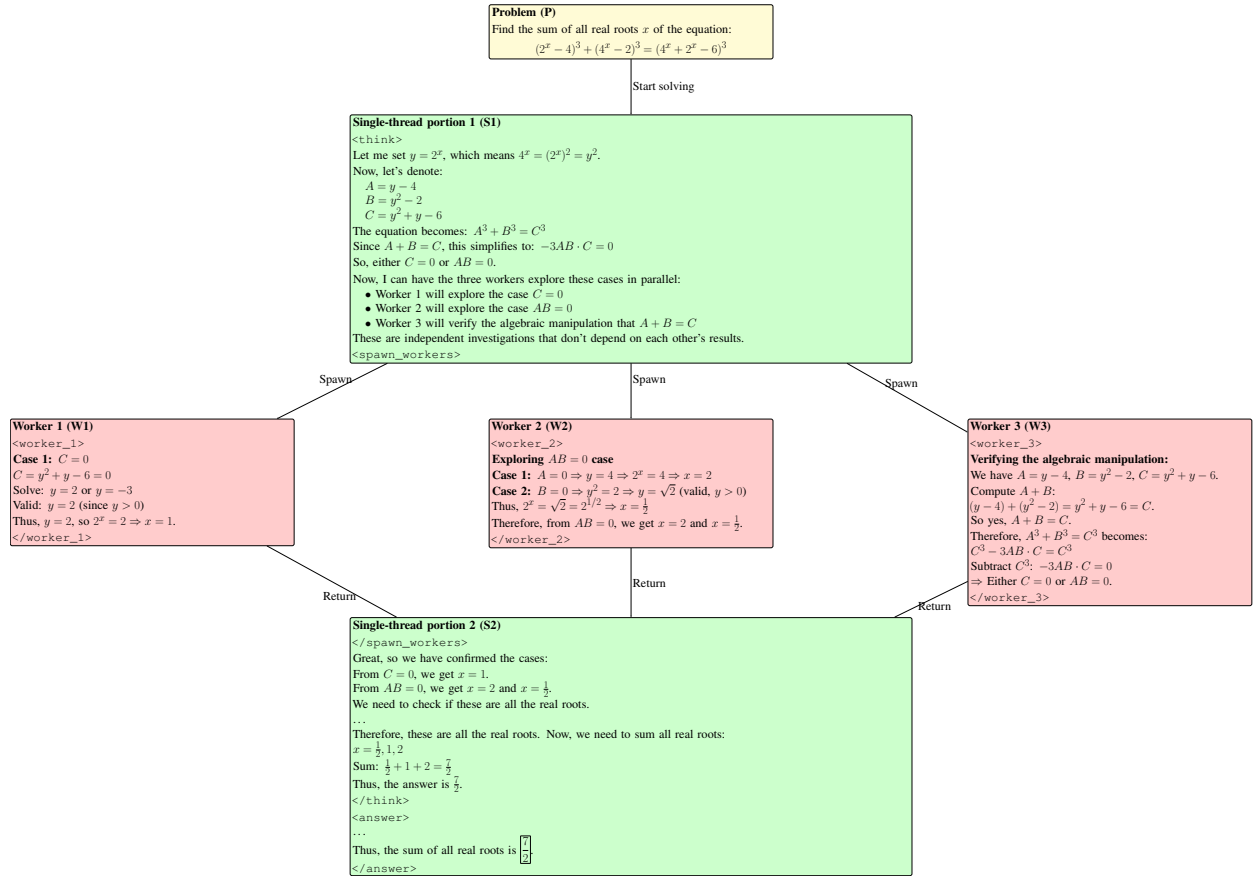

A response that our model generates, for the example problem shown in \Cref{fig:intro_flowchart}, is shown in \Cref{fig:inference_algorithm_example}. The model, in its role as director, initially performs some algebraic manipulations, and finds that either $C = 0$ or $AB = 0$ must hold, where $A = y - 4$, $B = y^2 - 2$, and $C = y^2 + y - 6$. It then assigns these two cases to different workers. Finally, after the workers finish their thinking processes, the model reads their outputs, and computes the sum of each of the solutions $x$ found by workers 1 and 2.

\clearpage
\section{Prompts for SFT and RL} \label{appendix:sft_rl_prompts}

\lstset{
    basicstyle=\ttfamily,
    breaklines=true,
    inputencoding=utf8,
    extendedchars=true,
    literate={｜}{{|}}1 {▁}{{\_}}1
}
\begin{mybox}[box:sft_rl_prompt]{Prompt Used During SFT and RL}
\begin{lstlisting}
prefix = "<｜begin▁of▁sentence｜><｜User｜>"
suffix = "<｜Assistant｜>"
instruction = "Let's think step by step and output the final answer within \\boxed{{}}."
parallel_instruction = "You can spawn multiple workers to solve this problem in parallel." \
                       " The workers' thoughts are enclosed within <spawn_workers></spawn_workers> tags, and each worker's" \
                       " thought is enclosed within <worker_i></worker_i> tags, where i is the worker number, i.e." \
                       " <spawn_workers><worker_1>worker 1's thought</worker_1><worker_2>worker 2's thought</worker_2>..." \
                       "</spawn_workers>."
deepseek_r1_prompt_parallel = prefix + "{question} " + parallel_instruction + " " + instruction + suffix
\end{lstlisting}

\end{mybox}

\clearpage
\section{Prompt for Evaluating DeepScaleR-1.5B-Preview} \label{appendix:deepscaler_prompt}

\lstset{
    basicstyle=\ttfamily,
    breaklines=true,
    inputencoding=utf8,
    extendedchars=true,
    literate={｜}{{|}}1 {▁}{{\_}}1
}
\begin{mybox}[box:deepscaler_prompt]{Prompt Used to Evaluate DeepScaleR-1.5B-Preview}
\begin{lstlisting}
prefix = "<｜begin▁of▁sentence｜><｜User｜>"
suffix = "<｜Assistant｜>"
instruction = "Let's think step by step and output the final answer within \\boxed{{}}."
think_token = "<think>\n"
deepseek_r1_prompt = prefix + "{question} " + instruction + suffix + think_token
\end{lstlisting}

\end{mybox}

\clearpage
\section{RL Methods --- Additional Details} \label{appendix:grpo_cispo}

Here we discuss algorithmic and implementation details that are omitted from \Cref{subsec:rl_algorithm_explanation}.

\paragraph{GRPO/DAPO.} Within an iteration, for each question, we generate $\groupsize$ responses. For each response $\response_i$ for $i = 1, \ldots, \groupsize$, we calculate a reward $\reward_i$. Following \citet{shao2024deepseekmathpushinglimitsmathematical} and \citet{yu2025dapoopensourcellmreinforcement}, we calculate the normalized advantages $\normadv_i := \frac{\reward_i - \advmean}{\advstd}$, where $\advmean$ and $\advstd$ are the mean and standard deviation, respectively, of $\reward_i$ for $i = 1, \ldots, \groupsize$. 

For a mini-batch with problems $\problem^{(j)}$ for $j = 1, \ldots, K$, with each question $\problem^{(j)}$ having responses $\response_i^{(j)}$ for $i = 1,\ldots, \groupsize$, the objective is 
\begin{align*}
\frac{1}{\sum_{j = 1}^K \sum_{i = 1}^{\groupsize} |\response_i^{(j)}|} \sum_{j = 1}^K \sum_{i = 1}^{\groupsize} \sum_{t = 1}^{|\response_i^{(j)}|} \Big( \min(r_{j, i, t}(\theta) \normadv_i^{(j)},\text{ } & \clip(\min(r_{j, i, t}(\theta), 1 - \epslow, 1 + \epshigh) \normadv_i^{(j)}) \\
& - \beta \klpenalty(\currentpolicy \mid \refmodel) \Big)
\end{align*}
where the ratios $r_{j, i, t}(\theta) := \frac{\currentpolicy(\response_{i, t}^{(j)} \mid \problem, \response_{i, < t}^{(j)})}{\oldpolicy(\response_{i, t}^{(j)} \mid \problem, \response_{i, < t}^{(j)})}$, $\currentpolicy$ denotes the current policy with up-to-date parameters, $\oldpolicy$ denotes the policy used to generate the responses, $\response_{i, t}^{(j)}$ denotes the $t^{th}$ token of the response, and $\response_{i, < t}^{(j)}$ denotes the prefix up to (but not including) the $t^{th}$ token. We estimate $\klpenalty(\currentpolicy \mid \refmodel)$ in a manner similar to \citet{shao2024deepseekmathpushinglimitsmathematical}, using the formula $\frac{\refmodel(\response_{i, t}^{(j)} \mid \problem, \response_{i, < t}^{(j)})}{\currentpolicy(\response_{i, t}^{(j)} \mid \problem, \response_{i, < t}^{(j)})} - \log \frac{\refmodel(\response_{i, t}^{(j)} \mid \problem, \response_{i, < t}^{(j)})}{\currentpolicy(\response_{i, t}^{(j)} \mid \problem, \response_{i, < t}^{(j)})} - 1$. The questions $\problem^{(j)}$ included in the batch are filtered according to certain criteria --- details of the filtering are discussed below. Following \citet{yu2025dapoopensourcellmreinforcement}, we set $\epshigh = 0.28$ and $\epslow = 0.2$.

\paragraph{CISPO.} Partway through RL training, we change the objective --- instead of the GRPO/DAPO objective, we use the objective of CISPO, introduced by \citet{minimax2025minimaxm1scalingtesttimecompute} and further found to work well by \citet{khatri2025artscalingreinforcementlearning}. For a mini-batch with questions $\problem^{(j)}$ for $j = 1, \ldots, K$, with each question $\problem^{(j)}$ having responses $\response_i^{(j)}$ for $i = 1,\ldots, \groupsize$, the objective is now
\begin{align*}
\frac{1}{\sum_{j = 1}^K \sum_{i = 1}^{\groupsize} |\response_i^{(j)}|} \sum_{j = 1}^K \sum_{i = 1}^{\groupsize} \sum_{t = 1}^{|\response_i^{(j)}|} \Big(\min(r_{j, i, t}(\theta), \epshigh) \normadv_i^{(j)} \log \currentpolicy(\response_{i, t}^{(j)} \mid \problem, \response_{i, < t}^{(j)}) - \beta \klpenalty(\currentpolicy \mid \refmodel) \Big)
\end{align*}
Observe that if only a single optimization step is performed for each rollout batch, then this objective reduces to REINFORCE \citep{reinforce1992}, due to being fully on-policy --- we perform 2 optimization steps per rollout batch. We continue computing the normalized advantages $\normadv$ by taking the standard deviation of the rewards across all responses for a problem $\problem$. This is different from \citet{khatri2025artscalingreinforcementlearning}, who find that normalizing the advantages across the entire mini-batch performs slightly better.

\clearpage
\section{Plateaus in Accuracy --- Stage 1 and Stage 3} \label{appendix:intermediate_plateaus_rl}

\begin{figure}[h]
    \centering
    \includegraphics[width=0.8\textwidth]{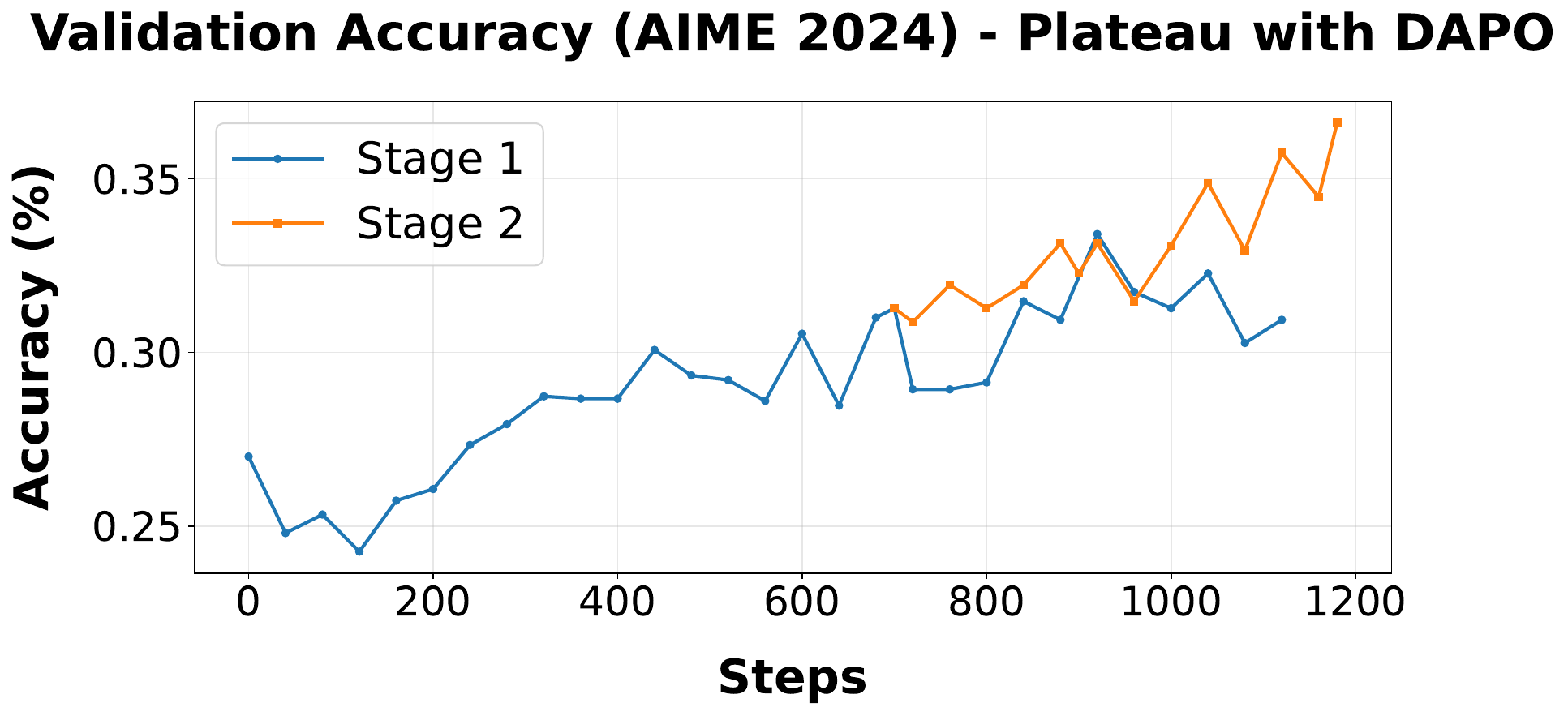}
    \caption{DAPO plateaus in accuracy during Stage 1, while accuracy continues improving monotonically once we switch to CISPO.}
    \label{fig:dapo_stage_1_plateau}
\end{figure}

\begin{figure}[h]
    \centering
    \includegraphics[width=0.8\textwidth]{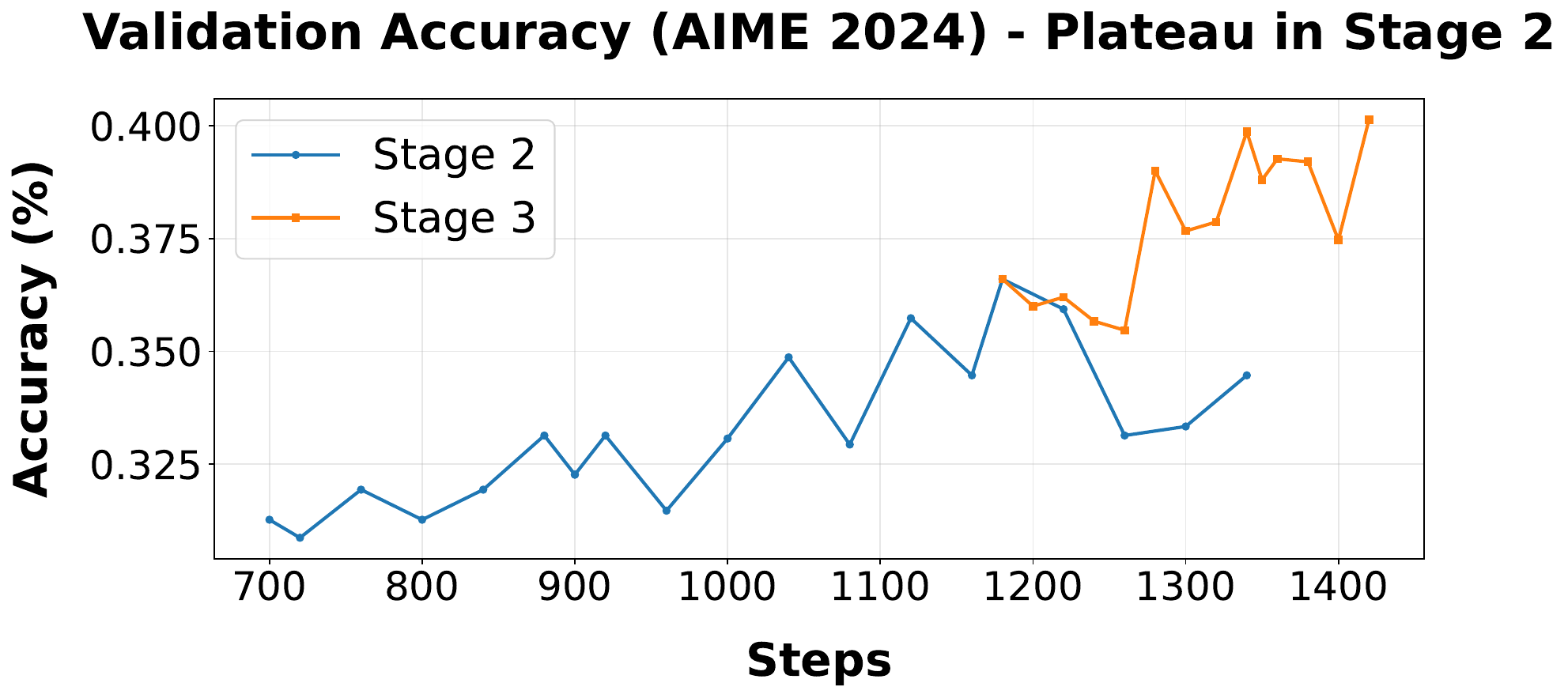}
    \caption{We encounter a plateau in accuracy in Stage 2, where we use filtering strategy $\firstfilter$. Accuracy continues improving monotonically in Stage 3 when we switch to filtering strategy $\secondfilter$.}
    \label{fig:data_filtering_stage_2_plateau}
\end{figure}

\clearpage
\section{Entropy Statistics During Stage 1} \label{appendix:entropy_statistics}

\begin{figure}[h]
    \centering
    \includegraphics[width=0.45\textwidth]{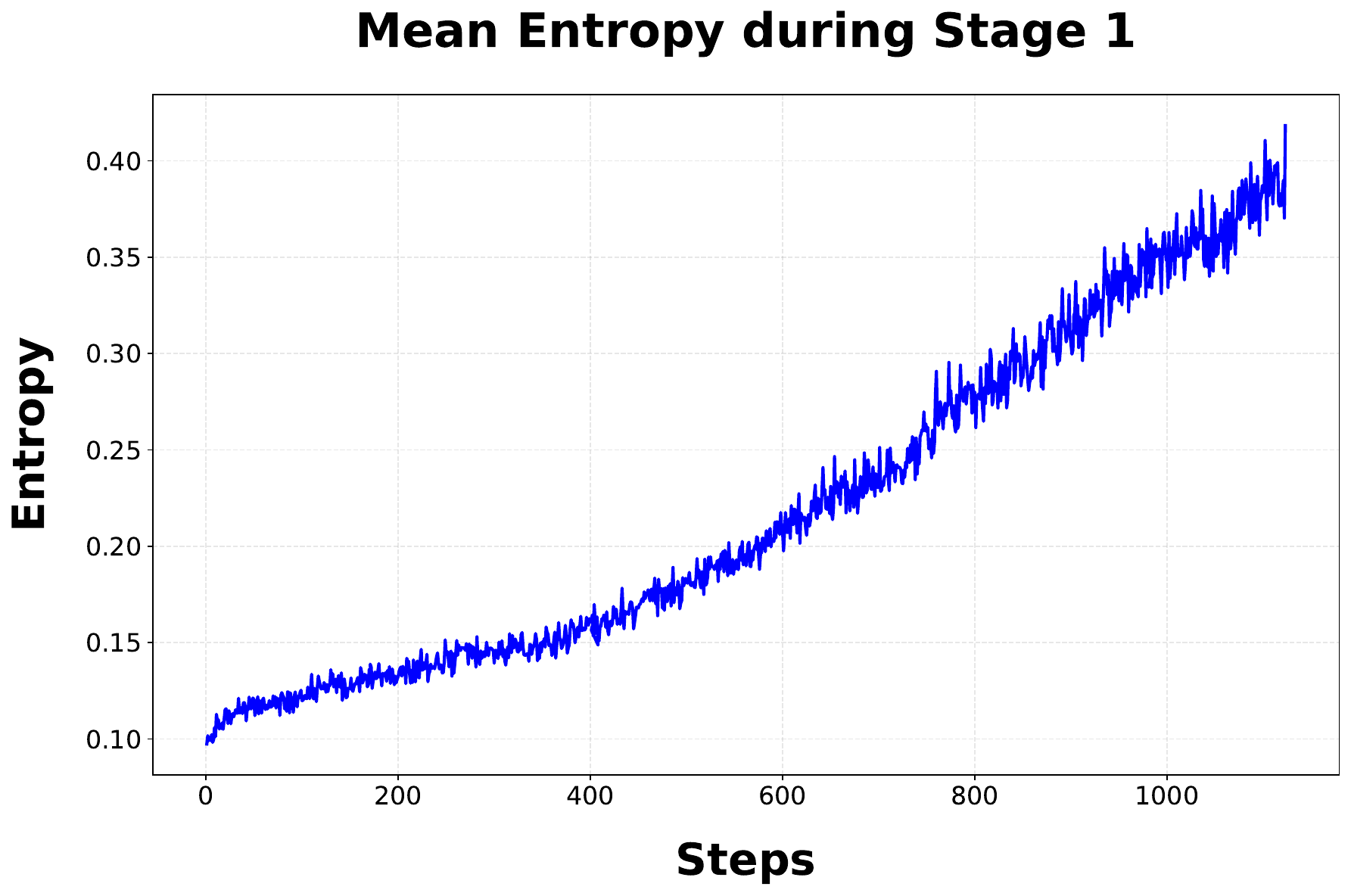}
    \hspace{1cm}
    \includegraphics[width=0.45\textwidth]{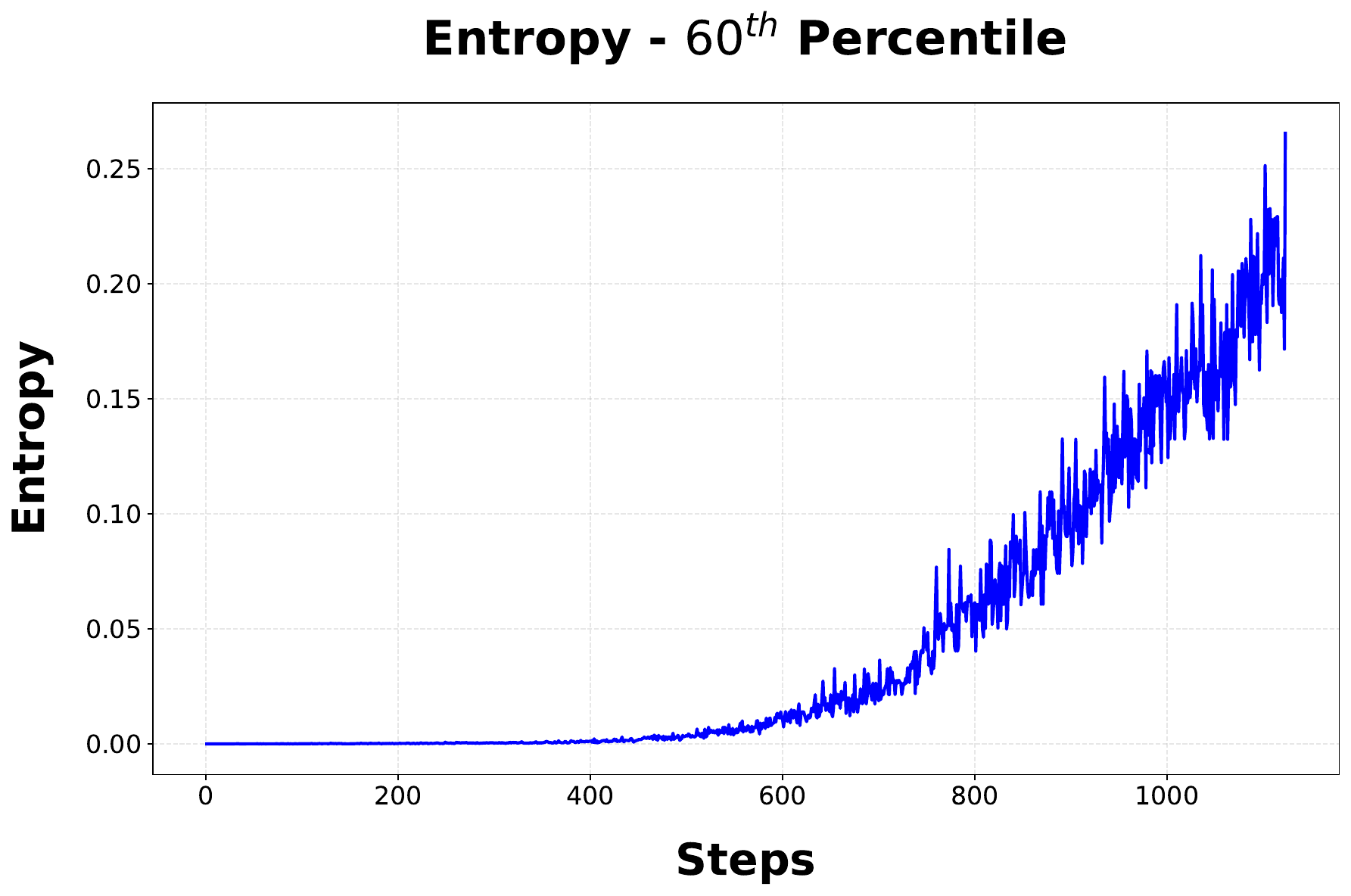}
    \vspace{0.5cm}
    \includegraphics[width=0.45\textwidth]{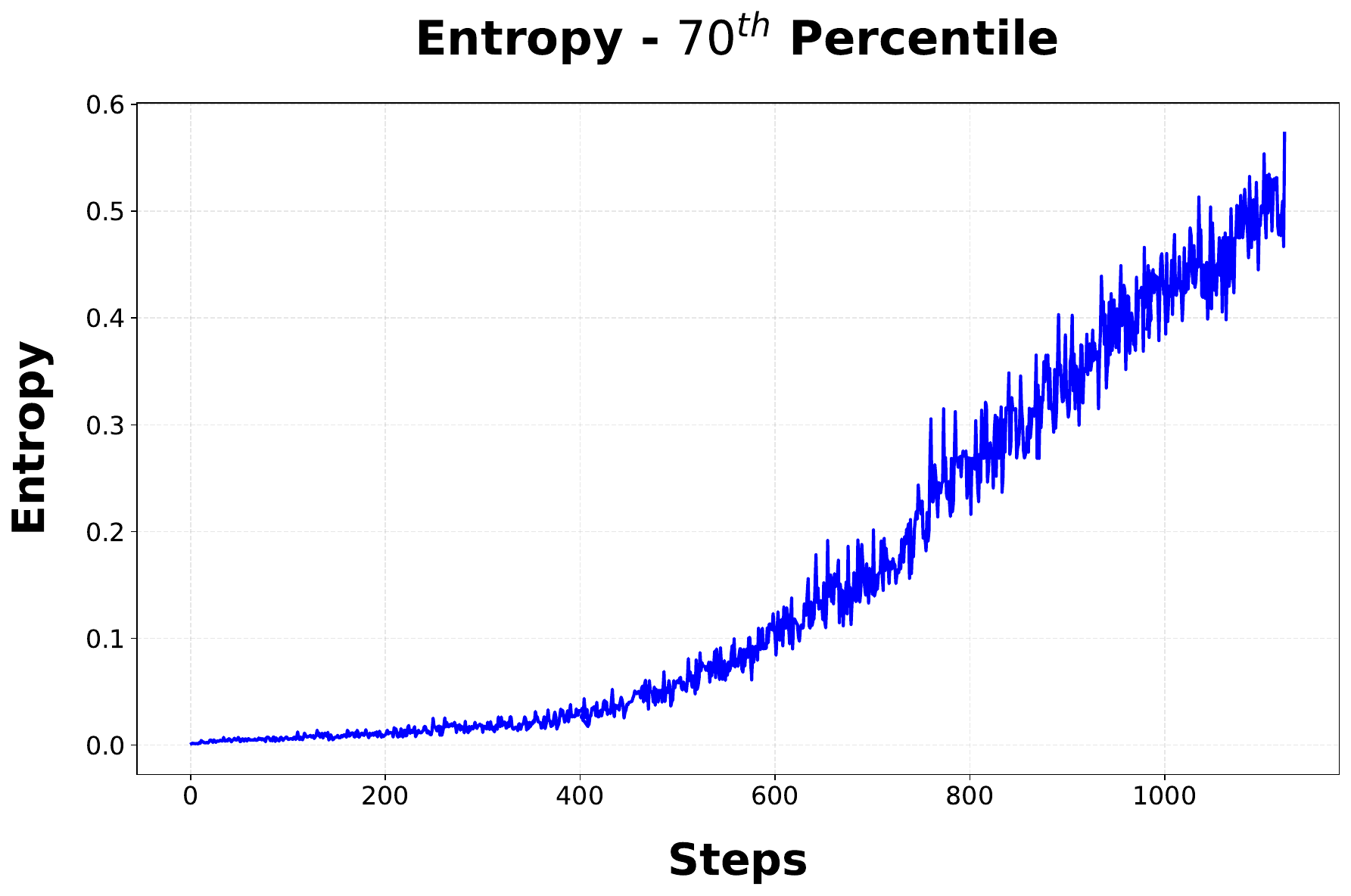}
    \hspace{1cm}
    \includegraphics[width=0.45\textwidth]{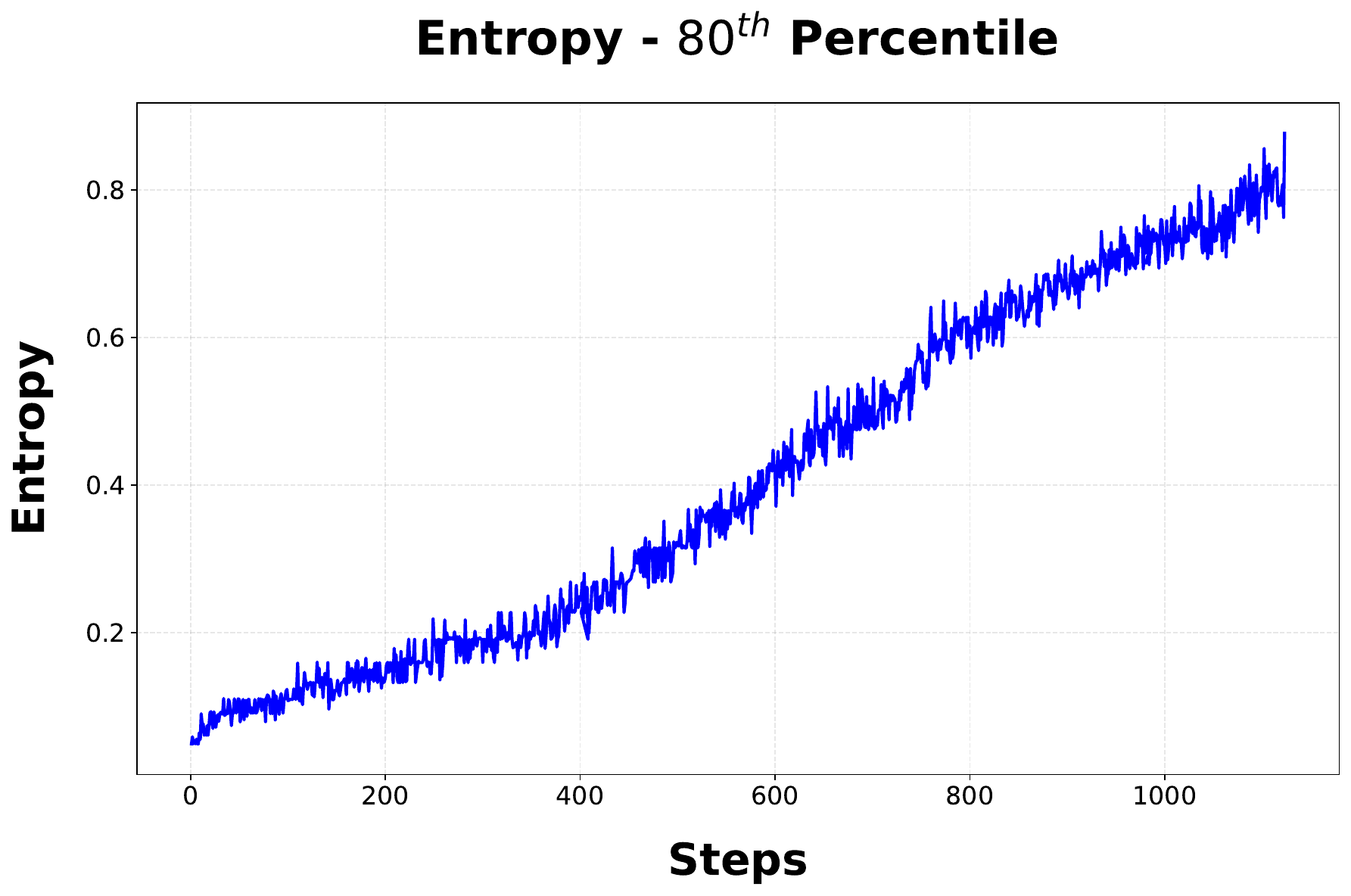}
    \vspace{0.5cm}
    \includegraphics[width=0.45\textwidth]{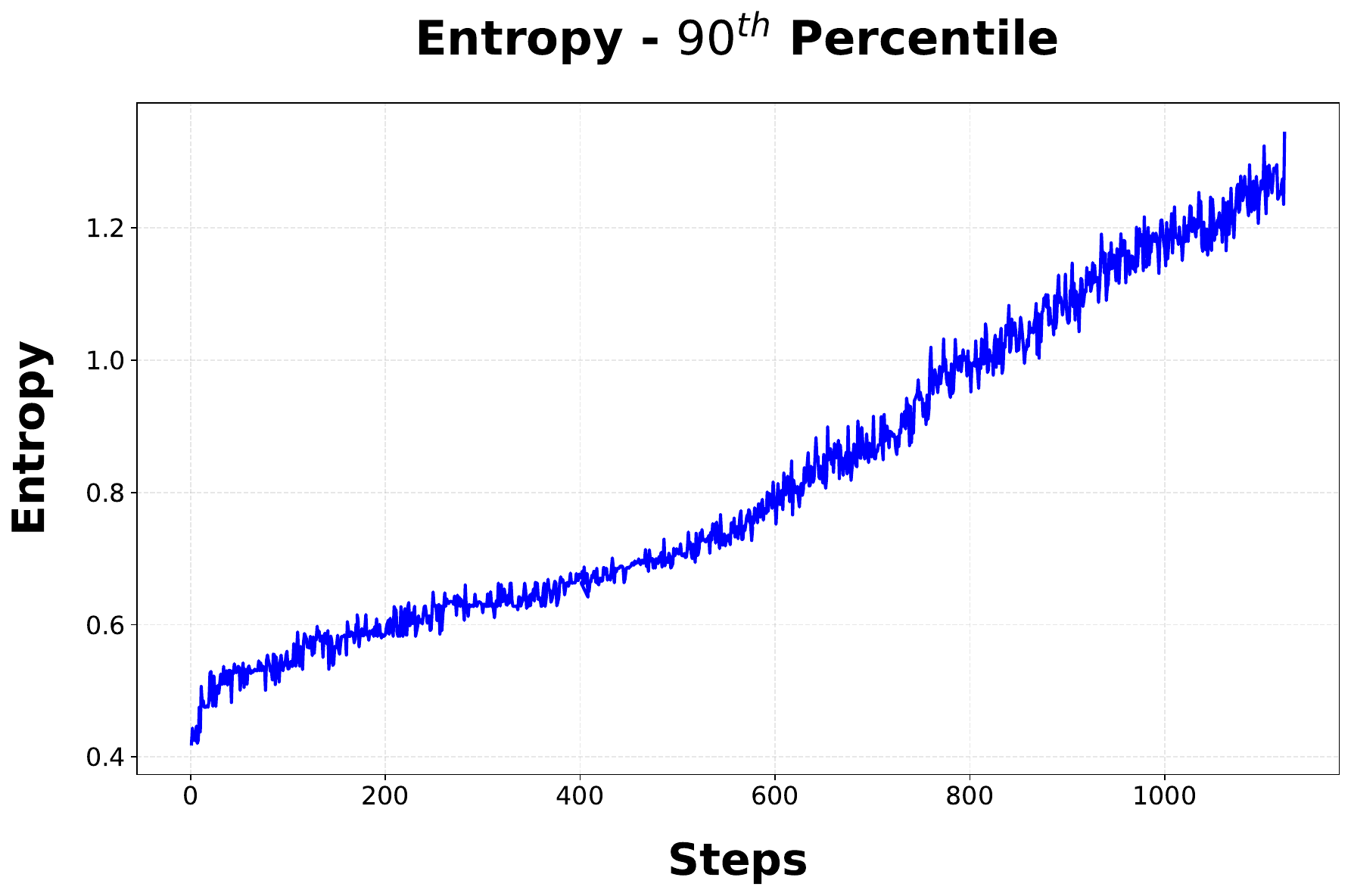}
    \hspace{1cm}
    \includegraphics[width=0.45\textwidth]{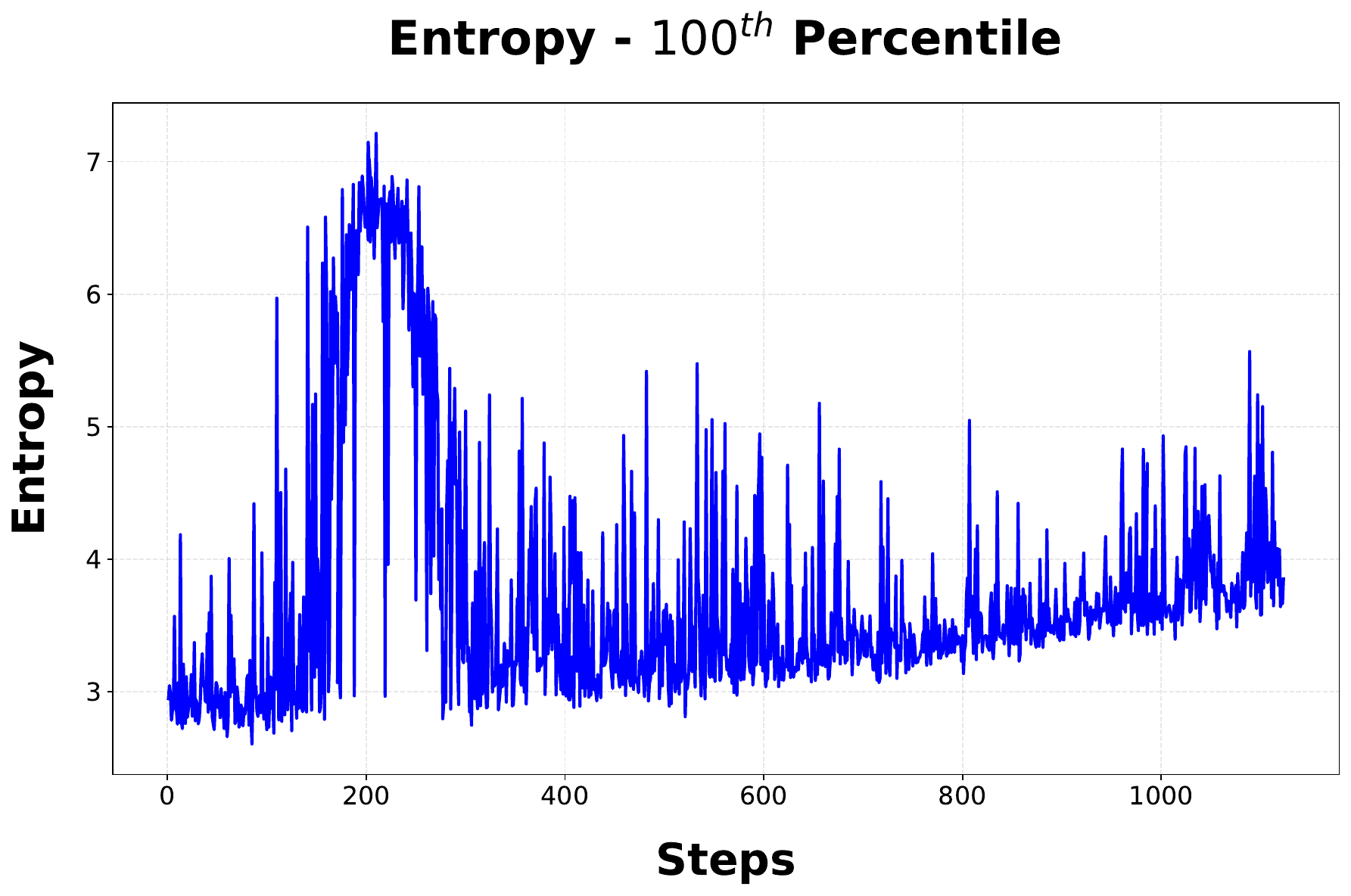}
    \caption{Entropy statistics during stage 1, when using DAPO with $\epshigh=0.28$.}
    \label{fig:entropy_statistics_stage_1}
\end{figure}

\clearpage
\section{Additional Experimental Details} \label{appendix:extra_hyperparameters_details}

In this section, we specify hyperparameters and experimental details that are omitted from \Cref{subsec:experimental_setup}.

\paragraph{\ourmethod RL hyperparameters} The hyperparameters that we use in each stage of our training run are as follows. In Stage 1, we use learning rate $7.071 \cdot 10^{-7}$, rollout batch size $288$, and training batch size $96$. In Stage 2, for the first $200$ steps, we use learning rate $1.414 \cdot 10^{-6}$, rollout batch size $288$, and training batch size $96$. Afterwards, we increase the training batch size to $192$, and the learning rate to $2.828 \cdot 10^{-6}$. Stage 2 lasts for $480$ steps in total. In Stage 3, the rollout batch size is $288$, the training batch size is $96$, and the learning rate is $2 \cdot 10^{-6}$. Finally, in Stage 4, the rollout batch size is $360$, the training batch size is $96$, and the learning rate is $2 \cdot 10^{-6}$. For each rollout batch, we perform $2$ optimization steps, making our algorithm off-policy. In each stage that uses CISPO, we set $\epshigh = 5.0$. For our high length penalty (HLP) stage, we begin from the final step of Stage 4. The learning rate is $2 \cdot 10^{-6}$, the rollout batch size is $360$, and the training batch size is $96$.

\paragraph{HLP with DeepScaleR-1.5B-Preview} For \textbf{DSR-HLP-24K} and \textbf{DSR-HLP-12K}, we use learning rate $2 \cdot 10^{-6}$, training batch size $96$ and rollout batch size $360$. We filter problems $\problem$ based solely on the correctness of the final answer of the responses to $\problem$ (i.e. using filtering strategy \secondfilter), as in \textbf{\ourmethod-HLP}. We use a length penalty with $\lpcoeff = 0.9$ and $\penaltycutoff = 2000$. As in the original training of DeepScaleR-1.5B-Preview \citep{deepscaler2025}, we assign reward $1.0$ to responses which are correct, which include a thinking process between \texttt{<think>} and \texttt{</think>} tags, and for which the answer comes after \texttt{</think>}, and $0$ otherwise. We ensure that \texttt{<think>} is included at the end of the prompt. We run \textbf{\ourmethod-HLP}, \textbf{DSR-HLP-24K}, and \textbf{DSR-HLP-12K} for $200$ steps each, and for each method, we select the best checkpoint by assessing the accuracy and longest path length on AIME 2024. We train and evaluate \textbf{DSR-HLP-24K} and \textbf{DSR-HLP-12K} using the prompt shown in \Cref{box:deepscaler_prompt}.

\paragraph{Majority Voting Baseline --- Implementation Details} As mentioned in \Cref{subsec:experimental_setup}, we generate $N$ responses per problem, where $N$ is $200$ or $16$ depending on the evaluation benchmark, using \textbf{DSR-12K}. To determine the pass rate and longest path length of \textbf{Maj-DSR-3} for a given problem, we take all $\binom{N}{3}$ subsets $S = \{\response_1, \response_2, \response_3\}$ of size $3$ from the responses to this problem, and compute the accuracy and longest path length when using $S$ for majority voting. For a particular subset $S$, we compute the accuracy and longest path length as follows. We first use \texttt{math\_verify} to determine the number of equivalence classes of final answers. If there is only a single equivalence class, we compute the accuracy of each of the $3$ responses and take the average (corresponding to sampling a representative uniformly at random). If there are two equivalence classes, one of which has $2$ elements, we use the equivalence class with $2$ elements, and take the average of the accuracies of those two elements. If there are $3$ equivalence classes, then we again take the average accuracy of all $3$ responses (corresponding to sampling a class uniformly at random). In all cases, the longest path length of majority voting is calculated as the maximum longest path length out of these $3$ responses.

\clearpage
\section{Additional Tables and Plots} \label{appendix:additional_tables_plots}

\begin{figure}[h]
    \centering
    \includegraphics[width=0.8\textwidth]{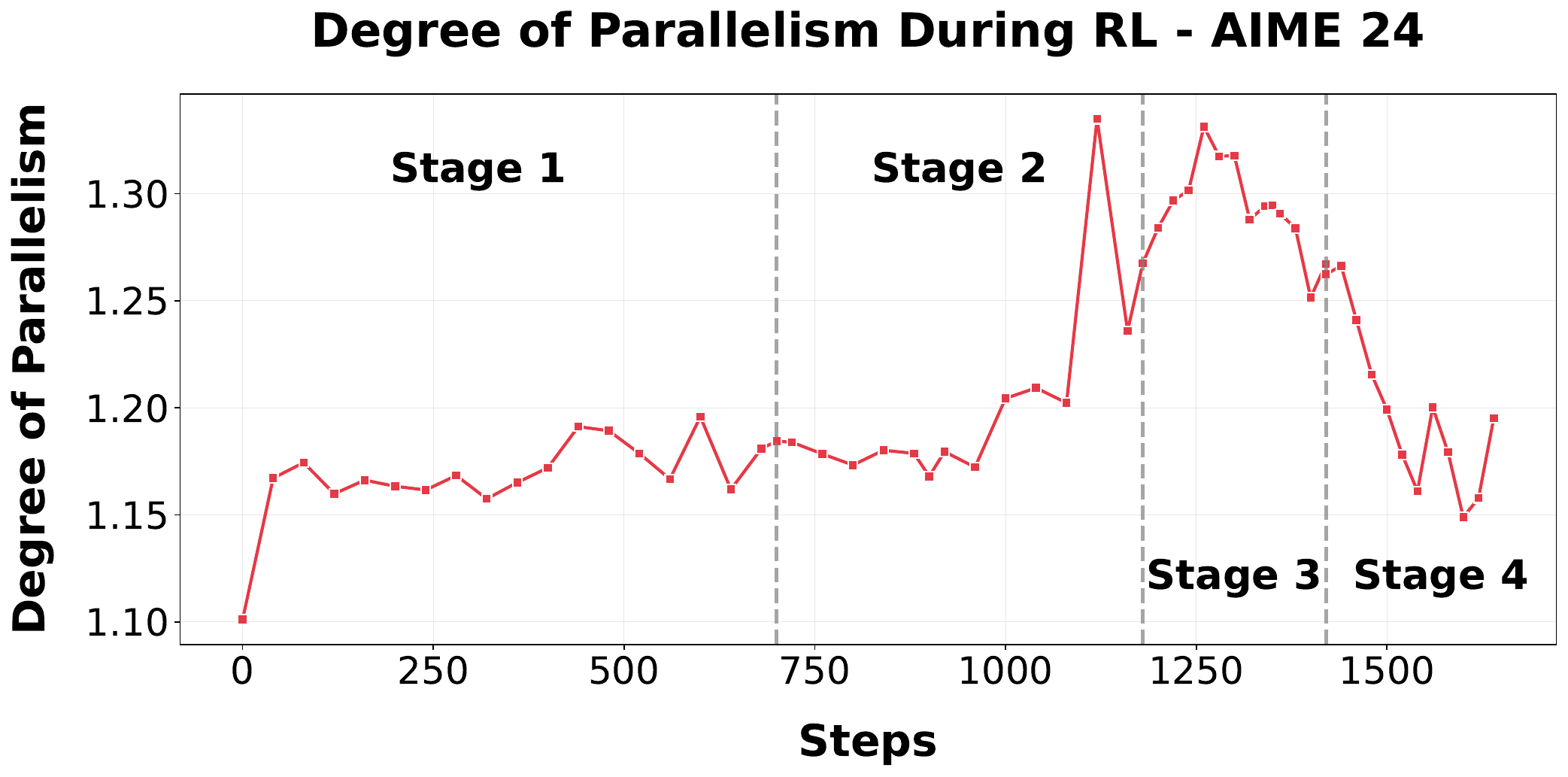}
    \caption{Degree of parallelism of \textbf{\ourmethod} on AIME 2024 during all stages of RL.}
    \label{fig:validation_dp}
\end{figure}

\begin{figure}[h]
    \centering
    \includegraphics[width=0.8\textwidth]{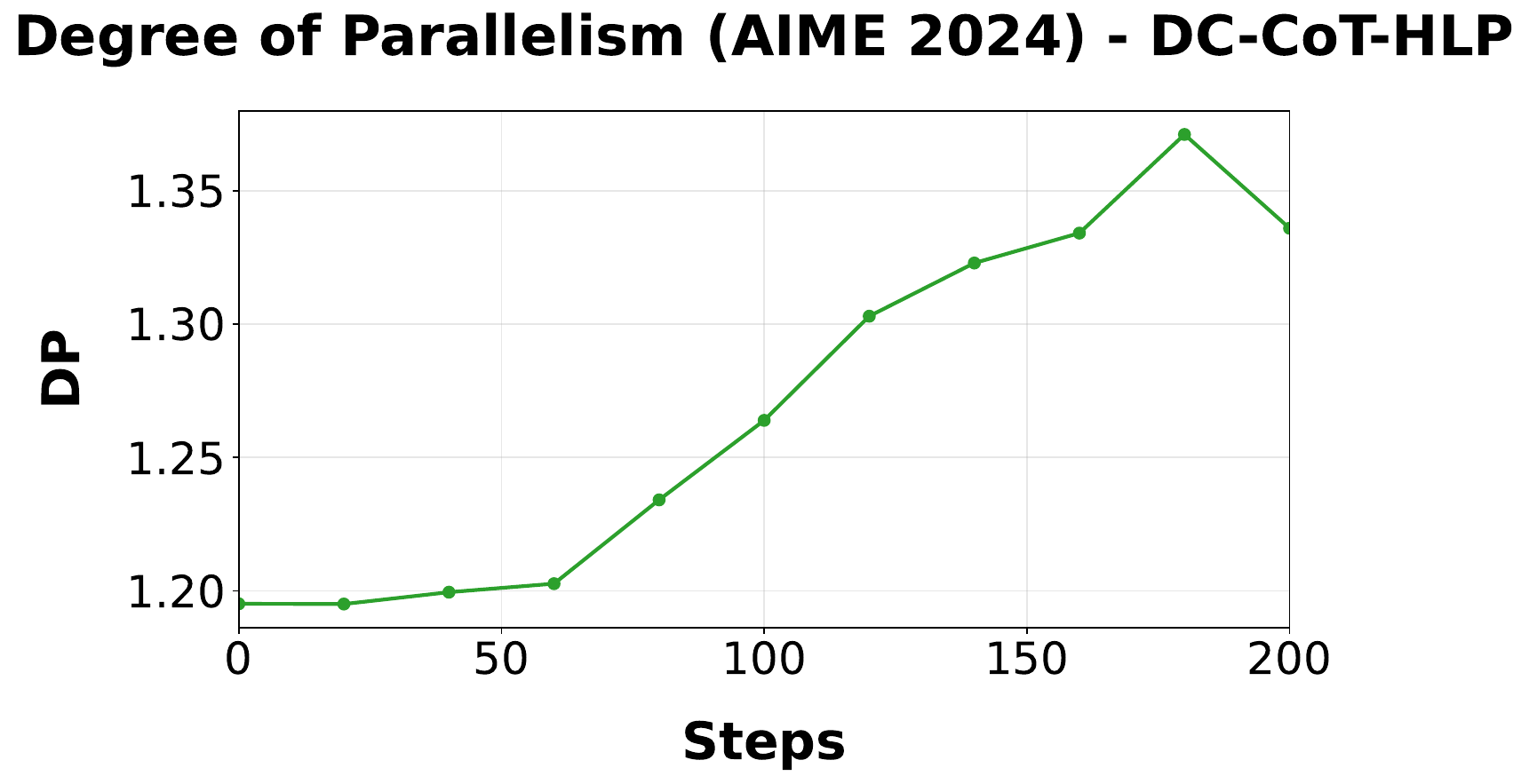}
    \caption{Degree of parallelism on AIME 2024 for \textcolor{green}{\ourmethod-HLP}}
    \label{fig:hlp_dp}
\end{figure}

Here we include tables and plots that are omitted from \Cref{subsec:results}. \Cref{tab:main_results_aime_amc_hmmt} and \Cref{tab:main_results_math_mm_ob} show the sample standard deviation for the accuracies and longest path lengths reported in \Cref{tab:main_results}. Different from the longest path length, \Cref{tab:total_tokens} shows the \textit{total amount of tokens} used by each response, averaged across responses. Finally, \Cref{fig:validation_dp} and \Cref{fig:hlp_dp} show the degree of parallelism during our main RL run and during the HLP stage, respectively.

\begin{table}[ht]
\centering
\caption{Results in \Cref{tab:main_results} with sample mean and sample standard deviation (AIME 2024, AMC 23, HMMT 2025).}
\label{tab:main_results_aime_amc_hmmt}

\small
\setlength{\tabcolsep}{3pt}
\renewcommand{\arraystretch}{1.15}

\begin{tabular}{l *{6}{c}}
\toprule
\textbf{Method}
& \multicolumn{2}{c}{\textbf{AIME 24}}
& \multicolumn{2}{c}{\textbf{AMC}}
& \multicolumn{2}{c}{\textbf{HMMT}} \\
\cmidrule(lr){2-3}\cmidrule(lr){4-5}\cmidrule(lr){6-7}
& \textbf{Acc (\%)} & \textbf{LPL}
& \textbf{Acc (\%)} & \textbf{LPL}
& \textbf{Acc (\%)} & \textbf{LPL} \\
\midrule

\textbf{DSR-32K}
& $40.20_{\pm 0.63}$ & $9540_{\pm 83}$
& $81.35_{\pm 0.44}$ & $5151_{\pm 51}$
& $18.28_{\pm 0.50}$ & $10685_{\pm 89}$ \\

\textbf{DSR-12K}
& $39.43_{\pm 0.63}$ & $8047_{\pm 43}$
& $80.66_{\pm 0.44}$ & $4792_{\pm 38}$
& $18.28_{\pm 0.50}$ & $8666_{\pm 41}$ \\

\textbf{Maj-DSR}
& $44.19_{\pm 0.01}$ & $9690_{\pm 0.0}$
& $84.55_{\pm 0.0}$ & $5803_{\pm 1}$
& $19.39_{\pm 0.01}$ & $9970_{\pm 0.0}$ \\

\textbf{DSR-HLP-24K}
& $40.08_{\pm 0.63}$ & $7903_{\pm 54}$
& $81.90_{\pm 0.43}$ & $4535_{\pm 40}$
& $18.32_{\pm 0.50}$ & $8375_{\pm 55}$ \\

\textbf{DSR-HLP-12K}
& $39.03_{\pm 0.63}$ & $6023_{\pm 38}$
& $80.83_{\pm 0.44}$ & $3574_{\pm 29}$
& $16.78_{\pm 0.48}$ & $6391_{\pm 38}$ \\
\midrule

\textbf{\ourmethod}
& $41.87_{\pm 0.64}$ & $5974_{\pm 38}$
& $80.38_{\pm 0.44}$ & $3368_{\pm 29}$
& $18.07_{\pm 0.50}$ & $6803_{\pm 38}$ \\

\textbf{\ourmethod-HLP}
& $40.85_{\pm 0.63}$ & $4740_{\pm 27}$
& $80.55_{\pm 0.44}$ & $2786_{\pm 21}$
& $16.48_{\pm 0.48}$ & $5175_{\pm 28}$ \\

\textbf{\ourmethod-Maj}
& $46.02_{\pm 0.01}$ & $7412_{\pm 0.0}$
& $83.70_{\pm 0.00}$ & $4239_{\pm 0.43}$
& $19.54_{\pm 0.01}$ & $8109_{\pm 0.0}$ \\

\bottomrule
\end{tabular}
\end{table}

\begin{table}[h]
\centering
\caption{Sample mean and sample standard deviation for \Cref{tab:main_results} (MATH 500, Minerva Math, Olympiad Bench).}
\label{tab:main_results_math_mm_ob}

\small
\setlength{\tabcolsep}{3pt}
\renewcommand{\arraystretch}{1.15}

\begin{tabular}{l *{6}{c}}
\toprule
\textbf{Method}
& \multicolumn{2}{c}{\textbf{MATH}}
& \multicolumn{2}{c}{\textbf{MM}}
& \multicolumn{2}{c}{\textbf{OB}} \\
\cmidrule(lr){2-3}\cmidrule(lr){4-5}\cmidrule(lr){6-7}
& \textbf{Acc (\%)} & \textbf{LPL}
& \textbf{Acc (\%)} & \textbf{LPL}
& \textbf{Acc (\%)} & \textbf{LPL} \\
\midrule

\textbf{DSR-32K}
& $89.04_{\pm 0.35}$ & $3146_{\pm 39}$
& $36.65_{\pm 0.73}$ & $4879_{\pm 71}$
& $60.94_{\pm 0.47}$ & $5875_{\pm 49}$ \\

\textbf{DSR-12K}
& $88.81_{\pm 0.35}$ & $2963_{\pm 29}$
& $36.63_{\pm 0.73}$ & $4496_{\pm 47}$
& $60.52_{\pm 0.47}$ & $5341_{\pm 33}$ \\

\textbf{Maj-DSR}
& $91.02_{\pm 0.05}$ & $3638_{\pm 6}$
& $38.69_{\pm 0.12}$ & $5716_{\pm 9}$
& $63.34_{\pm 0.08}$ & $6479_{\pm 6}$ \\

\textbf{DSR-HLP-24K}
& $89.29_{\pm 0.35}$ & $2790_{\pm 30}$
& $37.50_{\pm 0.73}$ & $4124_{\pm 49}$
& $61.72_{\pm 0.47}$ & $4909_{\pm 34}$ \\

\textbf{DSR-HLP-12K}
& $88.08_{\pm 0.36}$ & $2215_{\pm 22}$
& $37.36_{\pm 0.73}$ & $3122_{\pm 33}$
& $60.34_{\pm 0.47}$ & $3880_{\pm 25}$ \\
\midrule

\textbf{\ourmethod}
& $88.74_{\pm 0.35}$ & $1979_{\pm 22}$
& $35.16_{\pm 0.72}$ & $2865_{\pm 32}$
& $60.19_{\pm 0.47}$ & $3742_{\pm 26}$ \\

\textbf{\ourmethod-HLP}
& $88.44_{\pm 0.36}$ & $1734_{\pm 16}$
& $35.02_{\pm 0.72}$ & $2454_{\pm 25}$
& $60.07_{\pm 0.47}$ & $3040_{\pm 19}$ \\

\textbf{\ourmethod-Maj}
& $90.66_{\pm 0.05}$ & $2443_{\pm 5}$
& $37.05_{\pm 0.12}$ & $3668_{\pm 6}$
& $62.45_{\pm 0.08}$ & $4603_{\pm 5}$ \\

\bottomrule
\end{tabular}
\end{table}

\begin{table*}[h]
\centering
\caption{Total tokens used by each response, averaged across responses. For \textbf{\ourmethod-Maj}, we average across all subsets of $3$ responses to the same problem. For each subset of $3$ responses, we compute the sum of the tokens used by each of the $3$ responses.}
\label{tab:total_tokens}

\small
\setlength{\tabcolsep}{3pt}
\renewcommand{\arraystretch}{1.15}

\begin{tabular}{l *{6}{c}}
\toprule
\textbf{Method}
& \textbf{AIME 24}
& \textbf{AMC}
& \textbf{HMMT}
& \textbf{MATH}
& \textbf{MM}
& \textbf{OB} \\
\midrule

\textbf{\ourmethod} & $6831$ & $4185$ & $7603$ & $2554$ & $3526$ & $4489$ \\

\textbf{\ourmethod-HLP} & $6185$ & $4022$ & $6411$ & $2537$ & $3401$ & $4154$ \\

\textbf{\ourmethod-Maj} & $20493$ & $12554$ & $22809$ & $7662$ & $10579$ & $13468$ \\

\bottomrule
\end{tabular}
\end{table*}

\clearpage
\section{Simplifications of RL Algorithm with Fewer Stages} \label{appendix:fewer_stage_RL}

Our main RL run uses 4 stages --- we explore the following two simplified versions.
\begin{enumerate}
    \item Starting from the SFT checkpoint, we skip stage 1, and directly start using CISPO with the same hyperparameters as in stage 2. We then switch to the hyperparameters of stage 3 when the accuracy plateaus, followed by those of stage 4. Thus, this version of the RL algorithm does not use DAPO at any point.
    \item Starting from the SFT checkpoint, we skip stages 1 and 2, applying CISPO with the hyperparameters of stage 3, followed by those of stage 4 when the amount of truncated responses starts to increase. Thus, we never use the filtering strategy \firstfilter, instead using \secondfilter throughout the run.
\end{enumerate}
We refer to these simplified procedures as the \textit{three-stage algorithm} and \textit{two-stage algorithm} respectively.

\begin{figure}[h]
    \centering
    \includegraphics[width=\linewidth]{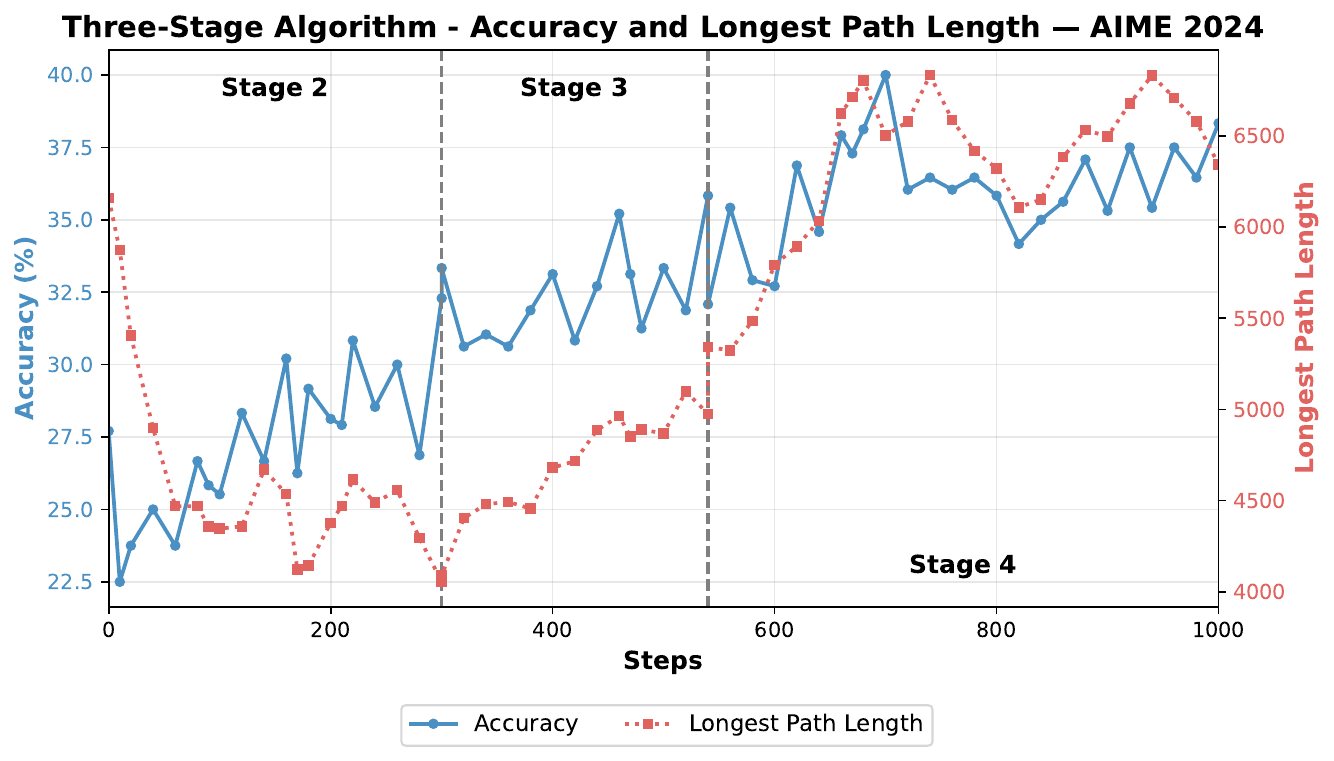}
    \caption{Validation accuracy and longest path length over the course of the three-stage algorithm. We write "stage 2" to indicate that the hyperparameters from stage 2 of our original run are being used, and similarly for the other stages.}
    \label{fig:three_stage_acc_lpl}
\end{figure}

\begin{figure}[h]
    \centering
    \includegraphics[width=\linewidth]{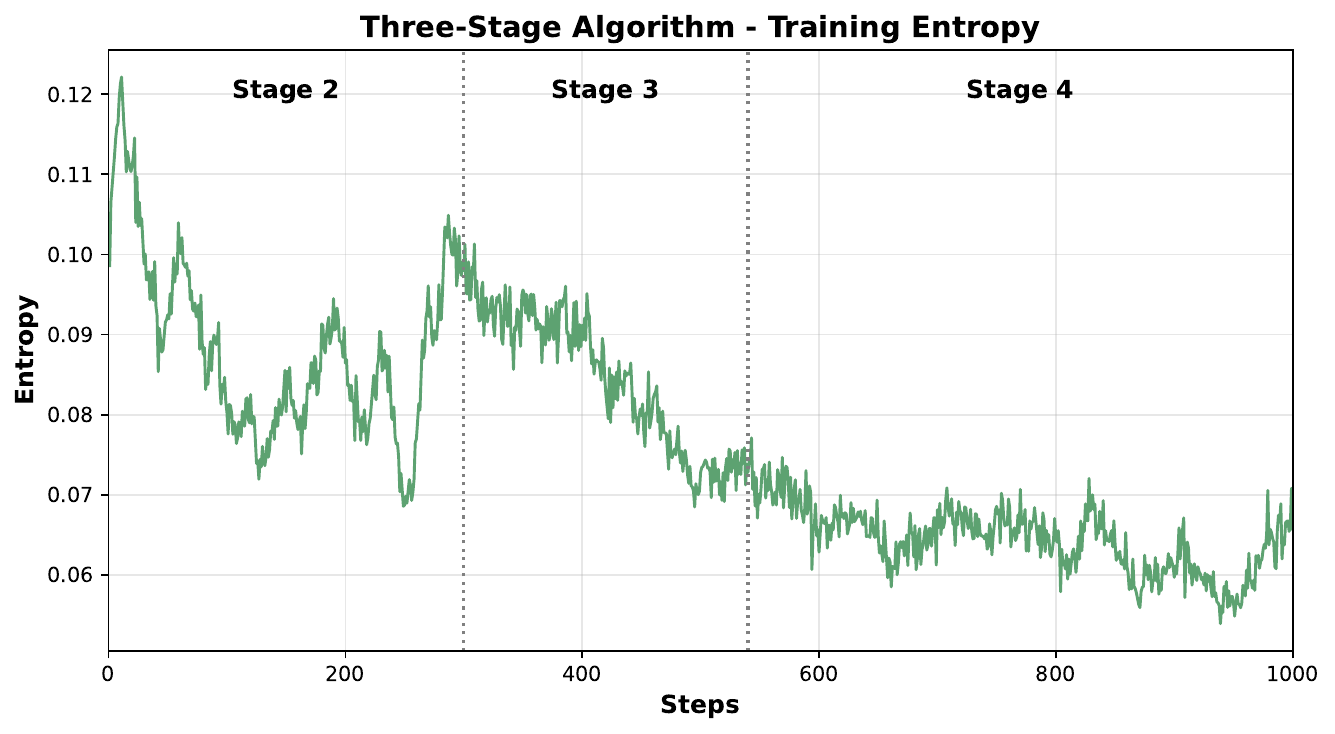}
    \caption{Training entropy over the course of the three-stage algorithm.}
    \label{fig:three_stage_entropy}
\end{figure}

For the three-stage algorithm, validation accuracy and longest path length are shown in \Cref{fig:three_stage_acc_lpl}, while the average token entropy for the generated responses on the training dataset is shown in \Cref{fig:three_stage_entropy}. While the accuracy improves monotonically for most of the run, it plateaus slightly below 40\% during Stage 4. We believe this is due to a very low entropy --- as seen in \cref{fig:three_stage_entropy}, the entropy decreases significantly compared to the SFT checkpoint. Thus, it is possible the initial increase in entropy in our main run, due to performing stage 1 with DAPO (see \Cref{fig:entropy_statistics_stage_1}), brings the entropy into a more beneficial regime by the end of stage 1.

\begin{figure}[h]
    \centering
    \includegraphics[width=\linewidth]{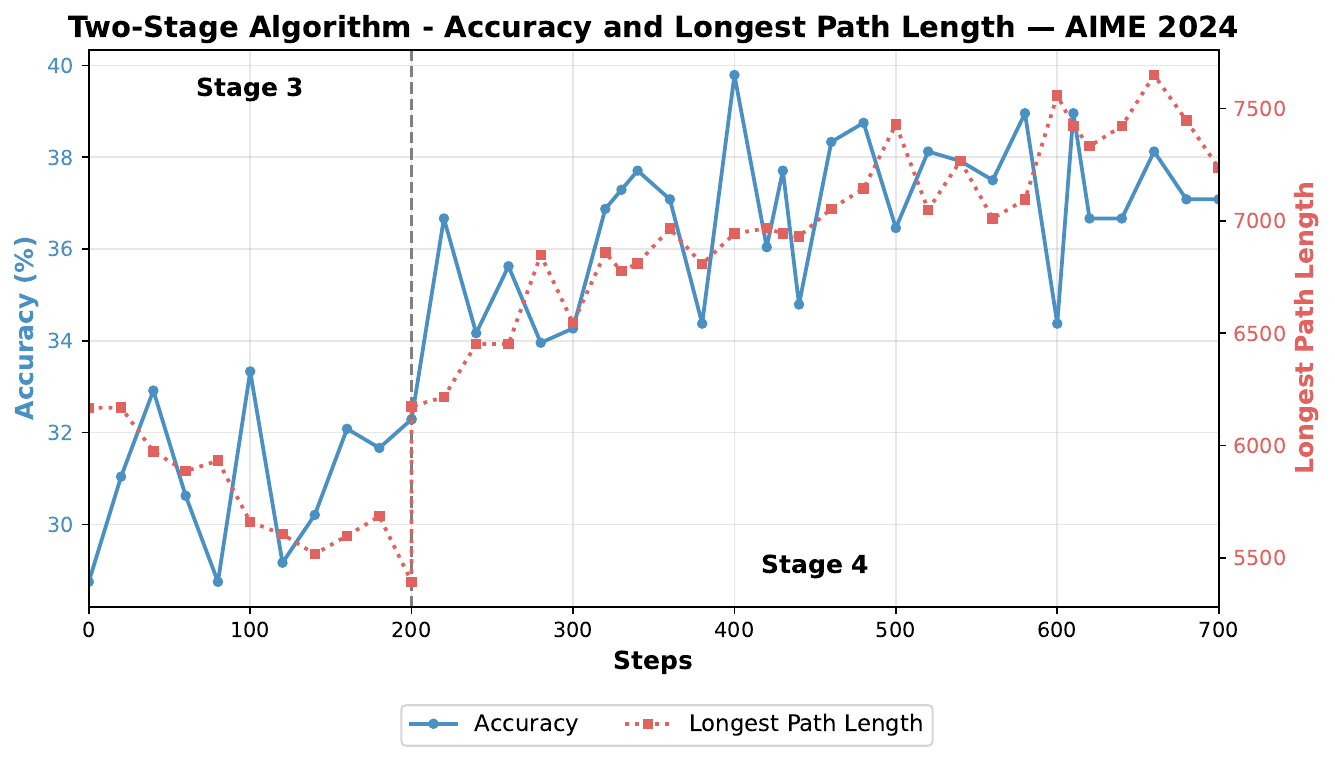}
    \caption{Validation accuracy and longest path length over the course of the two-stage algorithm. We write "stage 3" to indicate that the hyperparameters from stage 3 of our original run are being used, and similarly for "stage 4".}
    \label{fig:two_stage_acc_lpl}
\end{figure}

For the two-stage algorithm, validation accuracy and longest path length are shown in \Cref{fig:two_stage_acc_lpl}. Since we never use filtering strategy \firstfilter, the decrease in longest path length during stage 3, compared to the SFT checkpoint, is not very significant, compared to the decrease in longest path length during stages 1 and 2 in our main run, as well as the decrease in longest path length during stage 2 of the three-stage algorithm. This suggests that our initial use of \firstfilter in our main run plays an important role in its final performance.

For \Cref{fig:three_stage_acc_lpl} and \Cref{fig:two_stage_acc_lpl}, we report the average accuracy and longest path length across $16$ rollouts for each problem.

\clearpage
\section{Latency in Wall-Clock Time} \label{appendix:latency_table}

\paragraph{Inference with batch size 1.}

We evaluate the latency of each method on AIME 2024, averaged across problems, where we perform inference with batch size 1. For each problem, we use 3 different random seeds. We then average across problems and across random seeds. In all methods, generation is done using vLLM \citep{kwon2023efficient}. The results are shown in \Cref{tab:aime_2024_batch_size_1_latency_result}. Note that \textbf{\ourmethod} achieves substantial improvements in latency compared to \textbf{DSR-32K} and \textbf{DSR-12K}. Even compared to \textbf{DSR-12K}, which has the same maximum allowed longest path length as \textbf{\ourmethod}, \textbf{\ourmethod} shows a 25\% improvement in latency. Similarly, \textbf{DC-CoT-HLP} achieves a substantial improvement in latency compared to \textbf{DSR-HLP-24K} and \textbf{DSR-HLP-12K}, while achieving slightly higher accuracy than \textbf{DSR-HLP-24K} on AIME 2024.

\begin{table}[h]
\centering
\caption{Latency for generation with batch size 1 on AIME 2024.}
\label{tab:aime_2024_batch_size_1_latency_result}

\small
\setlength{\tabcolsep}{6pt}
\renewcommand{\arraystretch}{1.15}

\begin{tabular}{lcc}
\toprule
\textbf{Method} & \textbf{Avg time per Problem (s)} & \textbf{Std.\ Dev across seeds (s)} \\
\midrule
\textbf{DSR-32K}      & 84.33 & 31.19 \\
\textbf{DSR-12K}      & 63.74 &  2.29 \\
\textbf{DC-CoT}       & 47.51 &  5.69 \\
\textbf{DSR-HLP-24K}  & 61.53 &  5.68 \\
\textbf{DSR-HLP-12K}  & 48.94 &  1.56 \\
\textbf{DC-CoT-HLP}   & 35.49 &  1.68 \\
\bottomrule
\end{tabular}
\end{table}

\paragraph{Inference with larger batch sizes.} In \Cref{tab:aime_2024_larger_batch_size_latency}, we evaluate the latency of each method when different batch sizes are used. We evaluate each method with a batch size of $30 N$ rollouts on AIME 2024, where each problem is repeated $N$ times. Each run uses a single H100 GPU. For each method and each value of $N$, we use 3 different random seeds and report the mean and standard deviation across seeds. We find that DC-CoT and DC-CoT-HLP are significantly faster than the corresponding baselines.

\begin{table}[h]
\centering
\caption{Wall-clock time (seconds) for generation with batch size $30 \cdot N$, where $N$ is the number of rollouts per problem, on AIME 2024. Each entry is the mean $\pm$ standard deviation across 3 random seeds.}
\label{tab:aime_2024_larger_batch_size_latency}
\small
\setlength{\tabcolsep}{6pt}
\renewcommand{\arraystretch}{1.15}
\begin{tabular}{lcccc}
\toprule
\textbf{Method} & $N=1$ & $N=4$ & $N=8$ & $N=16$ \\
\midrule
\textbf{DC-CoT}       & 124.4 $\pm$ 2.6 & 289.7 $\pm$ 21.6 & 579.5 $\pm$ 16.6 & 1039.7 $\pm$ 12.4 \\
\textbf{DC-CoT-HLP}   & 90.8 $\pm$ 9.9 & 224.0 $\pm$ 9.3 & 435.2 $\pm$ 43.7 & 911.1 $\pm$ 16.6 \\
\textbf{DSR-12K}      & 152.2 $\pm$ 2.6 & 421.4 $\pm$ 23.9 & 862.4 $\pm$ 14.3 & 1706.5 $\pm$ 55.7 \\
\textbf{DSR-32K}      & 332.9 $\pm$ 93.9 & 725.4 $\pm$ 42.1 & 1499.4 $\pm$ 66.8 & 2756.0 $\pm$ 79.5 \\
\textbf{DSR-HLP-12K}  & 140.6 $\pm$ 3.2 & 307.5 $\pm$ 11.8 & 525.4 $\pm$ 38.7 & 1132.2 $\pm$ 5.3 \\
\textbf{DSR-HLP-24K}  & 208.4 $\pm$ 6.5 & 461.2 $\pm$ 44.8 & 892.7 $\pm$ 80.6 & 1815.0 $\pm$ 50.4 \\
\bottomrule
\end{tabular}
\end{table}

\clearpage
\section{Qualitative Analysis of Rollouts} \label{appendix:decomposability}

In this section, we qualitatively analyze the behavior of the spawned workers.

\paragraph{Setup.} From the last $5$ steps in our main DC-CoT RL run, we select 50 problems that were trained on, and select one correct rollout for each problem. We classify the rollouts, and the workers within them, according to the following criteria:
\begin{itemize}
    \item Are the workers doing distinct work, or identical work? 
    \begin{itemize}
        \item If they are doing distinct work, are they \textbf{dividing} the problem into separate tasks which are later combined, or are they \textbf{cross-verifying} the same result through different approaches?
        \item Within a rollout, we evaluate this separately for every pair of workers.
    \end{itemize}
    \item In the rollout, do two workers reach contradictory conclusions?
    \item Does the director hallucinate or incorrectly use the workers' results?
    \item Regarding the problem, is there a way for the problem to be decomposed into independent tasks that can be combined to find a solution?
    \item Can the problem be solved by multiple distinct approaches?
\end{itemize}

We then evaluate each problem with a separate Claude Sonnet 5 agent --- the prompt is shown in \Cref{appendix:claude_sonnet_5_decomposability_prompt}. We designed and wrote this prompt with the help of Claude Opus 5.

\paragraph{Results.} We find that out of the rollouts, 25/50 had at least two workers that were doing distinct work: 19/50 involved cross-verifications, while 6/50 involved dividing the problem into subtasks. We note that cross-verifications are still nontrivial, and it is typical for cross-verification to use a lot of tokens in standard CoTs. The workers reach contradictory conclusions in 4/50 rollouts, while the director misuses workers' results in 18/50 rollouts.

We also find that problems can be split into independent parts in 10/50 cases, while 36/50 problems admit several different approaches by which they can be solved in parallel.

\clearpage
\section{Prompt for Claude Sonnet 5 to Analyze Rollouts} \label{appendix:claude_sonnet_5_decomposability_prompt}

\begin{mybox}[box:claude_sonnet_5_prompt]{Prompt for Claude Sonnet 5 to Analyze Rollouts}
\begin{Verbatim}
You are analyzing ONE reasoning trajectory ("rollout") produced by a 1.5B math model trained with RL to decompose problems across three parallel workers ("DC-CoT"). Your job is to record specific structural facts about it. Be a careful, skeptical, neutral evaluator: do NOT assume any failure is present, and quote verbatim evidence for every claim.

DO NOT USE ANY TOOLS. Everything you need is inlined below. Return the structured object as your final response.

## How to read the trajectory

Structure:
1. **Director preamble** — reasons about the problem, usually states a plan for each worker.
2. **`<spawn_workers>` … `</spawn_workers>`** — contains `<worker_1>…</worker_1>`, `<worker_2>…`, `<worker_3>…`.
3. **Director synthesis** — continues after the workers return.
4. **`</think>`** then **`<answer>… \boxed{…}</answer>`**.

There may be **more than one round** of `<spawn_workers>`. Answer the per-round questions once per round.

Mechanics you must keep in mind:

- The three workers run **simultaneously and independently**. Each receives exactly the director's text up to `<spawn_workers>` plus its own `<worker_i>` tag — nothing else. **No worker can see any other worker's text**, and none can see the director's later synthesis.
- Therefore two workers producing the same reasoning is **not copying** — it is independent duplicated effort, which is what we are measuring.
- The director **does** see all workers' outputs before its synthesis.
- A worker may end in three ways: (i) emitting its `</worker_i>` tag; (ii) **stalling** — closing the tag having trailed off without a result; (iii) **self-terminating** with no close tag, often right after hallucinating a full `</think><answer>…</answer>` block. None of these is a failure by itself — record it.
- **The final answer in this rollout is already known to be CORRECT.** Judge the PROCESS, not the outcome. Do not solve the problem yourself and do not hunt for a wrong final answer.

## THE PROBLEM

{PROBLEM}

## THE TRAJECTORY

{TRAJECTORY}

---

# What to record

## Step 1 — per worker, per round

Two **independent** properties, never merged:

- `work_level`: `concluded` (reached a substantive result) | `partial` (real work, no result reached) | `minimal` (restates the director's setup, one sentence, or nothing)
- `termination`: `closed_tag` | `no_close_tag`
- `cut_off`: `0` or `1`. **This should be impossible.** A worker that hits its token cap consumes the entire remaining budget, so the director never runs and no answer is produced — such a rollout could not be in this sample. Set `1` only if you have direct evidence, and explain in `notes`.
- `task`: one line on what the worker actually **did**.

## Step 2 — classify every PAIR of workers into exactly one relation

- **`IDENTICAL`** — the same computation; a mistake in one would recur in the other.
- **`VERIFYING`** — independent routes to the **same** result. Requires **both**:
  1. **computationally independent** — an error in one would *not* appear in the other; and
  2. **substantive** — real multi-step work, not a one-line restatement or a rescaling.

  Also record `verification_target`: `answer` (recomputes the answer another way) | `method` (validates the technique, possibly on other inputs) | `input` (re-establishes a premise another worker uses); and `substantive`: `0`/`1`. **If `substantive = 0`, downgrade the pair to `IDENTICAL`.**
- **`DIVIDING`** — each worker produces a different **part** of the answer, and the parts combine.

**Deletion test** (use for `DIVIDING` vs the rest): *"If I delete worker j, is a PART of the answer now missing that worker i does not supply?"* Yes -> `DIVIDING`. No -> `IDENTICAL` or `VERIFYING`, decided by (1)+(2).

### Important refinement — prerequisites

A worker that re-establishes a **prerequisite** another worker needs is only "not distinct" when that prerequisite is **trivial relative to the main computation**. The test: **does splitting it off save the other worker real work?**

- Listing the primes <= 15 saves nothing -> `substantive = 0` -> `IDENTICAL`.
- But "factor this 10-digit number" while another worker builds the formula that *consumes* the factorization **is** `DIVIDING`.

### Worked rulings — apply these, do not re-litigate them

**`IDENTICAL`:**
- discriminant vs completing-the-square for "one real root" -> completing the square *is* how the discriminant is derived; one line; an algebra slip recurs in both -> not independent.
- `62.5 + 500/x = 250` vs `125 + 1000/x = 500` vs `1/8 + 1/x = 1/2` -> the second is the first ×2, the third is the first ÷500. **One relation at three unit scalings.**
- Two workers computing `C(6,5)·5!` identically; a third re-verifying the prime list -> that check has `verification_target=input`, `substantive=0` -> `IDENTICAL`.

**`VERIFYING`:**
- Integer recurrence `S_0…S_6` vs floating-point `a^6 = 4096/b^6` for `(3+sqrt(5))^6` -> independent and multi-step; the check actually **caught an arithmetic slip** in one route. `verification_target=answer`.
- multiply-then-reduce vs reduce-then-multiply (mod 6) -> independent intermediates. `verification_target=answer`.
- Validating the rule "units digit in base 6 = product mod 6" on **2×3 and 7×8**, numbers absent from the problem -> `verification_target=method`. It catches *method* errors but cannot catch an arithmetic slip in the real product. **Record it; do not silently discard it.**
- A telescoping algebraic identity vs numeric partial-sum estimation of the same series -> independent of the algebra -> `VERIFYING`; but two workers *both* doing the numeric estimation are `IDENTICAL` with each other.

**`DIVIDING`:**
- differentiate -> the ODE **+** evaluate at x=0 -> the initial condition (complementary **pieces**)
- enumerate 15 valid (A,D) pairs **×** derive the 56 (B,C) combinations per pair (**factors** that multiply)
- Case 1 positive factor pairs **+** Case 2 negative factor pairs (**cases** that sum)
- test divisor structures {1,2,3} vs {1,2,4} vs general {1,p,q} (independent **hypotheses**)

Group by **what a worker was doing, not whether it finished.** A `DIVIDING` or `VERIFYING` worker with `work_level = partial` still forms its own group.

## Step 3 — partition, and Step 4 — the Q1 label

Merge every `IDENTICAL` pair into one group; `VERIFYING` and `DIVIDING` groups stay separate. Every worker present appears in exactly one group.

`q1_label` **must** match the group count — there is **no "unclear" option**:
- 1 group -> `all_same`
- 2 groups (of 3 workers) -> `two_same_one_distinct`
- 3 groups -> `all_distinct`

For each `IDENTICAL` pair, give one short verbatim quote from **each** worker showing the overlap.

## Step 5 / 6 — relation flags and multi-round aggregation

Per round: `has_dividing_pair`, `has_verifying_pair`, `has_method_or_input_check` (each 0/1), and `nontrivial_round` = 1 iff >=2 of its workers have `work_level` in {`concluded`, `partial`}.

Trajectory level: `any_dividing` / `any_verifying` = OR over **nontrivial** rounds; `workers_non_redundant` = 1 iff some nontrivial round has >=2 groups.

## Step 7 — Q2 contradictory workers

`true` only if two workers **both with `work_level = concluded`** state incompatible substantive claims (different numeric answers, incompatible intermediate values or conclusions). Different methods are not a contradiction; a worker that never concluded cannot contradict anyone. If true, give `q2_claim_a` and `q2_claim_b` as **verbatim quotes** and say why they are incompatible.

## Step 8 — Q3 director misuses worker results

`q3_subcases` uses **single letters only**:
- `a` ignores the workers entirely and re-derives without reference
- `b` misreads or misattributes what a worker concluded
- `c` the workers disagreed and it adopted the wrong one, or failed to resolve it
- `d` claims the workers verified/agreed on something they did not
- `e` discards or contradicts correct, useful worker results

**Hard constraint: `q3_subcases` must be `[]` if and only if `q3_verdict` is false.** Put prose in `q3_subcase_notes`; put anything you considered and rejected in `q3_considered_and_rejected` — **never** in `q3_subcases`.

**Not** a misuse: correctly summarising the workers and then independently double-checking while still crediting them -> that is Q3b.

The common real mechanism: a worker that **stalled** or **self-terminated** without concluding is nonetheless credited with a conclusion. That is `b`/`d`, not `a`.

## Step 9 — Q3b director redundantly reworks

`true` if, after the workers return, the director redoes work they had already completed. Quote the overlap.

## Step 10 — Q4: two INDEPENDENT axes

Judge **the problem**, not this attempt.

**Axis A — can the problem be split into subtasks?**
1. First **enumerate** candidate subtasks (`q4_candidate_subtasks`; may be empty).
2. For each, mark `independent` (doable without another's result) and `substantive` (real progress, not a restatement or trivial check).
3. `a_count` = number that are independent **and** substantive **and** whose results **combine** into the answer.
4. `a_verdict` = `decomposable` if `a_count >= 2`, else `not_decomposable`.
5. `a_reason` = `too_trivial` | `inherently_sequential` | `n/a` (both roll up to `not_decomposable`).

**Axis B — does it admit alternative independent routes to the WHOLE answer?**
- `b_verdict` = `b_full` (>=2 routes each able to produce the complete answer alone) | `b_check` (only corroboration possible: estimate / method-check / prerequisite) | `b_none`
- `b_type` = `textbook_technique` | `identity_vs_direct` | `direct_vs_complementary` | `symbolic_vs_concrete` | `reordering` | `n/a`

**Axis B is not decomposability.** A problem with three alternative solution methods but no separable parts is `a_verdict = not_decomposable`, `b_verdict = b_full`.

## Step 11 — uncertainty: commit first, flag separately

For each of `q1_label`, `q2_verdict`, `q3_verdict`, `a_verdict`, `b_verdict`: **always commit to one value.** Then set `<field>_uncertain` = 0/1, and if 1, give `<field>_alternative` plus a one-line reason. Never use the flag to avoid choosing.

## Output contract

Return exactly these fields. Every field is required unless marked optional.
Do **not** include an `id` field — the rollout label is attached automatically and any value you supply would be ignored.

```json
{
  "rounds": [{
    "round_index": 1,
    "workers": {"worker_1": {"work_level": "concluded|partial|minimal",
                             "termination": "closed_tag|no_close_tag",
                             "cut_off": 0, "task": "one line"}},
    "pair_relations": [{"pair": "worker_1|worker_2",
                        "relation": "IDENTICAL|VERIFYING|DIVIDING",
                        "verification_target": "answer|method|input|n/a",
                        "substantive": 1,
                        "evidence_a": "verbatim", "evidence_b": "verbatim",
                        "why": "<=2 sentences"}],
    "partition": [["worker_1"], ["worker_2", "worker_3"]],
    "n_groups": 2,
    "q1_label": "all_same|two_same_one_distinct|all_distinct",
    "has_dividing_pair": 0, "has_verifying_pair": 1,
    "has_method_or_input_check": 0, "nontrivial_round": 1
  }],
  "any_dividing": 0, "any_verifying": 1, "workers_non_redundant": 1,
  "trajectory_relation": "divided|cross_verification|identical",
  "q1_label_trajectory": "all_same|two_same_one_distinct|all_distinct",
  "q1_uncertain": 0, "q1_alternative": "", "q1_uncertainty_reason": "",
  "q2_verdict": false, "q2_workers": [], "q2_claim_a": "", "q2_claim_b": "", "q2_why": "",
  "q2_uncertain": 0, "q2_alternative": "", "q2_uncertainty_reason": "",
  "q3_verdict": false, "q3_subcases": [], "q3_subcase_notes": "",
  "q3_considered_and_rejected": "", "q3_evidence": "",
  "q3_uncertain": 0, "q3_alternative": "", "q3_uncertainty_reason": "",
  "q3b_verdict": false, "q3b_evidence": "",
  "q4_candidate_subtasks": [{"description": "", "independent": 1, "substantive": 1}],
  "a_count": 0, "a_verdict": "decomposable|not_decomposable",
  "a_reason": "too_trivial|inherently_sequential|n/a",
  "a_uncertain": 0, "a_alternative": "", "a_uncertainty_reason": "",
  "b_verdict": "b_full|b_check|b_none",
  "b_type": "textbook_technique|identity_vs_direct|direct_vs_complementary|symbolic_vs_concrete|reordering|n/a",
  "b_uncertain": 0, "b_alternative": "", "b_uncertainty_reason": "",
  "confidence": "high|medium|low", "notes": ""
}
```

## Automated validation (your output is machine-checked)

Your response is rejected and re-run if: the partition does not cover exactly the workers present; `q1_label` disagrees with `n_groups`; the relation flags disagree with `pair_relations`; `q3_subcases` is non-empty while `q3_verdict` is false (or vice versa) or contains anything outside {a,b,c,d,e}; `a_count >= 2` disagrees with `a_verdict`; or any `cut_off = 1`.

Every failure verdict needs a verbatim quote. Keep each free-text reason to <=2 sentences. Where genuinely ambiguous, choose the conservative option (no failure) and flag it with the `_uncertain` fields — but always commit.
\end{Verbatim}
\end{mybox}

\clearpage
\section{Prompts for Creating SFT Dataset} \label{appendix:sft_data_gen_prompts}

\begin{mybox}[box:sft_sequential_cot_prompt]{Prompt for Generating Sequential CoTs}
\begin{Verbatim}[breaklines=true]
instruction = "Let's think step by step and output the final answer within \\boxed{{}}."
deepseek_r1_prompt = "{question} " + instruction + "<think>\n"
\end{Verbatim}
\end{mybox}

\begin{mybox}[box:claude_parallel_rewrite_prompt]{Prompt for $\claudemodel$ to Rewrite Sequential CoTs with Parallelization}
\begin{Verbatim}
Given a problem, Alice provides a solution in the format <think>some thoughts here</think>final answer here.
Once the <think></think> tags are done, she includes a little bit of discussion, followed by the final answer inside \boxed{{}}.
Your job is to rewrite Alice's solution in a new format, where multiple workers can sometimes collaborate to solve parts of the problem in parallel.
NOTE: When I say "rewrite" Alice's solution, I mean that you should copy most parts verbatim, but insert some parallelism when it makes sense.
There are 3 workers. At any point in the solution where you feel it is appropriate, you come up with 3 tasks, each of which is a subtask of the problem. The 3 workers then solve this task in parallel. To invoke the workers, you use the token <spawn_workers>. When the workers begin thinking, they output <worker_1>, <worker_2>, and <worker_3>, respectively. When the workers are done thinking, they output </worker_1>, </worker_2>, and </worker_3>, respectively. Multiple rounds of parallel thinking and answering may take place. Your response should thus have the following format:
<think> 
some thoughts (which could potentially be very long) and an outline of what each worker should do 
The thoughts here should be copied directly from Alice's solution - but spawn workers when it makes sense
Each time before you spawn workers - give a justification of why the calculation can be parallelized. (This justification should preserve the writing style of Alice's original solution as much as possible, and spell out the thought process step by step.)
- Feel free to make this justification as detailed as you would like. This justification might include some kind of thinking/reflection (for example, in Alice's original solution, you might see some thinking/second-guessing, such as "Wait, what if..." or other phrases along those lines). However if it is truly trivial then it doesn't have to be too long.
Additionally, when you propose subtasks for the workers, give a justification of why they are independent. Worker 1, Worker 2, and Worker 3's tasks should be precise, and should not rely on each other's results. E.g. worker 2 should be able to do its task without the results of Worker 1 or Worker 3's tasks. (This justification should preserve the writing style of Alice's original solution as much as possible, and spell out the thought process step by step.)
- Feel free to make this justification as detailed as you would like. This justification might include some kind of thinking/reflection (for example, in Alice's original solution, you might see some thinking/second-guessing, such as "Wait, what if..." or other phrases along those lines). However if it is truly trivial then it doesn't have to be too long.
<spawn_workers> 
<worker_1> worker 1 thinking - this should be copied directly from Alice's solution almost, but you can edit it very slightly ONLY to remove all references to tasks that are supposed to be done by other workers </worker_1>
<worker_2> worker 2 thinking - this should be copied directly from Alice's solution almost, but you can edit it very slightly ONLY to remove all references to tasks that are supposed to be done by other workers</worker_2>
<worker_3> worker 3 thinking - this should be copied directly from Alice's solution almost, but you can edit it very slightly ONLY to remove all references to tasks that are supposed to be done by other workers </worker_3>
</spawn_workers>
Do some further reflection based on the workers’ thoughts. If needed, use the parallel workers again and again - but if you decide to call the workers, then you must always explicitly state what worker 1, worker 2 and worker 3 should do in the upcoming block
Additionally, this portion should be almost directly copied from Alice's solution, aside from the part where you assign tasks to workers.
For assigning tasks to workers, again you should give justifications of why the calculations can be parallelized, why the tasks are independent, etc.
<spawn_workers>
<worker_1> worker 1 thinking - this should be copied directly from Alice's solution almost, but you can edit it very slightly ONLY to remove all references to tasks that are supposed to be done by other workers </worker_1>
<worker_2> worker 2 thinking - this should be copied directly from Alice's solution almost, but you can edit it very slightly ONLY to remove all references to tasks that are supposed to be done by other workers </worker_2>
<worker_3> worker 3 thinking - this should be copied directly from Alice's solution almost, but you can edit it very slightly ONLY to remove all references to tasks that are supposed to be done by other workers</worker_3>
</spawn_workers>
Further reflection (which can potentially be long), and optionally more rounds of parallel workers.
<spawn_workers>
<worker_1> worker 1 thinking </worker_1>
<worker_2> worker 2 thinking </worker_2>
<worker_3> worker 3 thinking </worker_3>
</spawn_workers>
….
….
Finally, wrap up the thought process after the last round of parallel workers is done
</think>
<answer>
answer here - this part should not contain any other tags such as the thinking tags/worker tags.
Within this block, you may include some discussion prior to the final answer.
The final answer should always be included in \boxed{{}}.
This part should be essentially copied from the portion of Alice's solution that comes after </think>.
</answer>

From a worker's perspective, it will not see the other workers' thoughts. Thus, worker 2 does not see <worker_1>, </worker_1>, or what goes between these two tags, for example.

====================

VERY IMPORTANT:
- As far as possible, copy text only from Alice's solution, instead of adding your own language.
- Only add your own language (that is not present in Alice's solution) when it is absolutely necessary. E.g. when you give justification for using parallel workers, since this will not be present in Alice's solution. Or for example, if a worker is performing a partial calculation, but does not have enough information to say exactly what Alice would have said.
- Basically, when writing the tasks that each worker is supposed to do, always keep thinking about what information this worker is supposed to have access to. For example, in some algebra problem, if Worker 1 finds that x=5, then within that same round of workers, Workers 2 and 3 should NOT have access to the knowledge that x=5.
- Within workers, make sure to use the same style as Alice. The portion WITHIN the workers should also be COPIED FROM ALICE AS MUCH AS POSSIBLE (with some caveats, as I discussed previously).

VERY IMPORTANT: 
- Pretend you are Alice, but with access to multiple workers. Do NOT use phrases such as "Alice said," etc. You are rewriting the solution so that someone who has not read Alice's solution will be able to completely follow it.
- Include all the trial and error that was done by Alice. In particular, Alice uses a lot of phrases such as "Wait, let me double-check" and "wait, no. Let's think again." Please include all of these phrases in your re-written version
- However, the solution should be completely self-contained with no references to Alice’s original work or to an ‘original solution.’
- Write as if you are solving the problem for the first time using the parallel workers approach, not rewriting someone else’s solution.
- Avoid any meta-commentary about the solution being a rewrite or referring to other solutions.
- Within a <spawn_workers></spawn_workers> block, each worker functions independently and cannot see the work of other workers until after the </spawn_workers> tag. Workers should never reference each others’ work or conclusions withi
n a <spawn_workers> block.
- In other words, when writing worker sections, imagine each worker is in a separate room with no communication until their work is collected at the end of a spawn cycle.

==================== EXAMPLE OF REWRITING TASK

The following is an example of the task that you must accomplish.

The math problem is as follows.

The problem was The matrices\n\\[\\begin{{pmatrix}} a & 1 & b \\\\ 2 & 2 & 3 \\\\ c & 5 & d \\end{{pmatrix}} \\quad \\text{{and}} \\quad \\begin{{pmatrix}} -5 & e & -11 \\\\ f & -13 & g \\\\ 2 & h & 4 \\end{{pmatrix}}\\]are inverses. 
 Find $a + b + c + d + e + f + g + h.$

Alice's solution is as follows:

<think>\nOkay, so I have this problem where two matrices are given, and they're supposed to be inverses of each other. I need to find the sum of all these variables: a, b, c, d, e, f, g, and h. Hmm, let me think about how to approach 
this.\n\nFirst, I remember that if two matrices are inverses, their product should be the identity matrix. So, if matrix A multiplied by matrix B gives the identity matrix, then B is the inverse of A, and vice versa. So, my plan is to
 multiply these two matrices together and set each element of the resulting matrix equal to the corresponding element of the identity matrix. That should give me a system of equations to solve for all the variables.\n\nLet me write do
wn the matrices again to make sure I have them correctly:\n\nMatrix A:\n\\[\n\\begin{{pmatrix}} \na & 1 & b \\\\ \n2 & 2 & 3 \\\\ \nc & 5 & d \n\\end{{pmatrix}}\n\\]\n\nMatrix B:\n\\[\n\\begin{{pmatrix}} \n-5 & e & -11 \\\\ \nf & -13 
& g \\\\ \n2 & h & 4 \n\\end{{pmatrix}}\n\\]\n\nAnd the identity matrix, which is:\n\\[\n\\begin{{pmatrix}} \n1 & 0 & 0 \\\\ \n0 & 1 & 0 \\\\ \n0 & 0 & 1 \n\\end{{pmatrix}}\n\\]\n\nSo, when I multiply A and B, I should get the identit
y matrix. Let's denote the product of A and B as matrix C, where C = A * B. Then, C should equal the identity matrix. So, I can set up equations for each element of C by equating them to the corresponding elements of the identity matr
ix.\n\nLet me compute each element one by one.\n\nFirst, let's compute the element in the first row and first column of matrix C.\n\nElement C11 = (a)(-5) + (1)(f) + (b)(2) = -5a + f + 2b\n\nThis should equal 1, because the identity m
atrix has 1 in the (1,1) position.\n\nSo, equation 1: -5a + f + 2b = 1\n\nNext, element C12 = (a)(e) + (1)(-13) + (b)(h) = ae - 13 + bh\n\nThis should equal 0, since the identity matrix has 0 in the (1,2) position.\n\nEquation 2: ae -
 13 + bh = 0\n\nThen, element C13 = (a)(-11) + (1)(g) + (b)(4) = -11a + g + 4b\n\nThis should equal 0 as well.\n\nEquation 3: -11a + g + 4b = 0\n\nMoving on to the second row.\n\nElement C21 = (2)(-5) + (2)(f) + (3)(2) = -10 + 2f + 6 
= 2f - 4\n\nThis should equal 0.\n\nEquation 4: 2f - 4 = 0\n\nElement C22 = (2)(e) + (2)(-13) + (3)(h) = 2e - 26 + 3h\n\nThis should equal 1.\n\nEquation 5: 2e - 26 + 3h = 1\n\nElement C23 = (2)(-11) + (2)(g) + (3)(4) = -22 + 2g + 12 
= 2g - 10\n\nThis should equal 0.\n\nEquation 6: 2g - 10 = 0\n\nProceeding to the third row.\n\nElement C31 = (c)(-5) + (5)(f) + (d)(2) = -5c + 5f + 2d\n\nThis should equal 0.\n\nEquation 7: -5c + 5f + 2d = 0\n\nElement C32 = (c)(e) +
 (5)(-13) + (d)(h) = ce - 65 + dh\n\nThis should equal 0.\n\nEquation 8: ce - 65 + dh = 0\n\nFinally, element C33 = (c)(-11) + (5)(g) + (d)(4) = -11c + 5g + 4d\n\nThis should equal 1.\n\nEquation 9: -11c + 5g + 4d = 1\n\nOkay, so now 
I have a system of 9 equations with 8 variables: a, b, c, d, e, f, g, h. Wait, 9 equations and 8 variables? That seems overdetermined, but maybe the equations are consistent.\n\nLet me write down all the equations again:\n\n1. -5a + f
 + 2b = 1\n2. ae - 13 + bh = 0\n3. -11a + g + 4b = 0\n4. 2f - 4 = 0\n5. 2e - 26 + 3h = 1\n6. 2g - 10 = 0\n7. -5c + 5f + 2d = 0\n8. ce - 65 + dh = 0\n9. -11c + 5g + 4d = 1\n\nHmm, okay, let's try to solve these step by step.\n\nStartin
g with equation 4: 2f - 4 = 0. That should be straightforward.\n\nEquation 4: 2f - 4 = 0 => 2f = 4 => f = 2.\n\nOkay, so f = 2.\n\nNow, equation 2: ae - 13 + bh = 0. Since we know f = 2, we can plug that in later.\n\nEquation 6: 2g - 
10 = 0 => 2g = 10 => g = 5.\n\nSo, g = 5.\n\nEquation 3: -11a + g + 4b = 0. We know g = 5, so:\n\n-11a + 5 + 4b = 0 => -11a + 4b = -5. Let's call this equation 3a.\n\nEquation 7: -5c + 5f + 2d = 0. f = 2, so:\n\n-5c + 10 + 2d = 0 => -
5c + 2d = -10. Let's call this equation 7a.\n\nEquation 5: 2e - 26 + 3h = 1 => 2e + 3h = 27. Let's call this equation 5a.\n\nEquation 1: -5a + f + 2b = 1. f = 2, so:\n\n-5a + 2 + 2b = 1 => -5a + 2b = -1. Let's call this equation 1a.\n
\nEquation 8: ce - 65 + dh = 0. Let's keep this as equation 8.\n\nEquation 9: -11c + 5g + 4d = 1. g = 5, so:\n\n-11c + 25 + 4d = 1 => -11c + 4d = -24. Let's call this equation 9a.\n\nSo now, let's recap the equations:\n\n1a: -5a + 2b 
= -1\n2: ae - 13 + bh = 0\n3a: -11a + 4b = -5\n4: f = 2\n5a: 2e + 3h = 27\n6: g = 5\n7a: -5c + 2d = -10\n8: ce - 65 + dh = 0\n9a: -11c + 4d = -24\n\nOkay, so equation 1a and equation 3a are two equations with variables a and b.\n\nLet
 me write them again:\n\n1a: -5a + 2b = -1\n3a: -11a + 4b = -5\n\nLet me solve these two equations for a and b.\n\nLet me use the elimination method.\n\nMultiply equation 1a by 2: -10a + 4b = -2\n\nEquation 3a: -11a + 4b = -5\n\nNow, 
subtract the first result from the second:\n\n(-11a + 4b) - (-10a + 4b) = (-5) - (-2)\n\n-11a + 4b +10a -4b = -5 +2\n\n(-1a) = -3\n\nSo, -a = -3 => a = 3\n\nOkay, so a = 3.\n\nNow, plug a = 3 into equation 1a: -5(3) + 2b = -1 => -15 +
 2b = -1 => 2b = 14 => b = 7\n\nSo, b = 7.\n\nSo now, we have a = 3, b =7, f=2, g=5, c and d are variables to solve, and e and h are variables to solve.\n\nNow, let's look at equations 7a and 9a:\n\n7a: -5c + 2d = -10\n\n9a: -11c + 4d
 = -24\n\nLet me write these as:\n\n-5c + 2d = -10  ...(7a)\n-11c + 4d = -24 ...(9a)\n\nLet me solve these two equations for c and d.\n\nLet me use elimination again.\n\nMultiply equation 7a by 2: -10c +4d = -20\n\nEquation 9a: -11c +
4d = -24\n\nNow, subtract the first result from the second:\n\n(-11c +4d) - (-10c +4d) = (-24) - (-20)\n\n-11c +4d +10c -4d = -24 +20\n\n(-1c) = -4\n\nSo, -c = -4 => c = 4\n\nOkay, so c = 4.\n\nNow, plug c =4 into equation 7a: -5(4) +
2d = -10 => -20 +2d = -10 => 2d =10 => d=5\n\nSo, d =5.\n\nNow, we have a=3, b=7, c=4, d=5, f=2, g=5.\n\nNow, we need to find e and h.\n\nLooking back at equation 5a: 2e +3h =27.\n\nAnd equation 8: ce -65 + dh =0. Let's plug c=4, d=5:
\n\n4e -65 +5h =0 => 4e +5h =65 ...(8a)\n\nSo, equation 5a: 2e +3h =27 ...(5a)\n\nEquation 8a: 4e +5h =65\n\nLet me solve these two equations for e and h.\n\nLet me use elimination.\n\nMultiply equation 5a by 2: 4e +6h =54\n\nEquation
 8a: 4e +5h =65\n\nSubtract the first result from the second:\n\n(4e +5h) - (4e +6h) =65 -54\n\n4e +5h -4e -6h =11\n\n(-h) =11 => h= -11\n\nSo, h= -11.\n\nNow, plug h= -11 into equation 5a: 2e +3*(-11)=27 => 2e -33=27 => 2e=60 => e=30
\n\nSo, e=30.\n\nLet me verify these values in equation 8a: 4e +5h=4*30 +5*(-11)=120 -55=65. Yes, that works.\n\nNow, let me check equation 8: ce -65 + dh=4*30 -65 +5*(-11)=120 -65 -55=0. Correct.\n\nGood, so all equations are satisfi
ed.\n\nNow, let's compile all the variables:\n\na=3, b=7, c=4, d=5, e=30, f=2, g=5, h=-11.\n\nSo, now, the question is to find the sum a + b + c + d + e + f + g + h.\n\nLet me compute that:\n\n3 + 7 + 4 +5 +30 +2 +5 + (-11)\n\nCompute
 step by step:\n\nStart with 3+7=10\n\n10+4=14\n\n14+5=19\n\n19+30=49\n\n49+2=51\n\n51+5=56\n\n56 + (-11)=56 -11=45\n\nSo, the total sum is 45.\n\nTherefore, the answer is 45.\n\n**Final Answer**\n\\boxed{{45}}\n</think>\n\nGiven two 
matrices which are inverses of each other, we need to find the sum of all variables in the matrices. \n\nThe matrices are:\n\\[\n\\begin{{pmatrix}} a & 1 & b \\\\ 2 & 2 & 3 \\\\ c & 5 & d \\end{{pmatrix}} \\quad \\text{{and}} \\quad \
\begin{{pmatrix}} -5 & e & -11 \\\\ f & -13 & g \\\\ 2 & h & 4 \\end{{pmatrix}}\n\\]\n\nWe know that the product of these matrices should be the identity matrix. By setting up the equations for each element of the resulting matrix and
 solving them step by step, we find the values of all variables:\n\n1. \\(a = 3\\)\n2. \\(b = 7\\)\n3. \\(c = 4\\)\n4. \\(d = 5\\)\n5. \\(e = 30\\)\n6. \\(f = 2\\)\n7. \\(g = 5\\)\n8. \\(h = -11\\)\n\nSumming these variables:\n\\[\na 
+ b + c + d + e + f + g + h = 3 + 7 + 4 + 5 + 30 + 2 + 5 + (-11) = 45\n\\]\n\nThus, the final answer is:\n\\[\n\\boxed{{45}}\n\\]

One good example of a rewritten solution is as follows.

==================== REWRITTEN SOLUTION

<think>
Okay, so I have this problem where two matrices are given, and they're supposed to be inverses of each other. I need to find the sum of all these variables: a, b, c, d, e, f, g, and h. Hmm, let me think about how to approach this.

First, I remember that if two matrices are inverses, their product should be the identity matrix. So, if matrix A multiplied by matrix B gives the identity matrix, then B is the inverse of A, and vice versa. So, my plan is to multiply
 these two matrices together and set each element of the resulting matrix equal to the corresponding element of the identity matrix. That should give me a system of equations to solve for all the variables.

Let me write down the matrices again to make sure I have them correctly:

Matrix A:
\[
\begin{{pmatrix}} 
a & 1 & b \\ 
2 & 2 & 3 \\ 
c & 5 & d 
\end{{pmatrix}}
\]

Matrix B:
\[
\begin{{pmatrix}} 
-5 & e & -11 \\ 
f & -13 & g \\ 
2 & h & 4 
\end{{pmatrix}}
\]

And the identity matrix, which is:
\[
\begin{{pmatrix}} 
1 & 0 & 0 \\ 
0 & 1 & 0 \\ 
0 & 0 & 1 
\end{{pmatrix}}
\]

So, when I multiply A and B, I should get the identity matrix. Let's denote the product of A and B as matrix C, where C = A * B. Then, C should equal the identity matrix. So, I can set up equations for each element of C by equating th
em to the corresponding elements of the identity matrix.

Let me compute each element one by one. Actually, wait - I notice that I'll need to compute 9 different elements of the resulting matrix. Let me think about whether I can parallelize this computation. Since each element of the product
 matrix is computed independently using the same input matrices, I can have three workers each compute three elements of the resulting matrix. This makes sense because computing C11 doesn't depend on knowing C12 or any other element -
 each element is calculated separately from the original matrices A and B.

Let me assign the work as follows:
- Worker 1 will compute the first row of the product matrix (elements C11, C12, C13) - equations 1 through 3
- Worker 2 will compute the second row of the product matrix (elements C21, C22, C23) - equations 4 through 6
- Worker 3 will compute the third row of the product matrix (elements C31, C32, C33) - equations 7 through 9

These tasks are independent because each worker is computing different elements of the product matrix, and the computation of one element doesn't depend on the result of another element's computation.

<spawn_workers>
<worker_1>
First, let's compute the element in the first row and first column of matrix C.

Element C11 = (a)(-5) + (1)(f) + (b)(2) = -5a + f + 2b

This should equal 1, because the identity matrix has 1 in the (1,1) position.

So, equation 1: -5a + f + 2b = 1

Next, element C12 = (a)(e) + (1)(-13) + (b)(h) = ae - 13 + bh

This should equal 0, since the identity matrix has 0 in the (1,2) position.

Equation 2: ae - 13 + bh = 0

Then, element C13 = (a)(-11) + (1)(g) + (b)(4) = -11a + g + 4b

This should equal 0 as well.

Equation 3: -11a + g + 4b = 0
</worker_1>
<worker_2>
Let's compute the elements in the second row of matrix C.

Element C21 = (2)(-5) + (2)(f) + (3)(2) = -10 + 2f + 6 = 2f - 4

This should equal 0.

Equation 4: 2f - 4 = 0

Element C22 = (2)(e) + (2)(-13) + (3)(h) = 2e - 26 + 3h

This should equal 1.

Equation 5: 2e - 26 + 3h = 1

Element C23 = (2)(-11) + (2)(g) + (3)(4) = -22 + 2g + 12 = 2g - 10

This should equal 0.

Equation 6: 2g - 10 = 0
</worker_2>
<worker_3>
Let's compute the elements in the third row of matrix C.

Element C31 = (c)(-5) + (5)(f) + (d)(2) = -5c + 5f + 2d

This should equal 0.

Equation 7: -5c + 5f + 2d = 0

Element C32 = (c)(e) + (5)(-13) + (d)(h) = ce - 65 + dh

This should equal 0.

Equation 8: ce - 65 + dh = 0

Finally, element C33 = (c)(-11) + (5)(g) + (d)(4) = -11c + 5g + 4d

This should equal 1.

Equation 9: -11c + 5g + 4d = 1
</worker_3>
</spawn_workers>

Okay, so now I have a system of 9 equations with 8 variables: a, b, c, d, e, f, g, h. Wait, 9 equations and 8 variables? That seems overdetermined, but maybe the equations are consistent.

Let me write down all the equations again:

1. -5a + f + 2b = 1
2. ae - 13 + bh = 0
3. -11a + g + 4b = 0
4. 2f - 4 = 0
5. 2e - 26 + 3h = 1
6. 2g - 10 = 0
7. -5c + 5f + 2d = 0
8. ce - 65 + dh = 0
9. -11c + 5g + 4d = 1

Hmm, okay, let's try to solve these step by step. Wait, I notice there are some simple equations that can be solved immediately, and then there are pairs of linear equations. Maybe I can parallelize this:

- Worker 1 can solve the simple equations for f and g (equations 4 and 6)
- Worker 2 can work on the system of equations involving a and b (equations 1 and 3, after substituting f and g)
- Worker 3 can work on the system of equations involving c and d (equations 7 and 9, after substituting f and g)

Actually wait, I need to solve for f and g first before I can work on the other systems. Let me reconsider. Worker 1 will solve for f and g from the simple equations. Then once we have those values, I can proceed with the other calcul
ations. But actually, the computation of f from equation 4 and g from equation 6 are independent of each other, and then using these values to simplify the other equations is also independent work that can be parallelized.

Let me think more carefully. Equations 4 and 6 give us f and g directly. Then equations 1 and 3 form a system for a and b. Equations 7 and 9 form a system for c and d. And equations 5 and 8 form a system for e and h. These three syste
ms are independent once we know f and g.

So let me have:
- Worker 1 solve equations 4 and 6 for f and g, then use these to solve the system of equations 1 and 3 for a and b
- Worker 2 solve equations 4 and 6 for f and g, then use these to solve the system of equations 7 and 9 for c and d  
- Worker 3 solve equations 4 and 6 for f and g, then prepare equations 5 and 8 for solving e and h

Wait, that's redundant. Each worker would be solving for f and g separately. Let me think again...

Wait, we can solve for f and g first, then assign tasks to workers:

Starting with equation 4: 2f - 4 = 0. That should be straightforward.

Equation 4: 2f - 4 = 0 => 2f = 4 => f = 2.

Okay, so f = 2.

Equation 6: 2g - 10 = 0 => 2g = 10 => g = 5.

So, g = 5.

Now I can substitute these values into the other equations to simplify them. Let me have three workers work on this in parallel:
- Worker 1 will substitute f = 2 and g = 5 into equations 1 and 3, then solve for a and b
- Worker 2 will substitute f = 2 and g = 5 into equations 7 and 9, then solve for c and d  
- Worker 3 will simplify equation 5 (which doesn't need f or g) to prepare for solving e and h

These tasks are independent because:
- Worker 1 is solving for a and b using only equations 1 and 3 with known values of f and g
- Worker 2 is solving for c and d using only equations 7 and 9 with known values of f and g
- Worker 3 is working with equation 5 which doesn't involve a, b, c, d, f, or g
Each worker can complete their task without needing results from the other workers.

<spawn_workers>
<worker_1>
Let's try to simplify equations 1 and 3. Let's start with equation 3.

Equation 3: -11a + g + 4b = 0. We know g = 5, so:

-11a + 5 + 4b = 0 => -11a + 4b = -5. Let's call this equation 3a.

Equation 1: -5a + f + 2b = 1. f = 2, so:

-5a + 2 + 2b = 1 => -5a + 2b = -1. Let's call this equation 1a.

So equation 1a and equation 3a are two equations with variables a and b.

Let me write them again:

1a: -5a + 2b = -1
3a: -11a + 4b = -5

Let me solve these two equations for a and b.

Let me use the elimination method.

Multiply equation 1a by 2: -10a + 4b = -2

Equation 3a: -11a + 4b = -5

Now, subtract the first result from the second:

(-11a + 4b) - (-10a + 4b) = (-5) - (-2)

-11a + 4b +10a -4b = -5 +2

(-1a) = -3

So, -a = -3 => a = 3

Okay, so a = 3.

Now, plug a = 3 into equation 1a: -5(3) + 2b = -1 => -15 + 2b = -1 => 2b = 14 => b = 7

So, b = 7.
</worker_1>
<worker_2>
Starting with equation 7:

Equation 7: -5c + 5f + 2d = 0. f = 2, so:

-5c + 10 + 2d = 0 => -5c + 2d = -10. Let's call this equation 7a.

Equation 9: -11c + 5g + 4d = 1. g = 5, so:

-11c + 25 + 4d = 1 => -11c + 4d = -24. Let's call this equation 9a.

Now, let's look at equations 7a and 9a:

7a: -5c + 2d = -10

9a: -11c + 4d = -24

Let me write these as:

-5c + 2d = -10  ...(7a)
-11c + 4d = -24 ...(9a)

Let me solve these two equations for c and d.

Let me use elimination again.

Multiply equation 7a by 2: -10c +4d = -20

Equation 9a: -11c +4d = -24

Now, subtract the first result from the second:

(-11c +4d) - (-10c +4d) = (-24) - (-20)

-11c +4d +10c -4d = -24 +20

(-1c) = -4

So, -c = -4 => c = 4

Okay, so c = 4.

Now, plug c =4 into equation 7a: -5(4) +2d = -10 => -20 +2d = -10 => 2d =10 => d=5

So, d =5.
</worker_2>
<worker_3>
Let's look at equation 5.

Equation 5: 2e - 26 + 3h = 1 => 2e + 3h = 27. Let's call this equation 5a.

So equation 5a is: 2e + 3h = 27

This equation involves e and h, which we'll need to solve for later once we know the other variables.
</worker_3>
</spawn_workers>

So now, we have a = 3, b = 7, f = 2, g = 5 from Worker 1, and c = 4, d = 5 from Worker 2, and we have equation 5a: 2e + 3h = 27 from Worker 3.

Now, we need to find e and h.

Looking back at equation 5a: 2e + 3h = 27.

And equation 8: ce - 65 + dh = 0. Let's plug c = 4, d = 5:

4e - 65 + 5h = 0 => 4e + 5h = 65 ...(8a)

So, equation 5a: 2e + 3h = 27 ...(5a)

Equation 8a: 4e + 5h = 65

Let me solve these two equations for e and h.

Let me use elimination.

Multiply equation 5a by 2: 4e + 6h = 54

Equation 8a: 4e + 5h = 65

Subtract the first result from the second:

(4e + 5h) - (4e + 6h) = 65 - 54

4e + 5h - 4e - 6h = 11

(-h) = 11 => h = -11

So, h = -11.

Now, plug h = -11 into equation 5a: 2e + 3*(-11) = 27 => 2e - 33 = 27 => 2e = 60 => e = 30

So, e = 30.

Let me verify these values in equation 8a: 4e + 5h = 4*30 + 5*(-11) = 120 - 55 = 65. Yes, that works.

Now, let me check equation 8: ce - 65 + dh = 4*30 - 65 + 5*(-11) = 120 - 65 - 55 = 0. Correct.

Good, so all equations are satisfied.

Now, let's compile all the variables:

a = 3, b = 7, c = 4, d = 5, e = 30, f = 2, g = 5, h = -11.

So, now, the question is to find the sum a + b + c + d + e + f + g + h.

Let me compute that:

3 + 7 + 4 + 5 + 30 + 2 + 5 + (-11)

Compute step by step:

Start with 3 + 7 = 10

10 + 4 = 14

14 + 5 = 19

19 + 30 = 49

49 + 2 = 51

51 + 5 = 56

56 + (-11) = 56 - 11 = 45

So, the total sum is 45.

Therefore, the answer is 45.
</think>

<answer>
Given two matrices which are inverses of each other, we need to find the sum of all variables in the matrices. 

The matrices are:
\[
\begin{{pmatrix}} a & 1 & b \\ 2 & 2 & 3 \\ c & 5 & d \end{{pmatrix}} \quad \text{{and}} \quad \begin{{pmatrix}} -5 & e & -11 \\ f & -13 & g \\ 2 & h & 4 \end{{pmatrix}}
\]

We know that the product of these matrices should be the identity matrix. By setting up the equations for each element of the resulting matrix and solving them step by step, we find the values of all variables:

1. \(a = 3\)
2. \(b = 7\)
3. \(c = 4\)
4. \(d = 5\)
5. \(e = 30\)
6. \(f = 2\)
7. \(g = 5\)
8. \(h = -11\)

Summing these variables:
\[
a + b + c + d + e + f + g + h = 3 + 7 + 4 + 5 + 30 + 2 + 5 + (-11) = 45
\]

Thus, the final answer is:
\[
\boxed{{45}}
\]
</answer>

==================== END REWRITTEN SOLUTION

COMMENTARY ON REWRITTEN SOLUTION
1. Notice that the rewritten solution is almost exactly the same as Alice's solution, with the text copied verbatim. The exceptions to this are when the rewritten solution assigns tasks to workers - this is new, because Alice's soluti
on does everything sequentially. In those portions, some extra reasoning is used, to assign the tasks to workers, and to justify that these tasks are independent.
2. In the rewritten solution, pay attention to the portion where the workers are called for the second time (right after equations 1-9 have been defined). Here, the rewritten solution initially comes up with some tasks, but realizes t
hat these tasks are in fact not independent, and therefore does some thinking to correct itself and come up with some new tasks which are in fact independent.
    (2a) There might be cases where you should do this, in the problems/solutions that I ask you to rewrite in a parallel format. If you realize that the tasks you assign to Workers 1, 2, and 3 are not independent, then feel free to c
orrect yourself, and include the type of dialogue that is there in this example rewritten solution (e.g. with phrases such as "Actually wait, ..." or "Let me reconsider...").
    (2b) However, you don't always have to do this - there may be a decent amount of cases where it is pretty clear to you what the tasks should be. If you feel that your initial assignment of tasks to Workers 1, 2 and 3 is good, then
 you don't have to second-guess yourself unless you want to.
3. Notice that the text inside the <worker_1></worker_1> tags, or the <worker_2></worker_2> and <worker_3></worker_3> tags are copied pretty much entirely from parts of Alice's solution. However, there are some occasional exceptions w
here we make some minor modifications:
    (3a) In the first round of parallel workers, we made some slight modifications so that the text within the different workers' tags are fully independent. For example, in the start of Worker 2's text, we say "Let's compute the elem
ents in the second row of matrix C." whereas in Alice's original solution, it said "Moving on to the second row." after Alice was done with the first row. It would not make much sense for Worker 2 to say "Moving on to the second row."
 since Worker 2 did not compute the first row to begin with - that was Worker 1's job.
    (3b) Similarly, for Worker 3, in the rewritten solution we start by saying "Let's compute the elements in the third row of matrix C." while in Alice's original solution it said "Proceeding to the third row."
4. Notice that in the task description in the first round of spawning parallel workers, we explicitly spell out that "Worker 1 will compute the first row of the product matrix (elements C11, C12, C13) - equations 1 through 3" - note h
ow we say "equations 1 through 3" and similarly, in the job descriptions for Worker 2, we say "equations 4 through 6" and in the job description for Worker 3, we say "equations 7 through 9".
    (4a) This is how we further spell out exactly what each worker is responsible for.
    (4b) Notice that in Worker 1's response, it defines Equations 1-3, Worker 2 defines Equations 4-6, and Worker 3 defines Equations 7-9. This is only possible because we specifically told each worker how to number their equations, w
hen we were assigning jobs. Otherwise, it would not have been possible for each worker to know how to number its equations, or at least it would be pretty ambiguous.

VERY IMPORTANT REMINDER:
- Include the final answer within the <answer></answer> tags inside \boxed{{}}, just as Alice does.

====================

The following is a problem, and Alice's solution. I would like you to rewrite Alice's solution. Please just respond with your parallelized rewritten version of Alice's solution, and don't say anything besides that.

Problem: {question}
Alice's solution: <think>\n{assistant_response}
\end{Verbatim}
\end{mybox}

\end{document}